\documentclass[11pt]{article}

\usepackage{fullpage}
\usepackage{natbib}

\usepackage[utf8]{inputenc} % allow utf-8 input
\usepackage[T1]{fontenc}    % use 8-bit T1 fonts
\usepackage{hyperref}       % hyperlinks
\usepackage{url}            % simple URL typesetting
\usepackage{booktabs}       % professional-quality tables
\usepackage{amsfonts}       % blackboard math symbols
\usepackage{nicefrac}       % compact symbols for 1/2, etc.
\usepackage{microtype}      % microtypography
\usepackage{xcolor}         % colors

\usepackage{graphicx}
\usepackage{amsmath}

\newtheorem{theorem}{Theorem}[section]
\newtheorem{lemma}{Lemma}[section]
\usepackage{algorithmic}
\usepackage[ruled,vlined]{algorithm2e}
\usepackage{subfigure}
\usepackage{float,psfrag,epsfig,color,xcolor,url}
\usepackage{diagbox}

\newcommand{\aname}{AIHF}

\usepackage{graphicx}
\usepackage{amsmath}
\usepackage{algorithmic}
\usepackage[ruled,vlined]{algorithm2e}
\usepackage{subfigure}
\newtheorem{assumption}{Assumption}

\def\ee{\mathbb{E}}
\def\soft{\text{soft}}

\title{Learning Reward and Policy Jointly from Demonstration and Preference Improves Alignment}

\author{Chenliang Li, Siliang Zeng, Zeyi Liao, Jiaxiang Li, \\
Dongyeop Kang, Alfredo Garcia, Mingyi Hong\footnote{S.~Zeng, J.~Li and M.~Hong are with Department of Electrical and Computer Engineering, University of Minnesota, Minneapolis, MN, USA. E-mails: \texttt{zeng0176@umn.edu}, \texttt{li003755@umn.edu}, \texttt{mhong@umn.edu}.  Z. Liao is with the  Department of Computer Science and Engineering, The Ohio State University, Columbus, OH, USA. E-mail: \texttt{liao.629@osu.edu}.
D. Kang is with the Department of Computer Science and Engineering, University of Minnesota, Minneapolis, MN, USA. E-mail: \texttt{dongyeop@umn.edu}. C.~Li and A.~Garcia are with Department of Industrial and Systems Engineering, Texas A\&M University, College Station, TX, USA. E-mails: \texttt{chenliangli@tamu.edu}, \texttt{alfredo.garcia@tamu.edu}.}}

\usepackage{amsfonts,amscd,dsfont,mathrsfs,nicefrac, bm,amsmath,amssymb}

\begin{document}

\maketitle

\begin{abstract}
Aligning to human preferences and/or intentions is an important requirement for contemporary foundation models. To ensure alignment, popular approaches such as reinforcement learning with human feedback (RLHF) break down the task into three stages: (i) a model is computed with supervised fine-tuning (SFT) based upon large demonstrations data, (ii) a reward model (RM) is estimated based upon human feedback data, and (iii) reinforcement learning (RL) is used to further refine the SFT model by optimizing the estimated reward model.  Demonstrations and human feedback data reflect human user preferences in different ways. As a result, the reward model estimate obtained from {\em only} human feedback data is likely not as accurate as a reward model estimate obtained from {\em both} demonstration and human feedback data. A policy model that optimizes the reward model estimate obtained from {\em both} demonstration and human feedback data will likely exhibit better alignment performance. We introduce a tractable algorithm for finding the reward and policy models and provide a finite-time performance guarantee. Additionally, we demonstrate the efficiency of the proposed solution with extensive experiments including alignment problems in LLMs and robotic control problems in MuJoCo. We observe that the proposed solutions outperform the existing alignment algorithm by large margins.
\end{abstract}

\section{Introduction}
% {\bf Background.} 
{As ChatGPT has taken the world by storm, it is clear that AI systems will soon become ubiquitous in our lives. For instance, Large Language Models (LLMs) have been used to solve hard problems including video gaming \citep{berner2019dota,mnih2015human}, autonomous control \citep{bellemare2020autonomous}, and robotic manipulation \citep{kalashnikov2018scalable,kober2008policy}. In this context, the notion of {\em alignment} plays an increasingly important role in the design and training of AI systems. Loosely speaking, alignment refers to the performance guarantee that the AI system will generate outcomes that are intended or preferred by the human user without undesirable side effects or behaviors such as deception \citep{park2023ai} or manipulation \citep{perez2022discovering}. As human user intentions or preferences may vary under specific contexts, it is critical that the AI system adapts to evolving user preferences and/or intentions \citep{leike2018scalable}.

The alignment problem is a learning problem with (at least) three types of input data: the demonstration data (consists of prompts and human-generated continuations), the preference data (consists of prompts and pairs of human-ranked responses), as well as prompts without any responses. Moreover, the process of aligning an LLM model is typically undertaken in successive stages. For example, the well-known RLHF approach adopted by \cite{ouyang2022training} starts with a supervised fine-tuning model (SFT) followed by reward model (RM) estimation based upon human-labeled preference data. The process closes with a final alignment stage in which reinforcement learning (RL) is used to optimize the estimated reward model. Similar strategies have been used in other related works such as \cite{rafailov2023direct,li2023remax,zhu2023fine,liu2023statistical}.
The approach to alignment based on successive stages may facilitate computation, but it is at the expense of inefficient exploitation of data. To illustrate, consider the three-stage RLHF approach proposed in \cite{ouyang2022training}, in the extreme case where the amount of high-quality preference data is quite limited, the reward model trained cannot adequately reflect the preferences of the human, which may lead to unsatisfactory performance in the RL stage. Further, the reward model estimate obtained from {\em only} the preference data fails to exploit the information about human users' preferences that are implicit in demonstration data. It is therefore reasonable to expect that a policy model that is fine-tuned with the reward model estimate obtained from {\em both} demonstration and human feedback may exhibit better alignment performance.

An alternative to the successive approach to alignment consists of {\em jointly} training the reward and policy models by leveraging demonstration and preference data. In contrast to the successive approach adopted in most of the current alignment approaches, the joint approach to reward and policy learning makes use of all available data, hence mitigating the risk of optimizing an inaccurate reward model. However, a joint approach to learning reward and policy models may improve alignment at the expense of potentially significant additional computational effort.

{\bf Contribution.} We introduce an algorithm jointly learning reward and policy models named Alignment with Integrated Human Feedback (AIHF) with a finite-time performance guarantee.} This approach leverages recent advances in Inverse Reinforcement Learning (IRL) {\citep{arora2021survey,zeng2022maximum}}, stochastic choice theory {\citep{blavatskyy2010models} and bi-level optimization \citep{hong2020two,ji2021bilevel,khanduri2021near}.  The proposed formulation integrates SFT, RM, and RL into a single stage, so that reward modeling and policy optimization can {\it fully} leverage all the available human feedback data.} More specifically, in the proposed algorithm, the policy is updated to improve alignment with the current reward model estimate and the reward model is updated to improve the fit to demonstration and human feedback data. As a result, upon convergence, the resulting reward and policy models are {\em consistent} in the sense that (i) the policy model is optimal with respect to the reward model and (ii) the reward model maximizes the fit to both demonstration and human feedback data. Several existing alignment schemes, such as RLHF \citep{ouyang2022training}  and DPO \citep{rafailov2023direct} and some of their extensions can be seen as particular instances of the proposed formulation.
We provide ample empirical evidence that the proposed AIHF solution outperforms the existing alignment algorithms by large margins, especially when the data is {\it unbalanced}, where the quality and/or quantity of one data category is worse/smaller than that of the other.
\section{Preliminaries and Related Work}
\subsection{Notation}
{\bf The Finite-Horizon MDP Model.}  A Markov decision process (MDP) is the tuple  $(\mathcal{S}, \mathcal{A}, P, \rho, r, \gamma)$, wherein $\mathcal{S}$ denotes the state space, $\mathcal{A}$ denotes the action space, $P:\mathcal{S} \times \mathcal{A} \times \mathcal{S} \to [0,1]$ denotes the transition probabilities, $\rho(\cdot)$ is the initial state distribution, $r:\mathcal{S} \times \mathcal{A} \to \mathbb{R}$ denotes the reward function and $\gamma \in (0,1)$ denotes the discount factor. For every $s_t \in \mathcal{S}$, a randomized policy $\pi(\cdot|s_t)$ is a probability distribution in $\Delta_{|\mathcal{A}|}$, the unit simplex in $\mathbb{R}^{|A|}$. Define $\tau :=\{(s_t,a_t)\}_{t=1}^{T}$ as a (finite horizon $T$) trajectory of state and action pairs. Let $\mathcal{H}_T \subset \prod_{t=1}^{T} \big(\mathcal{S} \times \mathcal{A}\big)$ denote all feasible state/action sequence of length $T$.

{\bf MDP Model of LLM.} The generation of text by a language model can be seen as sampling from policies in an MDP model. Specifically, each state $s_t= (x,y_{1:t-1})$ includes the prompt $x$ and all response tokens produced up to that point $y_{1:t-1}$. Each action $a_t= y_t$ represents a token from the vocabulary. The transition kernel $P$ is deterministic, i.e. given tokens $s_t=(x, y_{1:t-1})$ and $a_t = y_t$, the environment will transition to $s_{t+1}=(x, y_{1:t})$.
An LLM can be seen as a policy $\pi(\cdot|s_t)$ so that a response of length $T>0$ to prompt $x$ is obtained with probability:$\pi(y_{1:T}|x):=\prod_{i=1}^T \pi(y_i|x,y_{1:i-1})$

{\bf Human Feedback Data.} 
Let $\tau:= (y_{1:T},x)$ denote a finite text produced in response to prompt $x$. For a pair of sequences $(\tau_l, \tau_w)$ (which we assume of the same length $T$ for ease of exposition) we write $\tau_l \prec \tau_w$ to indicate the sequence $\tau_w$ is preferred over the sequence $\tau_l$. Following the Bradley-Terry-Luce (BTL) model \citep{bradley1952rank},  the distribution of preferences over pairs $(\tau_l, \tau_w)$ can be modeled as follows:
{\small \begin{align}
    P\big(\tau_w \succ \tau_l\big) =  \frac{\exp R(\tau_w; \theta)}{\exp R(\tau_w; \theta)+\exp R(\tau_l; \theta)}= \sigma \big( R(\tau_w; \theta) - R(\tau_l;\theta) \big)
\label{eq:BTL_models}
\end{align}}
where $\sigma$ is the sigmoid function and $R(\tau; \theta):= \sum_{t\ge 1}^T \gamma^t r(s_t,a_t; \theta)$ and $r(s_t,a_t; \theta)$ is a reward model parametrized by $\theta \in \mathbb{R}^d$.

\subsection{The RLHF Pipeline}

RLHF is a popular technique for fine-tuning AI systems to align with human preferences and values.  The RLHF approach proposed in \citet{stiennon2020learning,ouyang2022training} consists of the following three-stage: 1) the {\bf supervised fine-tuning (SFT}) stage, where the demonstration data is used to fine-tune the model in a supervised manner; 2) the {\bf reward modeling (RM)} stage, where the preference data is used to train a reward model; 3) {\bf the reinforcement learning (RL)} stage, where the SFT model is further fine-tuned by running RL using the trained reward model. Specifically, the RLHF pipeline can be formally described as follows:\\
\noindent {\bf Supervised Fine-Tuning (SFT)}: Given a demonstration dataset $\mathcal{D}$ consisting of sequences of the form $\tau =\{(s_t,a_t)\}_{t\geq0}$ the goal is the find the policy $\pi_{\rm SFT}(\cdot|s_t)$ that maximizes likelihood, i.e.:
\begin{align} \label{eq:SFT_loss}
\pi_{\rm SFT}&=\arg \max_{\pi}
\mathbb{E}_{\tau \sim \mathcal{D}}\Big[
\log \prod_{t \geq 0}  \Big(\pi(a_{t}|s_{t})\Big)^{\gamma^t}\Big]
\end{align}
\noindent {\bf Reward Modeling (RM)}: Based upon a dataset $\mathcal{P}$ of preferences over pairs $(\tau_l, \tau_w)$ the estimation of a reward learning problem can be formulated as the following Bradley-Terry-Luce (BTL) model \citep{bradley1952rank} (with $\beta>0$ a hyper-parameter):
\begin{align}
\max_{\theta \in \mathbb{R}^d} \ell_{\rm RM}(\theta)
:=  \mathbb{E}_{(\tau_w\succ\tau_l)\in \mathcal{P}} \Big[ \log \Big( \sigma \big( \frac{1}{\beta}\big(R(\tau_w; \theta) - R(\tau_l;\theta) \big) \Big) \Big] \label{eq:preference_gap_maximization}. 
\end{align}
\\
\noindent {\bf Reinforcement Learning (RL)}: Let $\hat{\theta}_{\mathcal{P}}$ denote the solution to problem (\ref{eq:preference_gap_maximization}). The last stage in the RLHF development pipeline consists of solving the problem:
\begin{align}
\pi_{\rm RLHF} &= \arg \max_{\pi} \mathbb{E}_{\tau \sim \pi} \Big[ \sum_{ t \geq 0} \gamma^{t}\big[ r(s_{t}, a_{t};\hat{\theta}_{\mathcal{P}})-\beta {D}_{\rm KL}\Big( \pi(\cdot | s_t) \| \pi_{\rm SFT}(\cdot | s_t) \Big)\Big]
\end{align}
where 
${D}_{\rm KL}\Big( \pi(\cdot | s_t) \| \pi_{\rm SFT}(\cdot | s_t) \Big):=
\sum_{a \in \mathcal{A}}\pi(a|s_{t})\log \frac{\pi(a|s_{t})}{\pi_{\rm SFT}(a| s_t)}$ is the Kullback-Leibler (KL) divergence, $\pi_{\rm SFT}$ is the supervised fine-tuning model. 

\subsection{Reward Learning using Demonstration Data}
% {\red TODO: add more related works}
In the RL literature, a line of work referred to as Inverse Reinforcement Learning (IRL) proposes to {\it jointly} learn the reward and policy from expert demonstration data.  Specifically, the target is to find the parameterized reward function  $r(s,a; \theta)$ (resp. an optimal polity $\pi^*(s,a)$) that best explains (resp. mimics) an expert policy $\pi^E$ given the demonstration data $\mathcal{D}$. 
For example, the well-known maximum entropy IRL  (MaxEnt-IRL) framework  \citep{ziebart2010modeling,ziebart2013principle,bloem2014infinite, zhou2017infinite} finds a policy maximizing entropy subject to
the expected features that match the empirical averages in the expert’s observation dataset. However, this approach can only be used to model linear rewards. 

Subsequent works such as \citet{levine2011nonlinear,wulfmeier2015maximum,zeng2022maximum} further improve the MaxEnt-IRL method so that nonlinear reward can be used.   
For example, \citet{zeng2022structural} proposed a maximum likelihood IRL (ML-IRL) formulation based on the Dynamic Discrete Choice (DDC) model, and a nonlinear reward function is utilized. 

It is also shown that when the reward function is linearly parameterized, then the MaxEnt is the Lagrangian dual of the ML-IRL problem \citep{zeng2022structural}.  On the other hand, it is worth mentioning that to our best knowledge, almost all IRL-based methods can only utilize demonstration data, which can be problematic because a large amount of high-quality demonstration data is typically hard to obtain. Further, it is well-known that using demonstration only cannot extract precise human preference, especially in safety-related tasks where the boundaries between permissible and impermissible actions need to be precisely determined; see, e.g., \citet{fischer2021sampling} which shows that insufficient demonstration dataset could lead to high generalization error.
\subsection{Joint Learning from demonstration and preference}

Combining data from demonstrations and human feedback to achieve alignment has also been studied in the robotics literature. {In \cite{ibarz2018reward}, the authors first combine two approaches to
learn from human feedback: expert demonstrations and trajectory preferences. The addition of demonstrations to learning from preferences typically results in substantial performance gains compared with using either demonstrations or preferences in isolation.}
In \cite{palan2019learning} and \cite{biyik2022learning}, the authors integrate diverse sources of human feedback including demonstrations and pairwise comparisons in a Bayesian approach to learn reward functions that are assumed to be linear in carefully selected features and evaluate their proposed method on robot learning platform. Moreover, their proposed methods need to actively generate preference queries, which are expensive to collect in practical applications. In contrast, the approach proposed in this paper is not Bayesian and does not include the requirement that the reward model is linear in pre-selected features. 
\subsection{Other Approaches to Alignment} 
\label{sec:integration}

Other approaches to alignment include Direct Preference optimization (DPO) \citep{rafailov2023direct} and Inverse Preference Learning (IPL) \citep{hejna2023inverse} both remove the need for explicit reward modeling, and they directly extract the policy from preferences.  This greatly reduced the training complexity, but it has been observed that these algorithms can be unstable in the training process \citep{azar2023general,xu2024dpo}. %and it will be impossible to re-train a new policy. 
There is also a large number of works that aim to learn reward functions from rating \citep{daniel2014active} {or ranking} \citep{yuan2023rrhf,myers2022learning}. %{\red[do you meant to say learning from rating and ranking together?]}. 
\cite{hong2024orpo}
proposed a single-stage supervised learning algorithm ORPO that can perform supervised
fine-tuning and preference alignment in one training session without maintaining.
However, all of these works highly rely on high-quality human feedback, which is often more difficult and expensive to obtain.

\section{Alignment with Integrated Human Feedback (AIHF)}
As mentioned before, the reward model obtained in (\ref{eq:preference_gap_maximization}) fails to exploit the information about human users' preferences that are implicit in demonstration data. As a result, the fine-tuned model obtained with RLHF may exhibit unsatisfactory alignment performance (this phenomenon will be discussed more concretely in Sec. \ref{sec:why}). Below we introduce a new approach to jointly train reward and policy models by simultaneously leveraging demonstration and human feedback data.

\begin{figure}[t]
\centering
\includegraphics[clip, trim={0cm 8.5cm 1.2cm 0cm}, width=0.99\textwidth]{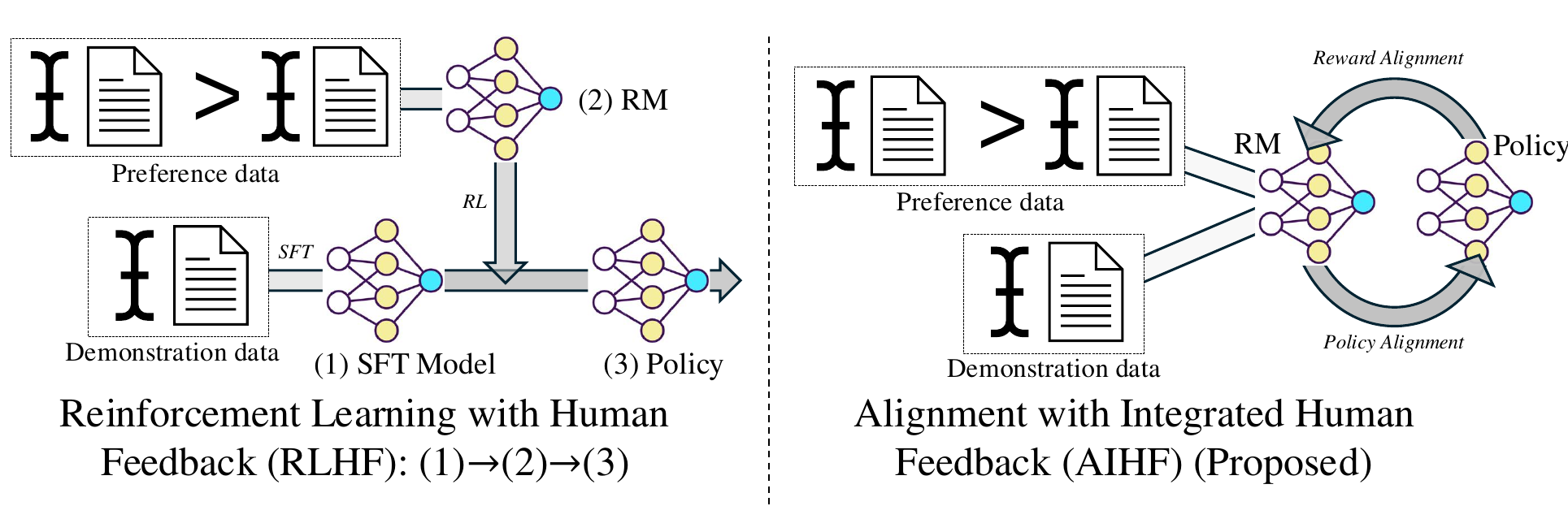}
\caption{Comparison of the RLHF (left) with the proposed \aname \ (right).
}
\label{fig:AIHF}
\end{figure}

\subsection{A Meta-Formulation}

Towards developing an approach that can model the {\it entire} alignment process with a common parametrization for both policy and reward models,  consider the following {\it meta}-formulation, termed Alignment with Integrated Human Feedback (\aname): 
\begin{subequations}\label{eq:AFHF}
\begin{align}
 {\bf (\aname)}\quad \max_{\theta} & \quad L (\theta):=  w_1 L_1(\pi_{\theta}) + L_2(R(\cdot;\theta)) \label{eq:upper}\\
{\rm s.t.} & \quad    \pi_\theta := \arg \max_{\pi} ~ L_3(\pi; R(\cdot;\theta))\label{eq:lower}
\end{align}
\end{subequations}
where $\theta\in\mathbb{R}^d$ is a parameter; $L_1(\pi_\theta)$ is a measure of fit of the parameterized policy $\pi_\theta$ to demonstration data and $L_2(R(\cdot;\theta))$ is a measure of fit of the parameterized reward model 
$R(\cdot;\theta)$ to the preference data and $L_3(\pi,R(\cdot;\theta))$ is a measure of performance of policy $\pi$ with respect to reward model $R(\cdot;\theta)$. $w_1 \geq 0$ is one balancing coefficient reflecting the relative size of demonstration versus preference data. Note that in the lower level policy optimization, the optimal policy corresponding to the reward model $R(\cdot;\theta)$ is actually determined by the reward parameter $\theta$, where we can denote the optimal policy under the reward model $R(\cdot;\theta)$ as $\pi^*_{R_{\theta}}:=\pi_{\theta}$. Therefore we simply put $\pi_{\theta}$ since the optimal policy under a certain reward model $R(\cdot;\theta)$ is actually determined by the reward parameters $\theta$.

See Fig.  \ref{fig:AIHF} for an illustration of AIHF. %We refer the above formulation as ; see  for an illustration of the proposed formulation. %\dk{Here is a good place to refer Figure 1 and mention the high-level difference with RLHF?}
The \aname~\eqref{eq:AFHF} is a {\it meta}-problem that models the alignment problem. It has two levels: an upper-level problem in which the goal is to find policy and reward models that jointly maximize a measure of fit to demonstrations and preference datasets; and a lower-level problem which ensures that the policy model optimizes performance with respect to the reward model. Its components can be customized to yield specific alignment formulations and algorithms. Before diving into various customizations, let us discuss the advantages of this formulation.

\noindent{\bf Generality.}  One can specialize the loss functions and problem parameters to yield a number of existing alignment formulations. Such generality implies that algorithms developed for \eqref{eq:AFHF} are easily applicable to different special formulations it covers. For more details see Sec. \ref{sub:special}.

\noindent{\bf Joint optimization.} The formulation jointly optimizes the reward and the policy. One benefit here is that it can strengthen the reward model through integrating both demonstrations and pairwise comparisons. Compared with the standard RLHF pipeline, through integrating additional data source such as demonstrations to train the reward model, it can further boost the policy optimization subroutine to achieve better alignment performance. See Sec. \ref{sec:algorithm_training} for a detailed discussion on how the reward parameter $\theta$ is updated by leveraging such demonstration, and see Tab. \ref{tab:reward_bench} for the experimental comparison between the reward model learned by RLHF and by our AIHF \eqref{eq:AFHF}.

\noindent{\bf Dataset Integration.} Clearly, the reward learning process leverages all the available data, therefore, we can expect that a high-quality reward model and its induced optimal policy can still be obtained even under unfavorable situations where the preference data is not sufficient. See Sec. \ref{sec:experiments} for experimental evidences.

\subsection{Specification of \aname}
In this section, we specify the formulation \eqref{eq:AFHF}. 
Let us begin with the choice of $L_1$. It can be directly instantiated by using one objective similar to \eqref{eq:SFT_loss}, which is the likelihood function over the collected expert demonstrations.  Note that we aim to optimize the reward parameter $\theta$ to align with human feedback in \eqref{eq:upper}, thus the objective of $L_1$ can be specialized as a maximum likelihood function 
over expert demonstrations as below:
\begin{align}
    L_1(\pi_{\theta}) :=\mathbb{E}_{\tau \sim \mathcal{D}} \Big[
\log \prod_{t \geq 0}  \Big(\pi_{\theta}(a_{t}|s_{t})\Big)^{\gamma^t}\Big]=\mathbb{E}_{\tau \sim \mathcal{D}} \Big[ \sum_{t\geq0} \gamma^t \log \pi_{\theta}(a_t | s_t) \Big]. \label{def:ml}
\end{align}
Here $\pi_{\theta}$ optimizes the measure of performance $L_3(\pi; R(\cdot;\theta))$ for a reward model $R(\cdot; \theta)$ as:
\begin{align}
     L_3(\pi; R(\cdot;\theta)) := \mathbb{E}_{s_0 \sim \rho, \tau \sim \pi} \bigg[R(\tau;\theta) - \beta \sum_{t\ge 0}\gamma^t {D}_{\rm KL}\Big( \pi(\cdot | s_t) \| \pi^{0}(\cdot | s_t) \Big)\bigg]   \label{eq:L3}
\end{align}
where $\pi^0$ is some initial policy and $\beta>0$ is temperature parameter.

Next, we specify $L_2$. To ensure internal model consistency, we identify the likelihood function for preference data so it is in accordance with the preferences implied by the reward model $R(\cdot;\theta)$ used in the definitions of $L_1$ and $L_3$. 
Thus, the optimal distribution $\mu_{\theta}$ over the set of $T$-long sequence of state-action pairs is defined as follows: 
\begin{align*}
 \mu_{\theta} :=\arg \max_{\mu \in \Delta^T} \mathbb{E}_{\tau \sim \mu}\big[ R(\tau; \theta) - \beta \mathcal{D}_{\rm KL}(\mu||\mu^0)\big]
%\label{ground_truth_sequence}
\end{align*}
where $\Delta_{T}$ denotes the simplex on $\mathcal{H}_T$ and $\mu^0$ is a prior distribution on the trajectories. It can be shown the solution of the above problem is of the form:
$$
\mu_{\theta}(\tau)=
\frac{\mu^0(\tau) \exp\big(R(\tau; \theta)/\beta\big)}{\sum_{\tau' \in \mathcal{H}_T}\mu^0(\tau') \exp\big(R(\tau';\theta)/\beta\big)\big)}.
$$

With this result, we can now obtain a model for the likelihood that sequence $\tau_j$ is preferred over $\tau_j$. 
By the {\em independence of irrelevant alternatives} property \citep{fudenberg2015stochastic} of the optimal choice $\mu_\theta$, when the set of feasible choices is reduced from $\mathcal{H}_T$ to just the the two-tuple
$\{\tau_l,\tau_w\}$, the likelihood that sequence $\tau_w$ is preferred over $\tau_l$ is given by $\mathbb{P}_{\theta}(\tau_w  \succ \tau_l):=\frac{\mu_{\theta}(\tau_w)}{\mu_{\theta}(\tau_l)+\mu_{\theta}(\tau_w)}.$ 
This motivates the choice of $L_2(\theta)$ as the following {\it likelihood function}:
{\small \begin{align*}
L_2(R(\cdot;\theta))  &= \mathbb{E}_{(\tau_w  \succ \tau_l)\in \mathcal{P}} \Big[ \log\frac{ \mu_{\theta}(\tau_w) }{\mu_{\theta}(\tau_w) + \mu_{\theta}(\tau_l)} \Big]\nonumber \\ 
& = \mathbb{E}_{(\tau_w  \succ \tau_l)\in \mathcal{P}} \Big[ \log\frac{\mu^0(\tau_w) \exp\big(R(\tau_w;\theta)\big)}{\mu^0(\tau_w) \exp\big(R(\tau_w;\theta)\big) + \mu^0(\tau_l) \exp\big(R(\tau_l;\theta)\big)} \Big].
\end{align*}}%
%Note that the above objective requires prior knowledge of the policy $\pi^0$ which determines the distribution $\mu^0(\tau)$. 
With $\mu^0$ equal to the uniform distribution on $\mathcal{H}_T$, this model is equivalent to
%Since this may be infeasible in practice when either the prior generative policy or the transition probability is unknown, we consider 
the BTL model \eqref{eq:preference_gap_maximization}: 
% {\red[can we change the template so that we remove all 'equation' in front of (xx)?]}:
\begin{align}
{L}^{\rm BTL}_2(\theta) = \ell_{\rm RM}(\theta) = \mathbb{E}_{(\tau_w  \succ \tau_l)\in\mathcal{P}} \Big[ \log \Big( \sigma \big( R(\tau_w; \theta) - R(\tau_l;\theta) \big) \Big) \Big].  \label{preference_loss:BTL}
\end{align}

\subsection{Special Cases of \aname}\label{sub:special}

Next, we discuss how formulation \eqref{eq:AFHF} can be specialized to some of the known alignment algorithms.

{\bf Specialization to RLHF-Type Approach.} First, if we set the coefficient $w_1=0$ in \eqref{eq:AFHF}, we obtain: 
\begin{align}
\begin{split}
\max_{\theta} & \quad L_2(\theta) \;\; {\rm s.t.}  \;\;    \pi_\theta := \arg \max_{\pi} L_3(\pi; R(\cdot;\theta)).
\label{eq:AFHF_redution_RLHF}
\end{split}
\end{align}
Noticed that now the upper- and lower-level problems are completely decomposable, since the upper-level problem solves for the reward parameterization $\theta$, while the lower-level problem solves for the policy (for the given reward), yielding two separate problems, which are exactly the RM and the RL problems in the typical RLHF approach. 

{\bf Specialization to DPO-Type Approach.} 
Consider the relationship between formulation \eqref{eq:AFHF} with the DPO-type approaches. 
Let us set the following objective function $L_1= \ell_{\rm SFT}$ and  $L_2=\ell_{\rm RM}$, and assume that $T=1$ for the generation process. Relaxing the constraint \eqref{eq:lower} which ensures the policy is optimal w.r.t. a certain parameterized model, we can obtain a DPO-type formulation:
\begin{equation}
\resizebox{\textwidth}{!}{
$
\max_{\pi} ~ L(\pi):= w_1 \cdot \mathbb{E}_{\tau^{\rm E} \sim \pi^{\rm E}} \bigg[ \log \pi(a^{\rm E}|s^{\rm E}) \bigg] +  \mathbb{E}_{(\tau_j\succ\tau_i) \sim \pi^P}\bigg[\log \bigg( \sigma\big(\beta \log \frac{\pi(a_{j}|s_j)}{\pi^0(a_{i}|s_j)}-\beta \log \frac{\pi(a_{i}|s_i)}{\pi^0(a_{j}|s_i)}\big) 
 \bigg) \bigg].
$
}\label{eq:AFHF_redution_dpo}
\end{equation}
The above formulation specializes to \cite{liu2024provably}, which is a slightly generalized version of DPO when {\it both} demonstration and preference data are used. Setting $w_1=0$ reduces to the problem solved by DPO; see \citet[Eq. (2)]{rafailov2023direct}, We refer to the algorithm defined by Eq.~\eqref{eq:AFHF} as Direct AIHF.

\textbf{Specialization to Self-Play Approach.} 
Define  $\ell(\cdot)$ as a monotonic and convex loss function,  consider setting $L_1:=w_1 \cdot \mathbb{E}_{\tau^{E}\in \pi^E, \alpha\in \pi(.|s^E)}\ell \left( R(\tau^E;\theta) - R(\tau;\theta) \right)$, and setting $L_2$ and $L_3$ according to \eqref{preference_loss:BTL} and \eqref{eq:L3}, respectively. Note that the choice of $L_1$ means that given demonstration data, we will find a policy that generates trajectories that match the rewards of the demonstration data.  
Again using the DPO type of reformulation, by substituting the reward expression obtained from the optimal policy \eqref{eq:dpo:pi} to  $L_1$ and selecting the $\sigma(\cdot)$ as $\ell(\cdot)$, then the AIHF problem in this case becomes:
\begin{align}\label{eq:AIHF_SPIN}
\max_{\pi} L(\pi) :=& w_1 \mathbb{E}_{\tau^{\rm E} \sim \pi^{\mathrm{E}}, \tilde{a} \sim \pi(\cdot|s)}\left[\log \sigma \left(\beta \log \frac{\pi(a^{\rm E}|s^{\rm E})}{\pi^0(a^{\rm E}|s^{\rm E})} - \beta \log\frac{\pi(\Tilde{a}|s^{\rm E})}{\pi^0(\Tilde{a}|s^{\rm E})}\right)\right] \nonumber \\
&+ \mathbb{E}_{(\tau_j\succ\tau_i) \sim \pi^P}\bigg[\log \sigma\left(\beta \log \frac{\pi(a_{j}|s_j)}{\pi^0(a_{j}|s_j)}-\beta \log \frac{\pi(a_{i}|s_i)}{\pi^0(a_{i}|s_i)}\right) \bigg].
\end{align}

Note that the first part of the above formulation is similar to what has been proposed in SPIN \citep{chen2024self}, which only utilizes the SFT data. We refer to the algorithm defined by Eq.~\eqref{eq:AIHF_SPIN} as Self-Play AIHF.

\subsection{Why AIHF can outperform two-stage alignment approaches?}
\label{sec:why}

To understand the difference between the proposed approach and the successive stages approach of the standard alignment pipeline, let us consider the a {\em static} setting with action set is $A:=\{\tau_1,\tau_2,\cdots\tau_N\}$, reward function  $R(\cdot):A \mapsto \mathbb{R}$, and demonstration $\mathcal{D}$ and preference dataset $\mathcal{P}$. In what follows, we compare the optimal solutions for policies obtained by different alignment approaches. %\textcolor{red}%{Our main finding is that for either $|\mathcal{D}|\gg |\mathcal{P}|$ or $|\mathcal{P}|\gg |\mathcal{D}|$ situations, AIHF policy consistently ...}. 
Due to space limitation, all derivation in this section is relegated to Appendix \ref{app:why}.

\noindent {\bf Policy with Demonstration Data.} It can be easily shown that when only the demonstration data $\mathcal{D}$ is available, the probability of generating $i$-th data equals its empirical probability, i.e., $\pi_{\rm SFT}(\tau_i)=\frac{\#\{\tau_i~ \mbox{in}~ \mathcal{D}\}}{|\mathcal{D}|}$.
Assume that such a policy is parameterized by an {\it implicit} reward function $R_D$, using the following softmax choice model where $\tau_i \in A$ is selected with probability $
\pi^*_i(R)=\frac{\exp (R_i/\beta)}{\sum_{j=1}^N\exp (R_j/\beta)}$ where $R_i:=R(\tau_i)$. Assuming a reference value $\widehat{R}_{\mathcal{D}}(\tau_1)=\bar{R}_1$, then %according to \citet[Proposition 1]{hotz1993conditional}, one can solve the following system of equations to obtain the optimal rewards:
the optimal rewards satisfies (See Sec. \ref{app:rlhf}):
\begin{align}
\frac{\#\{\tau_i \in \mathcal{D}\}}{|\mathcal{D}|} = \pi^*_i(\widehat{R}_{\mathcal{D}})=\frac{\exp (\widehat{R}_{\mathcal{D}}(\tau_i)/\beta)}{\sum_{j=1}^N\exp (\widehat{R}_{\mathcal{D}}(\tau_j)/\beta)} ~~~~i\in \{2,\dots,N\}
\label{eq:SFT_estimator}.
\end{align}
This implicit reward will be used shortly to characterize the RLHF policy.

\noindent {\bf Policy with Preference Only Data.}  %{\red[for each of the cases, write down the policy expression][move as much as derivation to appendix as possible][make this section concise and readable.]} 
Next, it can be shown that when only the preference data $\mathcal{P}$ is available, the reward estimation problem is defined as:
\begin{align}
\widehat{R}_{\mathcal{P}}= \arg \max_R \ell_{RM}(R):=\mathbb{E}_{(\tau_i \succ \tau_j) \sim \mathcal{P}}\Big[\log \frac{\pi^*_i(R)}{\pi^*_i(R) + \pi^*_j(R)}\Big].
\label{eq:likelihood:preference}
\end{align}
Again with a fixed reference value $\widehat{R}_{\mathcal{P}}(\tau_1)=\bar{R}_1$, the solution is (see Sec. \ref{app:rlhf}): 
\begin{align}
\pi^*_i(\widehat{R}_{\mathcal{P}}) &=\frac{
\sum_{j:j \neq i}|\mathcal{P}_{i\succ j}| }
{
\sum_{j:j\neq i}|\mathcal{P}_{i,j}|
\rho_{-(i,j)}(\pi^*(\widehat{R}_{\mathcal{P}}))}
\label{eq:estimator_preferences}
\end{align}
where $|\mathcal{P}_{i\succ j}|:=\# \{\tau_i \succ \tau_j ~\mbox{in}~ \mathcal{P}\}$ and $|\mathcal{P}_{i,j}|:=|\mathcal{P}_{i\succ j}|+|\mathcal{P}_{j\succ i}|$ %and $\rho_{-i}(\pi):=\sum_{j \neq i}\rho_{-(i,j)}(\pi)$ 
and $\rho_{-(i,j)}(\pi):=\Big(1-\sum_{k\in A \backslash \{i,j\}}\pi_k\Big)^{-1}$ is the expected number of times an action {\em other} than $\tau_i$ or $\tau_j$ is selected when sampling actions from $\pi$ infinitely many times.

\noindent {\bf RLHF Policy.} Based on the above results, in Sec. \ref{appendix:aihf} we show that the RLHF approach has the optimal policy $\pi^{\rm RLHF}(\tau_i)=\pi^*_i\Big(\widehat{R}_{\mathcal{D}}
+\widehat{R}_{\mathcal{P}}\Big)$. {That is, the RLHF policy can be seen as the softmax policy for the {\em sum} of reward estimators obtained from demonstrations and preferences separately.}

\noindent {\bf AIHF Policy.} Finally, we also find that the AIHF policy is of the form: % (see Sec. \ref{appendix:aihf}):
\begin{equation}\label{eq:AIHF_estimator}
    \pi^{\rm AIHF}(\tau_i) = \frac{\# \{\tau_i ~\mbox{in}~ \mathcal{D}\} + \sum_{j \neq i}|\mathcal{P}_{i \succ j}|} {|\mathcal{D}|+\sum_{j \neq i} |\mathcal{P}_{i,j}|\rho_{-(i,j)}\big(\pi^*(\widehat{R}^{\rm AIHF})\big)}. 
\end{equation}

{\bf Discussion.} Let us summarize our findings. First, $\pi^{\rm RLHF}$, which takes the form of the softmax of the {\it sum} of two rewards—one learned during the SFT stage and another during the reward training stage—offers some interesting insights into this popular approach. Second, $\pi^{\rm AIHF}$ is more robust than $\pi^{\rm RLHF}$. To see this, suppose that $|\mathcal{D}|\gg|\mathcal{P}|$, i.e. there is more demonstration than preference data. In this case, the policy estimator in \eqref{eq:AIHF_estimator}  will be largely defined by the demonstration data (which is reasonable) whereas the RLHF policy can be noisy since it (soft) maximizes %{\red[why maximize? it is a function of the two?]} 
the sum of two reward estimators: one that is more accurate (i.e. the one based on demonstrations, $\widehat{R}_{\mathcal{D}}$) and one that is less accurate (i.e. the one based on preferences $\widehat{R}_{\mathcal{P}}$). A similar observation can be made when $|\mathcal{D}|\ll|\mathcal{P}|$; see Sec. \ref{appendix:aihf} for more details. Third, $\pi^{\rm AIHF}$ can have less variance as compared with $\pi^{\rm RLHF}$. Indeed, $\pi^{\rm AIHF}$ takes a form of a weighted average of the policies estimated separately with demonstration and preference data, as by using (\ref{eq:SFT_estimator}) and (\ref{eq:estimator_preferences}), we can re-write (\ref{eq:AIHF_estimator}) as:
\begin{equation*}
\resizebox{\textwidth}{!}{
$
\pi^{*}_i(\widehat{R}^{\rm AIHF})=
\frac{|\mathcal{D}| } 
{|\mathcal{D}|+\sum_{j \neq i} |\mathcal{P}_{i,j}|
\rho_{-(i,j)}\big(\pi^*(\widehat{R}^{\rm AIHF})\big)}\pi^{*}_i(\widehat{R}_{\mathcal{D}})+
\frac{ \sum_{j \neq i} |\mathcal{P}_{i,j}|
\rho_{-(i,j)}\big(\pi^*(\widehat{R}_{\mathcal{P}})\big)} 
{|\mathcal{D}|+\sum_{j \neq i} |\mathcal{P}_{i,j}|
\rho_{-(i,j)}\big(\pi^*(\widehat{R}^{\rm AIHF})\big)}\pi^{*}_i(\widehat{R}_{\mathcal{P}})
$
}
\end{equation*}
Such averaging entails reduced variance. We also include simple numerical examples in the Appendix \ref{sec:numerical_example} to further illustrate this point.

\section{Proposed Algorithm for \aname{}  Training} \label{sec:algorithm_training} 
We are now ready to design algorithms for the proposed \aname~formulation \eqref{eq:AFHF}.
To begin with, first note that \eqref{eq:AFHF} takes a hierarchical form, and it belongs to the class of problem named {\it bi-level} optimization, first developed in the 70s {\citep{fiacco1990nonlinear}}, and recently found many applications in machine learning {\citep{wang2021bi,liu2021investigating,liu2022general}}. Generically speaking, bi-level problems are not easy to optimize; more specifically, in \eqref{eq:AFHF}, the upper-level problem \eqref{eq:upper} is a function of {\it both} the lower-level optimal solution $\pi_{\theta}$ and the true parameter $\theta$. It follows that a (stochastic) first-order algorithm for $L(\theta)$ involves some (potentially non-trivial) implicit gradient computation which often involves computing the Hessian matrix for the lower-level objective function. Fortunately, as we will show shortly, with some special choices of $L_1$, $L_2$, and $L_3$, one can design some simple and very efficient algorithms. 

{Before we go to details, we note that throughout this section, we assume that we are searching for a good policy $\pi_{\theta}$ {\it and} a reward estimate $r(\cdot, \cdot;\theta)$ to align with human feedback, where the policy $\pi_{\theta}$ is an optimal solution w.r.t. the certain reward estimate $r(\cdot, \cdot;\theta)$ according to the policy optimization problem \eqref{eq:lower}. Due to such optimal policy constraint w.r.t. one explicit reward estimate, we design an algorithm to solve such a single-stage, bi-level problem which is different from DPO \citep{rafailov2023direct} that simply optimizes the fixed loss function \eqref{eq:AFHF_redution_dpo} directly.}

On a high level, the proposed algorithm alternates between a policy alignment step (which updates $\pi$ with a fixed reward $r(\cdot, \cdot;\theta)$), and a reward alignment step (which updates  $\theta$ using a stochastic gradient, a function of the demonstration and preference data). Next, we study these steps in detail.

\textbf{Policy Alignment Step.} One can adopt the standard approaches such as the well-known proximal policy optimization (PPO) \citep{schulman2017proximal} algorithm to obtain an approximate optimal policy that solves \eqref{eq:L3}. It is worth noting that, when considering  $T=1$, our discussion leading to \eqref{eq:dpo:pi} indicates the optimal policy takes a much simpler form. {In this case, it is possible to consider a simpler method than running PPO to obtain the optimal policy. One alternative way is to use a baseline estimated reward value to perform variance reduction \citep{li2023remax}, thus reducing the computational complexity.

It is important to note that, the point of the above discussion is that these different choices for solving the policy alignment problem can be incorporated into our overall approach. 

%{\red[more details.][if we adopt this, do we have a new algorithm?]}

\textbf{Reward Alignment.} In this step, we use a stochastic gradient-type algorithm to optimize $L(\theta)$. Towards this end, first, observe that 
	\begin{align}\label{eq:reward_gradient}
		\nabla  L(\theta) &=  w_1 \nabla L_1(\pi_\theta) +  \nabla L_2(\theta).
	\end{align}
Clearly, regardless of the choice of $L_2$, $\nabla L_2$ is relatively easy to compute because the objective is directly related to $\theta$ since $L_2(\theta)$ can be regarded as one supervised learning loss and do not involve the optimal policy $\pi_{\theta}$. In particular, we have the following expressions:
{\small
\begin{subequations}
    \begin{align}
    \nabla {L}^{\rm BTL}_2(\theta) & = \mathbb{E}_{(\tau_w \succ \tau_l)\sim \pi^P} \Big[ \nabla_\theta \log \Big( \sigma \big( R(\tau_w; \theta) - R(\tau_l;\theta) \big) \Big) \Big]. \label{eq:l2:2}
    %\nabla {L}^{\rm LLR}_2(\theta) & =\mathbb{E}_{(\tau_i\prec\tau_j)\sim \pi^P(\Xi)}  [ \nabla_{\theta} R(\tau_j; \theta) - \nabla_{\theta} R(\tau_i; \theta) ].\label{eq:l2:3}
    \end{align}
\end{subequations}}%

On the contrary, the computation of $\nabla L_1(\pi_{\theta})$ is more involved, since $L_1$ depends on $\theta$ {\it implicitly} through the corresponding optimal policy $\pi_{\theta}$. Fortunately, the following lemma indicates that this gradient has a simple and intuitive form as well, and the proof can be found in Appendix \ref{appendix:Lemma:gradient}.

\begin{lemma} 
    Suppose that $L_1$ takes the form of the objective \eqref{def:ml} for reward learning from demonstrations, and suppose that $L_3$ takes the form \eqref{eq:L3} with $c(\cdot)$ being the KL-divergence w.r.t. some initial policy $\pi^0$. Then we have the following expression:   
 {\small
\begin{align}
     \nabla_{\theta}  L_1(\pi_{\theta}) &= \mathbb{E}_{\tau \sim \pi^{\rm E}, \tau' \sim \pi_{\theta}}[\nabla_{\theta} \big( R(\tau; \theta) - R(\tau';\theta) \big) ] 
     \label{eq:term_IRL}
\end{align}
}
where $\pi_{\theta}$ is the optimal policy given the reward model parameterized by $\theta$, with the expression \eqref{eq:opt:policy}.
\label{Lemma:gradient}
\end{lemma}

Intuitively, if the current policy $\pi_{\theta}$ has not matched $\pi^{\rm E}$ yet, then the reward should be improved by going towards the direction suggested by the expert trajectories, while {\it going away} from those generated by the current policy. Similar to the BTL model, from the gradient expression \eqref{eq:term_IRL}, it is clear that the optimization is toward the direction of increasing the gap between the reward of the real samples (demonstrations) and the synthetic ones (model generated continuations).

In practice, a few approximations need to be made to obtain a stochastic gradient of $L_1$. First, similarly, as before, the precise expectation cannot be obtained because the ground truth policy  $\pi^{\rm E}$ is unknown. Denote an offline demonstration dataset as $\mathcal{D}^{\rm E}:=\{\tau\}$, then one can replace the expectations $\mathbb{E}_{\tau\sim\pi^E}$ by $\mathbb{E}_{\tau\sim \mathcal{D}^{\rm E}}$. Second, in the second expectation in \eqref{eq:term_IRL}, the trajectories $\tau'$ are sampled from $\pi_\theta$, the optimal policy for a fixed reward parameterization by $\theta$. This means that the {\it policy alignment} step has to identify the optimal policy $\pi_{\theta}$ first, which, due to limitations such as computational constraints, and non-linear parameterization, is generally not possible. Instead, we propose to sample from the {\it current} policy $\pi^{k+1}$ obtained from the previous policy optimization step, where index $k$ represents the iteration counter. Following the approximation steps mentioned above, we construct a stochastic estimator $g_k$ to approximate the exact gradient $\nabla L(\theta_k)$ in \eqref{eq:reward_gradient} as follows:
\begin{align} \label{eq:sto_gradient_expression}
    g_k := w_1 g^k_1 + g^k_2 :=& w_1\big(\nabla_\theta R(\tau^E_{k},\theta_{k}) -  \nabla_\theta R(\tau^A_{k},\theta_{k}) \big)+\big(1-\sigma(R(\tau^W_{k},\theta_{k}) 
    -  R(\tau^L_{k},\theta_{k}))\big) \nonumber \\ &\quad{\times} \big(\nabla_\theta R(\tau^W_{k},\theta_{k}) -  \nabla_\theta R(\tau^L_{k},\theta_{k})\big).
\end{align}
The above two steps are summarized in Algorithm  \ref{alg:offline_ML_IRL}.
let us remark on the computational complexity of the proposed algorithm. Note that our algorithm is motivated by a class of popular algorithms in bi-level optimization, where the upper-level and lower-level problems are updated alternatingly using stochastic optimization \citep{hong2023two}. We conclude the section by theoretically inspecting the proposed algorithms. 
{\begin{algorithm}[t] 
	\caption{{{\it \small Alignment with Integrated Human Feedback (\aname)}}}
 \small
	{\begin{algorithmic}			\STATE {\bfseries Input:} Initialize reward parameter $ \theta^{0} $ and policy model $ \pi^{0} $, the stepsize of reward update $\eta$. Let $\mathcal{P}$, $\mathcal{D^{\rm E}}$ denote the preference and the demonstration data, respectively.
        % \STATE {\bfseries Initialization} initialize reward function parameter $\theta_0$

	\FOR{ \textbf{Iteration} $k=0,1,\ldots, K-1$}
        \STATE \textbf{{Policy Alignment:}} Optimizing $L_3$ by RL subroutine, e.g. PPO, to obtain one improved policy $\pi^{k+1}$  
	\STATE \textbf{Data Sample I:} Sample an expert trajectory $\tau\sim \mathcal{D^{\rm E}}$ and agent trajectory from $\tau'\sim \pi^{k+1}$ 
        \STATE \textbf{Data Sample II:} Sample preference pair $(\tau_{w}\succ \tau_{l})\sim \mathcal{P}$
        \STATE \textbf{Estimating Gradient:} Calculate one gradient estimator $g^k := w_1 g_1^k + g_2^k$ of $\nabla_{\theta} L(\theta) = w_1 \nabla_{\theta} L_1(\theta) + \nabla_{\theta} L_2(\theta)$ 
	\STATE \textbf{Reward Alignment:} $ \theta^{k+1} := \theta^{k} + \eta g^k $
	\ENDFOR
 
	\end{algorithmic}}
\label{alg:offline_ML_IRL}
\end{algorithm}}

\begin{theorem}
\label{theorem:main_convergence_results}
	Suppose Assumptions \ref{Assumption:Ergodicity_Markov_chain} - \ref{Assumption:reward_grad_bound} hold. Selecting stepsize $\alpha := \frac{\alpha_0}{K^\sigma} $ for the reward update step  \eqref{eq:sto_gradient_expression} where $\alpha_0>0$ and $ \sigma\in(0,1) $ are some fixed constants, and $K$ is the total number of iterations to be run by the algorithm. Then the following result holds:
	\begin{subequations}
	    \begin{align}
		&\frac{1}{K}\sum_{k = 0}^{K-1} \mathbb{E} \left[ \big \|  \log\pi_{k+1} - \log\pi_{\theta_k} \big \|_{\infty} \right] = \mathcal{O}(K^{-1}) + \mathcal{O}(K^{-\sigma}) \label{rate:lower_error} \\
		&\frac{1}{K} \sum_{k = 0}^{K - 1} \mathbb{E} \left[ \|  \nabla L(\theta_{k})  \|^2 \right]  = \mathcal{O}(K^{-\sigma}) + \mathcal{O}(K^{-1 + \sigma}) + \mathcal{O}(K^{-1}) \label{rate:upper_grad_norm}
	\end{align}
	\end{subequations}
	where $\|  \log\pi_{k+1} - \log\pi_{\theta_k} \|_{\infty} := \max_{s \in \mathcal{S}, a \in \mathcal{A}} \big| \log\pi_{k+1}(a|s) - \log\pi_{\theta_k}(a|s) \big|$. In particular, setting $\sigma=1/2$, then both quantities in \eqref{rate:lower_error} and \eqref{rate:upper_grad_norm} converge with the rate $\mathcal{O}(K^{-1/2}).$ 
\end{theorem}

The above theorem shows that Alg. \ref{alg:offline_ML_IRL} could converge to the stationary point if we take a large loop number $K$. Note that details and proofs of the result above are delegated to Appendix \ref{sec:convergence_result}.

\section{Experiments}\label{sec:experiments}
In this section, we provide numerical evaluations of the proposed method \eqref{eq:AFHF} (Alg. \ref{alg:offline_ML_IRL}) and its variants \eqref{eq:AFHF_redution_dpo} and \eqref{eq:AIHF_SPIN}, and comparing them with state-of-the-art methods RLHF \citep{ouyang2022training}, DPO \citep{rafailov2023direct}, IPO \citep{calandriello2024human} and SPIN \citep{chen2024self}. Our experiments demonstrate the advantages of the proposed methods in the following aspects: (1) Reward learning from demonstration and preference is the key to improving over standard RLHF. (2) Using demonstration in reward learning could increase model improvement efficiency (w.r.t. the KL divergence violation) (3) AIHF could reduce the effect of distribution mismatch caused by the sequential alignment method, thus breaking the performance limits of the state-of-the-art methods.

\noindent{\bf Models and datasets.} In the first setting, we test Alg. \ref{alg:offline_ML_IRL} on Anthropic-HH \citep{bai2022training} dataset\footnote{Dataset available at \url{https://huggingface.co/datasets/Anthropic/hh-rlhf}.} with (relatively small) Pythia \citep{biderman2023pythia} models\footnote{Models available at \url{https://huggingface.co/EleutherAI}.} as policy models. Anthropic-HH is a preference dataset collected from 52B LLMs that provide two continuations based on helpfulness and harmlessness, and we pick 10k chosen/preferred continuation data to form the demonstration dataset, while others serve as preference dataset and RL prompt dataset. For the HH dataset, We first fine-tune the language models (Pythia-160M/1B/2.8B) through supervised fine-tuning over all chosen responses from the HH dataset for 1 epoch, we call it {\it full-SFT model} and use it as our base model. Moreover, we also SFT the language model using the selected top 10k chosen responses and name it as {\it demonstration-SFT model}. For each policy model, we use the exact same model Pythia-1.4B as the reward model.

The other setting we test is on  7B models. 
% We also tested two variants of AIHF: Direct AIHF \eqref{eq:AFHF_redution_dpo}, and Self-Play AIHF \eqref{eq:AIHF_SPIN}. 
We use Ultrafeedback\footnote{Available at \url{https://huggingface.co/datasets/HuggingFaceH4/ultrafeedback_binarized}.} as our preference dataset (61.1k preference data) and Ultrachat200k\footnote{Available at \url{https://huggingface.co/datasets/HuggingFaceH4/ultrachat_200k}.} as the demonstration dataset (208k demonstration data), with mistral-7b-sft-beta \footnote{Available at \url{https://huggingface.co/HuggingFaceH4/mistral-7b-sft-beta}.} \citep{jiang2023mistral} as our base model. We use the same mistral-7b-sft-beta model as the initialization of the reward model.

\noindent{\bf Evaluation.} For the Anthropic-HH dataset, we present the reward evaluated by the PKU-Alignment/beaver-7b-v3.0-reward model\citep{ji2024beavertails}. In our 7B model experiments, we adopt the widely recognized HuggingFace Open LLM Leaderboard framework \citep{open-llm-leaderboard}. This evaluation suite measures LLM performance across six tasks: commonsense reasoning (Arc \citep{clark2018think}, HellaSwag \citep{zellers2019hellaswag}, Winogrande \citep{sakaguchi2021winogrande}), multi-task language understanding (MMLU \citep{hendrycks2020measuring}), mimicking human falsehoods (TruthfulQA \citep{lin2021truthfulqa}), and math problem-solving (GSM8K \citep{cobbe2021training}). Additional implementation details can be found in Appendix \ref{app:sec:experiments}.

\begin{figure*}[t]
    \begin{center}
    \subfigure[Pythia-160M]{\includegraphics[clip, trim={0cm 0cm 0cm 0.7cm}, width=0.32\textwidth]{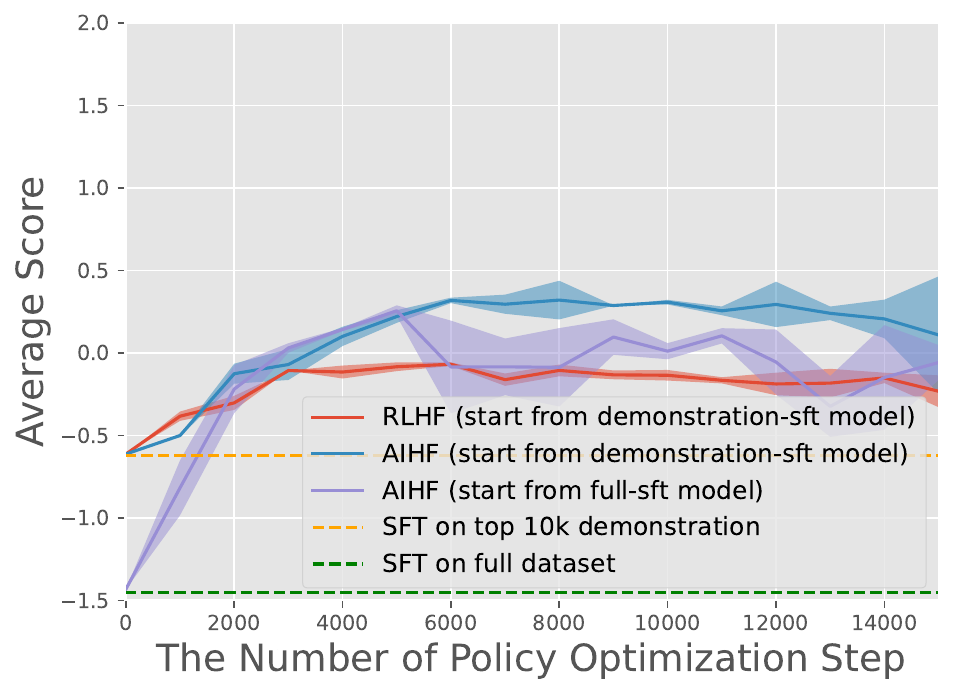}\label{fig:160M}}
    \subfigure[Pythia-1B]{\includegraphics[clip, trim={0cm 0cm 0cm 0.7cm}, width=0.32\textwidth]{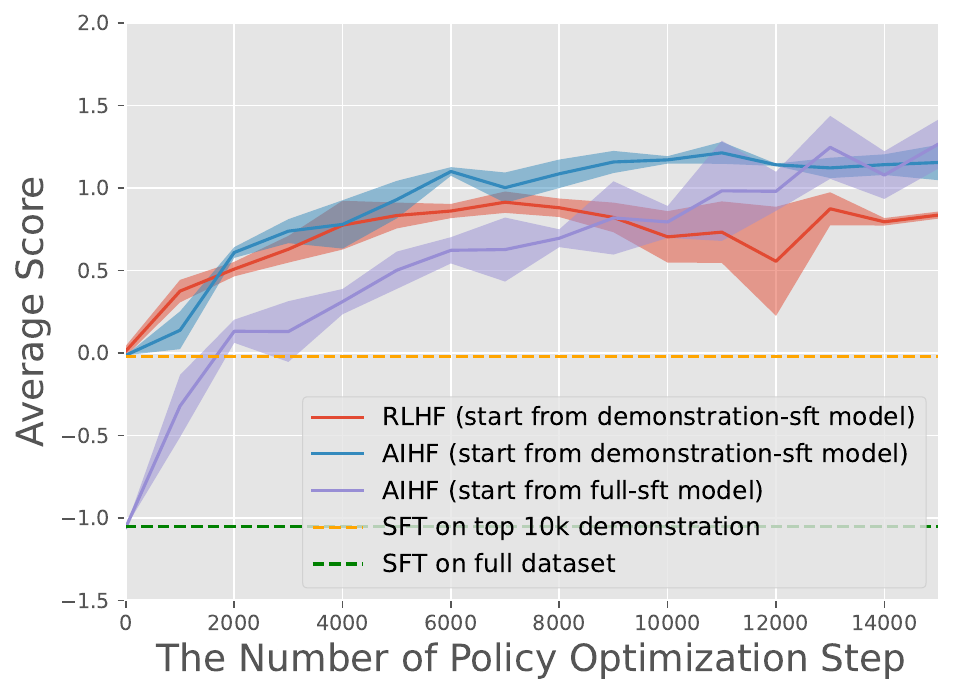}\label{fig:1B}}
    \subfigure[Pythia-2.8B]{\includegraphics[clip, trim={0cm 0cm 0cm 0.7cm}, width=0.32\textwidth]{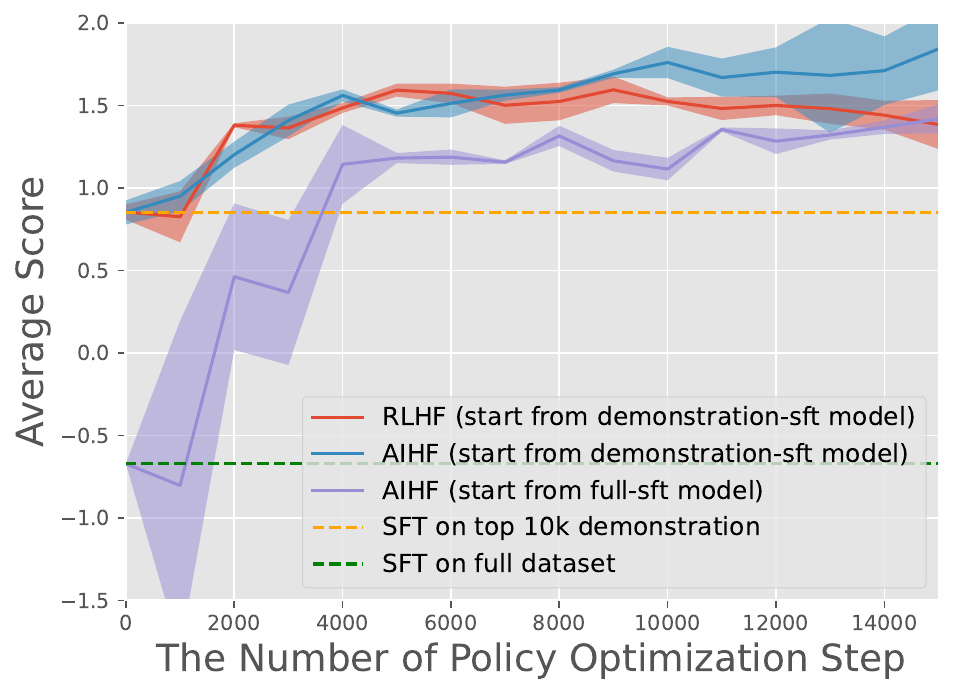}\label{fig:2.8B}}
    
    \caption{Experiment results of Pythia-160M/1B/2.8B policy models, with the reward model trained from Pythia-1.4B. We record the average scores (across three trials) of AIHF and RLHF on the Anthropic-HH test dataset (See Tab. \ref{tab:policy_quality} in Appendix for more comparisons with other algorithms)}.
    
    \label{fig:Reward Evaluation HH}
    \end{center}
\end{figure*}

\noindent \textbf{Results of small model (160M, 1B, and 2.8B) experiments.} We observe that the proposed AIHF performs effectively when initiated from both the demonstration-SFT model and the full-SFT model. %{\color{red}[what's the difference? is it defined?]}. 
As shown in Fig.\ref{fig:Reward Evaluation HH}, utilizing the same data, \aname~ algorithm can eventually outperform RLHF irrespective of the initial model. Furthermore, according to the numerical results as shown in Fig.\ref{fig:KL}, we see that the proposed AIHF algorithm has a smaller deviation from the base model compared with the RLHF algorithm. This benefit of the AIHF approach is due to the fact that we incorporate the maximum likelihood IRL objective for both reward learning and policy learning. In this case, both the reward model and policy model will be trained to align with the demonstrations, which are also used in the training process of the SFT stage. %{\color{red}[do these explained the reason that we have smaller distance well?]}. 
We also conducted a study on the demonstration/preference data ratio in Fig. \ref{fig:1B_ratio1} and \ref{fig:1B_ratio2}. We observe that AIHF consistently outperforms RLHF across different demonstration/preference data ratios. Furthermore, we also record the performance of the two variants, namely Direct AIHF \eqref{eq:AFHF_redution_dpo} and Self-Play-AIHF \eqref{eq:AIHF_SPIN}, also RLHF, DPO, IPO, and SPIN in Tab. \ref{tab:policy_quality}. Our proposed AIHF still outperforms all these methods in this experiment setting.

\begin{figure*}[t]
    \begin{center}
    \subfigure[KL divergence to demonstration-SFT policy]{\includegraphics[clip, trim={0cm 0cm 0cm 1cm}, width=0.34\textwidth]{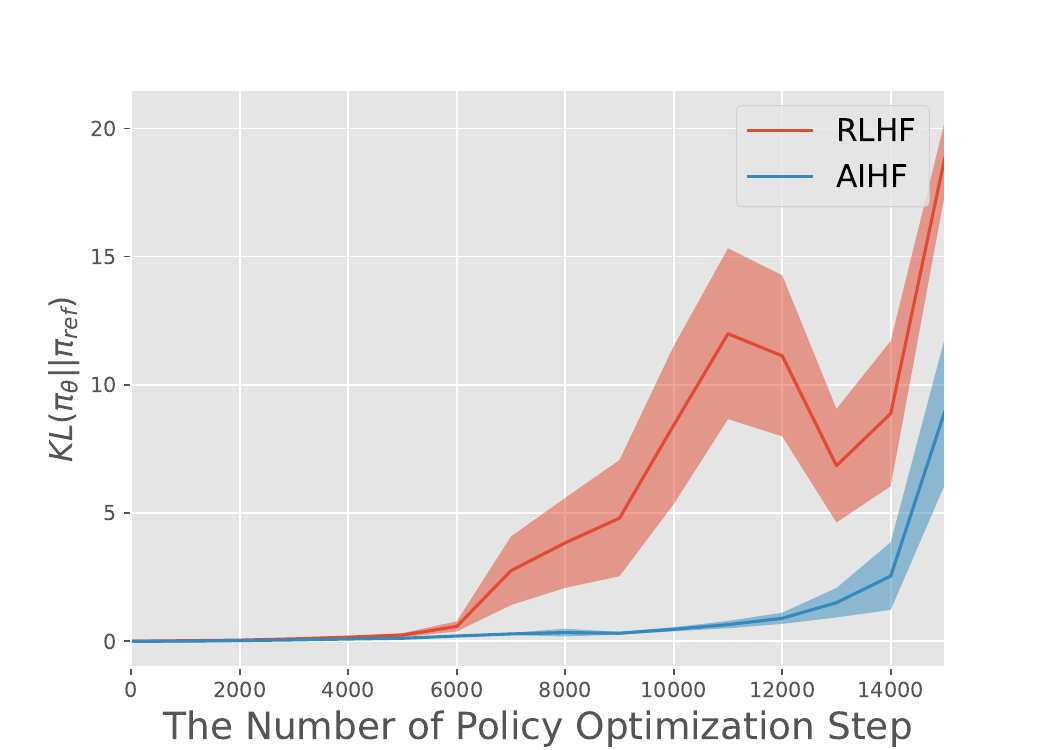}\label{fig:KL}}
    \subfigure[AIHF vs RLHF with 10k demonstration, 5k preference]{\includegraphics[clip, trim={0cm 0cm 0cm 1cm}, width=0.32\textwidth]{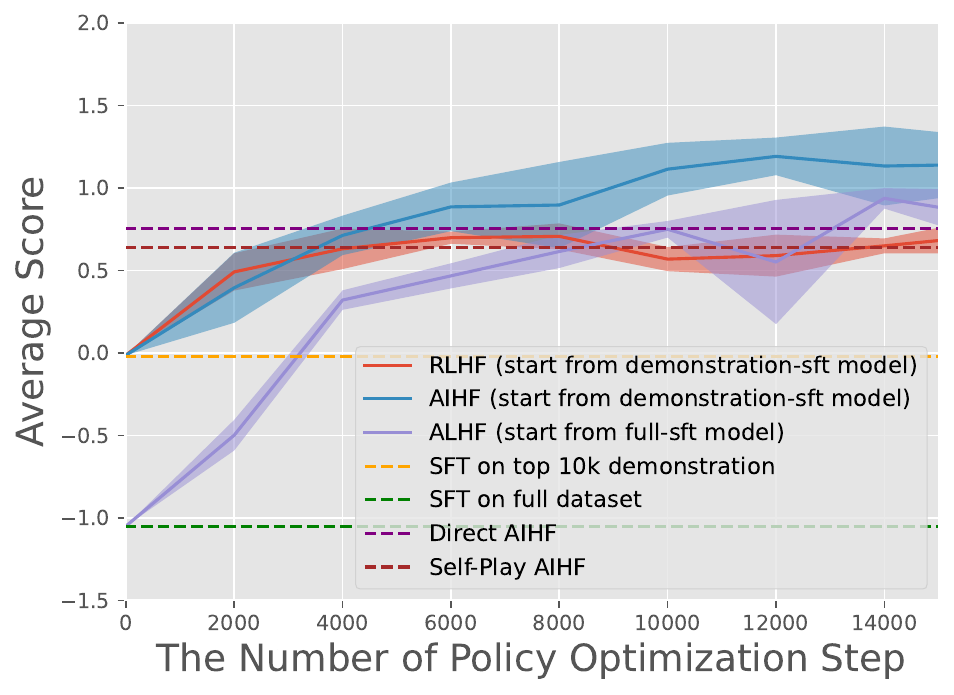}}\label{fig:1B_ratio1}
    \subfigure[AIHF vs RLHF with 10k demonstration, 10k preference]{\includegraphics[clip, trim={0cm 0cm 0cm 1cm}, width=0.32\textwidth]{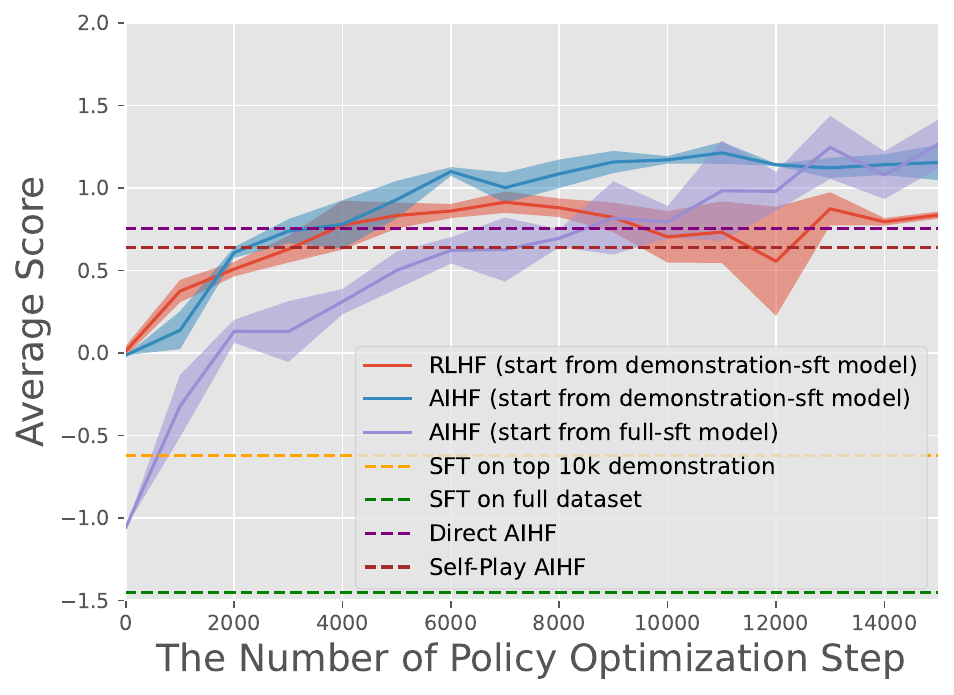}\label{fig:1B_ratio2}}
    \caption{Experiment results on Pythia-1B policy models, where the reward model is trained from Pythia-1.4B models. We record the average scores of AIHF and RLHF on the Anthropic-HH test dataset, reporting the results across three different trials.}
    \label{fig:1B_evaluation}
    \end{center}
\end{figure*}

\noindent \textbf{Results of large model (7B) experiments.} 
We run AIHF, Direct AIHF, and Self-Play AIHF along with other methods on the 7B experiment setting. The results are presented in Fig. \ref{fig:7B_downstream} (the numbers are recorded in Tab. \ref{tab:leaderboard}), where we can see that similar to the 1B setting, AIHF is consistently outperforming other methods. Additionally, both Direct AIHF and Self-Play AIHF effectively outperform the RLHF model (zephyr-7b-beta). The success of AIHF, as well as Self-Play AIHF and Direct AIHF further suggests that joint learning from demonstration and preference is indeed beneficial for the alignment. We also conduct an ablation study with different choices of $w_1$ in \eqref{eq:AFHF_redution_dpo}, as shown in Tab. \ref{tab:AIHF_weight_ablation},  the improvement of joint learning methods over baseline is robust. Furthermore, We also evaluate the reward models estimated using different methods (DPO, standard preference learning, and AIHF) over the widely used RewardBench \citep{lambert2024rewardbench}. The results, illustrated in Tab. \ref{tab:reward_bench} in the Appendix, show that the reward model trained through the AIHF can achieve significant improvement (especially on reasoning tasks) compared to both the standard BTL reward model in \eqref{eq:preference_gap_maximization} and the implicit reward model in DPO.

\begin{figure}[t]
    \centering
    \includegraphics[clip,width=0.75\linewidth]{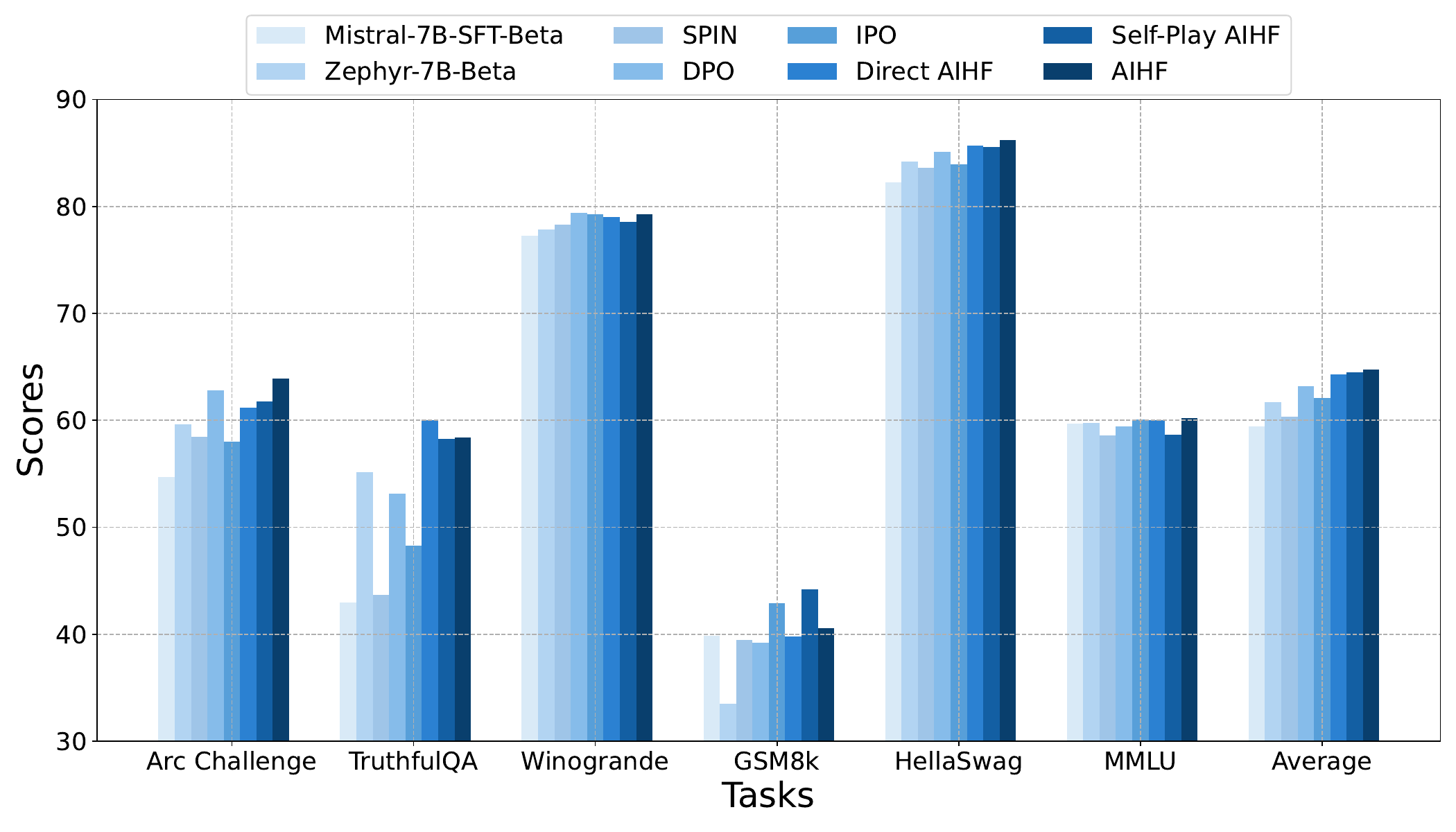}
    \caption{Performance comparison between Direct AIHF, Self-Play AIHF training across the six benchmark datasets (See also Table \ref{tab:leaderboard} in the Appendix).}
    \label{fig:7B_downstream}
\end{figure}

\noindent \textbf{Other Results.} Due to the page limits, we leave two additional experiments in the appendix: 1) movie review generation with positive sentiment on IMDb dataset \citep{maas2011learning}, 2) experiment on Robotics control tasks in MuJoCo \citep{todorov2012mujoco}. For the result of MuJoCo Experiment \ref{mujuco}, we observe that even though Behavior Cloning (BC)/SFT could provide a high-performing initialization, RLHF still fails to improve policy quality in the following RL stage. In the contract, AIHF can effectively integrate preferences and demonstrations, leading to a more robust reward function and consequently, a high-quality policy. For the IMDB result (Fig. \ref{fig:Mismatch-label}), We show that AIHF is able to alleviate the distribution mismatch between the generated trajectories by the policy, and the data that the learned reward model is able to rank.

\section{Conclusion}\label{app:sec:conclusion}
In this work, we study the alignment problem when diverse data sources from human feedback are available. Furthermore, we have developed an algorithmic framework that can integrate both expert demonstration and pairwise comparison data from human feedback to learn the reward functions for further guiding policy learning/model fine-tuning in the alignment pipeline. Through extensive evaluations on robotic control tasks and large language model alignment tasks, we demonstrate that our proposed method can outperform existing benchmarks on alignment tasks and is able to recover a better reward model to guide policy learning.

\newpage
\bibliography{iclr2025_conference}

\begin{thebibliography}{72}
\providecommand{\natexlab}[1]{#1}
\providecommand{\url}[1]{\texttt{#1}}
\expandafter\ifx\csname urlstyle\endcsname\relax
  \providecommand{\doi}[1]{doi: #1}\else
  \providecommand{\doi}{doi: \begingroup \urlstyle{rm}\Url}\fi

\bibitem[Arora \& Doshi(2021)Arora and Doshi]{arora2021survey}
Saurabh Arora and Prashant Doshi.
\newblock A survey of inverse reinforcement learning: Challenges, methods and
  progress.
\newblock \emph{Artificial Intelligence}, 297:\penalty0 103500, 2021.

\bibitem[Azar et~al.(2023)Azar, Rowland, Piot, Guo, Calandriello, Valko, and
  Munos]{azar2023general}
Mohammad~Gheshlaghi Azar, Mark Rowland, Bilal Piot, Daniel Guo, Daniele
  Calandriello, Michal Valko, and R{\'e}mi Munos.
\newblock A general theoretical paradigm to understand learning from human
  preferences.
\newblock \emph{arXiv preprint arXiv:2310.12036}, 2023.

\bibitem[Bai et~al.(2022)Bai, Jones, Ndousse, Askell, Chen, DasSarma, Drain,
  Fort, Ganguli, Henighan, et~al.]{bai2022training}
Yuntao Bai, Andy Jones, Kamal Ndousse, Amanda Askell, Anna Chen, Nova DasSarma,
  Dawn Drain, Stanislav Fort, Deep Ganguli, Tom Henighan, et~al.
\newblock Training a helpful and harmless assistant with reinforcement learning
  from human feedback.
\newblock \emph{arXiv preprint arXiv:2204.05862}, 2022.

\bibitem[Beeching et~al.(2023)Beeching, Fourrier, Habib, Han, Lambert, Rajani,
  Sanseviero, Tunstall, and Wolf]{open-llm-leaderboard}
Edward Beeching, Clémentine Fourrier, Nathan Habib, Sheon Han, Nathan Lambert,
  Nazneen Rajani, Omar Sanseviero, Lewis Tunstall, and Thomas Wolf.
\newblock Open llm leaderboard.
\newblock
  \url{https://huggingface.co/spaces/HuggingFaceH4/open_llm_leaderboard}, 2023.

\bibitem[Bellemare et~al.(2020)Bellemare, Candido, Castro, Gong, Machado,
  Moitra, Ponda, and Wang]{bellemare2020autonomous}
Marc~G Bellemare, Salvatore Candido, Pablo~Samuel Castro, Jun Gong, Marlos~C
  Machado, Subhodeep Moitra, Sameera~S Ponda, and Ziyu Wang.
\newblock Autonomous navigation of stratospheric balloons using reinforcement
  learning.
\newblock \emph{Nature}, 588\penalty0 (7836):\penalty0 77--82, 2020.

\bibitem[Berner et~al.(2019)Berner, Brockman, Chan, Cheung, D{\k{e}}biak,
  Dennison, Farhi, Fischer, Hashme, Hesse, et~al.]{berner2019dota}
Christopher Berner, Greg Brockman, Brooke Chan, Vicki Cheung, Przemys{\l}aw
  D{\k{e}}biak, Christy Dennison, David Farhi, Quirin Fischer, Shariq Hashme,
  Chris Hesse, et~al.
\newblock Dota 2 with large scale deep reinforcement learning.
\newblock \emph{arXiv preprint arXiv:1912.06680}, 2019.

\bibitem[Biderman et~al.(2023)Biderman, Schoelkopf, Anthony, Bradley,
  O’Brien, Hallahan, Khan, Purohit, Prashanth, Raff,
  et~al.]{biderman2023pythia}
Stella Biderman, Hailey Schoelkopf, Quentin~Gregory Anthony, Herbie Bradley,
  Kyle O’Brien, Eric Hallahan, Mohammad~Aflah Khan, Shivanshu Purohit,
  USVSN~Sai Prashanth, Edward Raff, et~al.
\newblock Pythia: A suite for analyzing large language models across training
  and scaling.
\newblock In \emph{International Conference on Machine Learning}, pp.\
  2397--2430. PMLR, 2023.

\bibitem[B{\i}y{\i}k et~al.(2022)B{\i}y{\i}k, Losey, Palan, Landolfi, Shevchuk,
  and Sadigh]{biyik2022learning}
Erdem B{\i}y{\i}k, Dylan~P Losey, Malayandi Palan, Nicholas~C Landolfi, Gleb
  Shevchuk, and Dorsa Sadigh.
\newblock Learning reward functions from diverse sources of human feedback:
  Optimally integrating demonstrations and preferences.
\newblock \emph{The International Journal of Robotics Research}, 41\penalty0
  (1):\penalty0 45--67, 2022.

\bibitem[Blavatskyy \& Pogrebna(2010)Blavatskyy and
  Pogrebna]{blavatskyy2010models}
Pavlo~R Blavatskyy and Ganna Pogrebna.
\newblock Models of stochastic choice and decision theories: Why both are
  important for analyzing decisions.
\newblock \emph{Journal of Applied Econometrics}, 25\penalty0 (6):\penalty0
  963--986, 2010.

\bibitem[Bloem \& Bambos(2014)Bloem and Bambos]{bloem2014infinite}
Michael Bloem and Nicholas Bambos.
\newblock Infinite time horizon maximum causal entropy inverse reinforcement
  learning.
\newblock In \emph{53rd IEEE conference on decision and control}, pp.\
  4911--4916. IEEE, 2014.

\bibitem[Bradley \& Terry(1952)Bradley and Terry]{bradley1952rank}
Ralph~Allan Bradley and Milton~E Terry.
\newblock Rank analysis of incomplete block designs: I. the method of paired
  comparisons.
\newblock \emph{Biometrika}, 39\penalty0 (3/4):\penalty0 324--345, 1952.

\bibitem[Brown et~al.(2019)Brown, Goo, Nagarajan, and
  Niekum]{brown2019extrapolating}
Daniel Brown, Wonjoon Goo, Prabhat Nagarajan, and Scott Niekum.
\newblock Extrapolating beyond suboptimal demonstrations via inverse
  reinforcement learning from observations.
\newblock In \emph{International conference on machine learning}, pp.\
  783--792. PMLR, 2019.

\bibitem[Calandriello et~al.(2024)Calandriello, Guo, Munos, Rowland, Tang,
  Pires, Richemond, Lan, Valko, Liu, et~al.]{calandriello2024human}
Daniele Calandriello, Daniel Guo, Remi Munos, Mark Rowland, Yunhao Tang,
  Bernardo~Avila Pires, Pierre~Harvey Richemond, Charline~Le Lan, Michal Valko,
  Tianqi Liu, et~al.
\newblock Human alignment of large language models through online preference
  optimisation.
\newblock \emph{arXiv preprint arXiv:2403.08635}, 2024.

\bibitem[Chen et~al.(2024)Chen, Deng, Yuan, Ji, and Gu]{chen2024self}
Zixiang Chen, Yihe Deng, Huizhuo Yuan, Kaixuan Ji, and Quanquan Gu.
\newblock Self-play fine-tuning converts weak language models to strong
  language models.
\newblock \emph{arXiv preprint arXiv:2401.01335}, 2024.

\bibitem[Clark et~al.(2018)Clark, Cowhey, Etzioni, Khot, Sabharwal, Schoenick,
  and Tafjord]{clark2018think}
Peter Clark, Isaac Cowhey, Oren Etzioni, Tushar Khot, Ashish Sabharwal, Carissa
  Schoenick, and Oyvind Tafjord.
\newblock Think you have solved question answering? try arc, the ai2 reasoning
  challenge.
\newblock \emph{arXiv preprint arXiv:1803.05457}, 2018.

\bibitem[Cobbe et~al.(2021)Cobbe, Kosaraju, Bavarian, Chen, Jun, Kaiser,
  Plappert, Tworek, Hilton, Nakano, et~al.]{cobbe2021training}
Karl Cobbe, Vineet Kosaraju, Mohammad Bavarian, Mark Chen, Heewoo Jun, Lukasz
  Kaiser, Matthias Plappert, Jerry Tworek, Jacob Hilton, Reiichiro Nakano,
  et~al.
\newblock Training verifiers to solve math word problems.
\newblock \emph{arXiv preprint arXiv:2110.14168}, 2021.

\bibitem[Daniel et~al.(2014)Daniel, Viering, Metz, Kroemer, and
  Peters]{daniel2014active}
Christian Daniel, Malte Viering, Jan Metz, Oliver Kroemer, and Jan Peters.
\newblock Active reward learning.
\newblock In \emph{Robotics: Science and systems}, volume~98, 2014.

\bibitem[Fiacco \& McCormick(1990)Fiacco and McCormick]{fiacco1990nonlinear}
Anthony~V Fiacco and Garth~P McCormick.
\newblock \emph{Nonlinear programming: sequential unconstrained minimization
  techniques}.
\newblock SIAM, 1990.

\bibitem[Fischer et~al.(2021)Fischer, Eyberg, Werling, and
  Lauer]{fischer2021sampling}
Johannes Fischer, Christoph Eyberg, Moritz Werling, and Martin Lauer.
\newblock Sampling-based inverse reinforcement learning algorithms with safety
  constraints.
\newblock In \emph{2021 IEEE/RSJ International Conference on Intelligent Robots
  and Systems (IROS)}, pp.\  791--798. IEEE, 2021.

\bibitem[Fudenberg et~al.(2015)Fudenberg, Iijima, and
  Strzalecki]{fudenberg2015stochastic}
Drew Fudenberg, Ryota Iijima, and Tomasz Strzalecki.
\newblock Stochastic choice and revealed perturbed utility.
\newblock \emph{Econometrica}, 83\penalty0 (6):\penalty0 2371--2409, 2015.

\bibitem[Haarnoja et~al.(2018)Haarnoja, Zhou, Abbeel, and
  Levine]{haarnoja2018soft}
Tuomas Haarnoja, Aurick Zhou, Pieter Abbeel, and Sergey Levine.
\newblock Soft actor-critic: Off-policy maximum entropy deep reinforcement
  learning with a stochastic actor.
\newblock In \emph{International conference on machine learning}, pp.\
  1861--1870. PMLR, 2018.

\bibitem[Hejna \& Sadigh(2023)Hejna and Sadigh]{hejna2023inverse}
Joey Hejna and Dorsa Sadigh.
\newblock Inverse preference learning: Preference-based rl without a reward
  function.
\newblock \emph{arXiv preprint arXiv:2305.15363}, 2023.

\bibitem[Hejna \& Sadigh(2024)Hejna and Sadigh]{hejna2024inverse}
Joey Hejna and Dorsa Sadigh.
\newblock Inverse preference learning: Preference-based rl without a reward
  function.
\newblock \emph{Advances in Neural Information Processing Systems}, 36, 2024.

\bibitem[Hendrycks et~al.(2020)Hendrycks, Burns, Basart, Zou, Mazeika, Song,
  and Steinhardt]{hendrycks2020measuring}
Dan Hendrycks, Collin Burns, Steven Basart, Andy Zou, Mantas Mazeika, Dawn
  Song, and Jacob Steinhardt.
\newblock Measuring massive multitask language understanding.
\newblock \emph{arXiv preprint arXiv:2009.03300}, 2020.

\bibitem[Hong et~al.(2024)Hong, Lee, and Thorne]{hong2024orpo}
Jiwoo Hong, Noah Lee, and James Thorne.
\newblock Orpo: Monolithic preference optimization without reference model.
\newblock \emph{arXiv preprint arXiv:2403.07691}, 2024.

\bibitem[Hong et~al.(2020)Hong, Wai, Wang, and Yang]{hong2020two}
Mingyi Hong, Hoi-To Wai, Zhaoran Wang, and Zhuoran Yang.
\newblock A two-timescale framework for bilevel optimization: Complexity
  analysis and application to actor-critic.
\newblock \emph{arXiv preprint arXiv:2007.05170}, 2020.

\bibitem[Hong et~al.(2023)Hong, Wai, Wang, and Yang]{hong2023two}
Mingyi Hong, Hoi-To Wai, Zhaoran Wang, and Zhuoran Yang.
\newblock A two-timescale stochastic algorithm framework for bilevel
  optimization: Complexity analysis and application to actor-critic.
\newblock \emph{SIAM Journal on Optimization}, 33\penalty0 (1):\penalty0
  147--180, 2023.

\bibitem[Ibarz et~al.(2018)Ibarz, Leike, Pohlen, Irving, Legg, and
  Amodei]{ibarz2018reward}
Borja Ibarz, Jan Leike, Tobias Pohlen, Geoffrey Irving, Shane Legg, and Dario
  Amodei.
\newblock Reward learning from human preferences and demonstrations in atari.
\newblock \emph{Advances in neural information processing systems}, 31, 2018.

\bibitem[Ji et~al.(2024)Ji, Liu, Dai, Pan, Zhang, Bian, Chen, Sun, Wang, and
  Yang]{ji2024beavertails}
Jiaming Ji, Mickel Liu, Josef Dai, Xuehai Pan, Chi Zhang, Ce~Bian, Boyuan Chen,
  Ruiyang Sun, Yizhou Wang, and Yaodong Yang.
\newblock Beavertails: Towards improved safety alignment of llm via a
  human-preference dataset.
\newblock \emph{Advances in Neural Information Processing Systems}, 36, 2024.

\bibitem[Ji et~al.(2021)Ji, Yang, and Liang]{ji2021bilevel}
Kaiyi Ji, Junjie Yang, and Yingbin Liang.
\newblock Bilevel optimization: Convergence analysis and enhanced design.
\newblock In \emph{International Conference on Machine Learning}, pp.\
  4882--4892. PMLR, 2021.

\bibitem[Jiang et~al.(2023)Jiang, Sablayrolles, Mensch, Bamford, Chaplot,
  Casas, Bressand, Lengyel, Lample, Saulnier, et~al.]{jiang2023mistral}
Albert~Q Jiang, Alexandre Sablayrolles, Arthur Mensch, Chris Bamford,
  Devendra~Singh Chaplot, Diego de~las Casas, Florian Bressand, Gianna Lengyel,
  Guillaume Lample, Lucile Saulnier, et~al.
\newblock Mistral 7b.
\newblock \emph{arXiv preprint arXiv:2310.06825}, 2023.

\bibitem[Kalashnikov et~al.(2018)Kalashnikov, Irpan, Pastor, Ibarz, Herzog,
  Jang, Quillen, Holly, Kalakrishnan, Vanhoucke,
  et~al.]{kalashnikov2018scalable}
Dmitry Kalashnikov, Alex Irpan, Peter Pastor, Julian Ibarz, Alexander Herzog,
  Eric Jang, Deirdre Quillen, Ethan Holly, Mrinal Kalakrishnan, Vincent
  Vanhoucke, et~al.
\newblock Scalable deep reinforcement learning for vision-based robotic
  manipulation.
\newblock In \emph{Conference on Robot Learning}, pp.\  651--673. PMLR, 2018.

\bibitem[Khanduri et~al.(2021)Khanduri, Zeng, Hong, Wai, Wang, and
  Yang]{khanduri2021near}
Prashant Khanduri, Siliang Zeng, Mingyi Hong, Hoi-To Wai, Zhaoran Wang, and
  Zhuoran Yang.
\newblock A near-optimal algorithm for stochastic bilevel optimization via
  double-momentum.
\newblock \emph{Advances in Neural Information Processing Systems}, 34, 2021.

\bibitem[Kober \& Peters(2008)Kober and Peters]{kober2008policy}
Jens Kober and Jan Peters.
\newblock Policy search for motor primitives in robotics.
\newblock \emph{Advances in neural information processing systems}, 21, 2008.

\bibitem[Kwon et~al.(2023)Kwon, Li, Zhuang, Sheng, Zheng, Yu, Gonzalez, Zhang,
  and Stoica]{kwon2023efficient}
Woosuk Kwon, Zhuohan Li, Siyuan Zhuang, Ying Sheng, Lianmin Zheng, Cody~Hao Yu,
  Joseph Gonzalez, Hao Zhang, and Ion Stoica.
\newblock Efficient memory management for large language model serving with
  pagedattention.
\newblock In \emph{Proceedings of the 29th Symposium on Operating Systems
  Principles}, pp.\  611--626, 2023.

\bibitem[Lambert et~al.(2024)Lambert, Pyatkin, Morrison, Miranda, Lin, Chandu,
  Dziri, Kumar, Zick, Choi, et~al.]{lambert2024rewardbench}
Nathan Lambert, Valentina Pyatkin, Jacob Morrison, LJ~Miranda, Bill~Yuchen Lin,
  Khyathi Chandu, Nouha Dziri, Sachin Kumar, Tom Zick, Yejin Choi, et~al.
\newblock Rewardbench: Evaluating reward models for language modeling.
\newblock \emph{arXiv preprint arXiv:2403.13787}, 2024.

\bibitem[Leike et~al.(2018)Leike, Krueger, Everitt, Martic, Maini, and
  Legg]{leike2018scalable}
Jan Leike, David Krueger, Tom Everitt, Miljan Martic, Vishal Maini, and Shane
  Legg.
\newblock Scalable agent alignment via reward modeling: a research direction.
\newblock \emph{arXiv preprint arXiv:1811.07871}, 2018.

\bibitem[Levine et~al.(2011)Levine, Popovic, and Koltun]{levine2011nonlinear}
Sergey Levine, Zoran Popovic, and Vladlen Koltun.
\newblock Nonlinear inverse reinforcement learning with gaussian processes.
\newblock \emph{Advances in neural information processing systems}, 24, 2011.

\bibitem[Li et~al.(2023)Li, Xu, Zhang, Yu, Sun, and Luo]{li2023remax}
Ziniu Li, Tian Xu, Yushun Zhang, Yang Yu, Ruoyu Sun, and Zhi-Quan Luo.
\newblock Remax: A simple, effective, and efficient method for aligning large
  language models.
\newblock \emph{arXiv preprint arXiv:2310.10505}, 2023.

\bibitem[Lin et~al.(2021)Lin, Hilton, and Evans]{lin2021truthfulqa}
Stephanie Lin, Jacob Hilton, and Owain Evans.
\newblock Truthfulqa: Measuring how models mimic human falsehoods.
\newblock \emph{arXiv preprint arXiv:2109.07958}, 2021.

\bibitem[Liu et~al.(2021)Liu, Gao, Zhang, Meng, and Lin]{liu2021investigating}
Risheng Liu, Jiaxin Gao, Jin Zhang, Deyu Meng, and Zhouchen Lin.
\newblock Investigating bi-level optimization for learning and vision from a
  unified perspective: A survey and beyond.
\newblock \emph{IEEE Transactions on Pattern Analysis and Machine
  Intelligence}, 44\penalty0 (12):\penalty0 10045--10067, 2021.

\bibitem[Liu et~al.(2022)Liu, Mu, Yuan, Zeng, and Zhang]{liu2022general}
Risheng Liu, Pan Mu, Xiaoming Yuan, Shangzhi Zeng, and Jin Zhang.
\newblock A general descent aggregation framework for gradient-based bi-level
  optimization.
\newblock \emph{IEEE Transactions on Pattern Analysis and Machine
  Intelligence}, 45\penalty0 (1):\penalty0 38--57, 2022.

\bibitem[Liu et~al.(2023)Liu, Zhao, Joshi, Khalman, Saleh, Liu, and
  Liu]{liu2023statistical}
Tianqi Liu, Yao Zhao, Rishabh Joshi, Misha Khalman, Mohammad Saleh, Peter~J
  Liu, and Jialu Liu.
\newblock Statistical rejection sampling improves preference optimization.
\newblock \emph{arXiv preprint arXiv:2309.06657}, 2023.

\bibitem[Liu et~al.(2024)Liu, Lu, Zhang, Liu, Guo, Yang, Blanchet, and
  Wang]{liu2024provably}
Zhihan Liu, Miao Lu, Shenao Zhang, Boyi Liu, Hongyi Guo, Yingxiang Yang, Jose
  Blanchet, and Zhaoran Wang.
\newblock Provably mitigating overoptimization in rlhf: Your sft loss is
  implicitly an adversarial regularizer.
\newblock \emph{arXiv preprint arXiv:2405.16436}, 2024.

\bibitem[Maas et~al.(2011)Maas, Daly, Pham, Huang, Ng, and
  Potts]{maas2011learning}
Andrew Maas, Raymond~E Daly, Peter~T Pham, Dan Huang, Andrew~Y Ng, and
  Christopher Potts.
\newblock Learning word vectors for sentiment analysis.
\newblock In \emph{Proceedings of the 49th annual meeting of the association
  for computational linguistics: Human language technologies}, pp.\  142--150,
  2011.

\bibitem[Mnih et~al.(2015)Mnih, Kavukcuoglu, Silver, Rusu, Veness, Bellemare,
  Graves, Riedmiller, Fidjeland, Ostrovski, et~al.]{mnih2015human}
Volodymyr Mnih, Koray Kavukcuoglu, David Silver, Andrei~A Rusu, Joel Veness,
  Marc~G Bellemare, Alex Graves, Martin Riedmiller, Andreas~K Fidjeland, Georg
  Ostrovski, et~al.
\newblock Human-level control through deep reinforcement learning.
\newblock \emph{nature}, 518\penalty0 (7540):\penalty0 529--533, 2015.

\bibitem[Myers et~al.(2022)Myers, Biyik, Anari, and Sadigh]{myers2022learning}
Vivek Myers, Erdem Biyik, Nima Anari, and Dorsa Sadigh.
\newblock Learning multimodal rewards from rankings.
\newblock In \emph{Conference on Robot Learning}, pp.\  342--352. PMLR, 2022.

\bibitem[Ouyang et~al.(2022)Ouyang, Wu, Jiang, Almeida, Wainwright, Mishkin,
  Zhang, Agarwal, Slama, Ray, et~al.]{ouyang2022training}
Long Ouyang, Jeffrey Wu, Xu~Jiang, Diogo Almeida, Carroll Wainwright, Pamela
  Mishkin, Chong Zhang, Sandhini Agarwal, Katarina Slama, Alex Ray, et~al.
\newblock Training language models to follow instructions with human feedback.
\newblock \emph{Advances in Neural Information Processing Systems},
  35:\penalty0 27730--27744, 2022.

\bibitem[Palan et~al.(2019)Palan, Landolfi, Shevchuk, and
  Sadigh]{palan2019learning}
Malayandi Palan, Nicholas~C Landolfi, Gleb Shevchuk, and Dorsa Sadigh.
\newblock Learning reward functions by integrating human demonstrations and
  preferences.
\newblock \emph{arXiv preprint arXiv:1906.08928}, 2019.

\bibitem[Park et~al.(2023)Park, Goldstein, O'Gara, Chen, and
  Hendrycks]{park2023ai}
Peter~S Park, Simon Goldstein, Aidan O'Gara, Michael Chen, and Dan Hendrycks.
\newblock Ai deception: A survey of examples, risks, and potential solutions.
\newblock \emph{arXiv preprint arXiv:2308.14752}, 2023.

\bibitem[Perez et~al.(2022)Perez, Ringer, Luko{\v{s}}i{\=u}t{\.e}, Nguyen,
  Chen, Heiner, Pettit, Olsson, Kundu, Kadavath, et~al.]{perez2022discovering}
Ethan Perez, Sam Ringer, Kamil{\.e} Luko{\v{s}}i{\=u}t{\.e}, Karina Nguyen,
  Edwin Chen, Scott Heiner, Craig Pettit, Catherine Olsson, Sandipan Kundu,
  Saurav Kadavath, et~al.
\newblock Discovering language model behaviors with model-written evaluations.
\newblock \emph{arXiv preprint arXiv:2212.09251}, 2022.

\bibitem[Rafailov et~al.(2023)Rafailov, Sharma, Mitchell, Ermon, Manning, and
  Finn]{rafailov2023direct}
Rafael Rafailov, Archit Sharma, Eric Mitchell, Stefano Ermon, Christopher~D
  Manning, and Chelsea Finn.
\newblock Direct preference optimization: Your language model is secretly a
  reward model.
\newblock \emph{arXiv preprint arXiv:2305.18290}, 2023.

\bibitem[Rajbhandari et~al.(2020)Rajbhandari, Rasley, Ruwase, and
  He]{Rajbhandari2020}
Samyam Rajbhandari, Jeff Rasley, Olatunji Ruwase, and Yuxiong He.
\newblock Zero: Memory optimizations toward training trillion parameter models.
\newblock In \emph{SC20: International Conference for High Performance
  Computing, Networking, Storage and Analysis}, pp.\  1--16, 2020.
\newblock \doi{10.1109/SC41405.2020.00024}.

\bibitem[Ross \& Bagnell(2010)Ross and Bagnell]{ross2010efficient}
St{\'e}phane Ross and Drew Bagnell.
\newblock Efficient reductions for imitation learning.
\newblock In \emph{Proceedings of the thirteenth international conference on
  artificial intelligence and statistics}, pp.\  661--668. JMLR Workshop and
  Conference Proceedings, 2010.

\bibitem[Sakaguchi et~al.(2021)Sakaguchi, Bras, Bhagavatula, and
  Choi]{sakaguchi2021winogrande}
Keisuke Sakaguchi, Ronan~Le Bras, Chandra Bhagavatula, and Yejin Choi.
\newblock Winogrande: An adversarial winograd schema challenge at scale.
\newblock \emph{Communications of the ACM}, 64\penalty0 (9):\penalty0 99--106,
  2021.

\bibitem[Schulman et~al.(2017)Schulman, Wolski, Dhariwal, Radford, and
  Klimov]{schulman2017proximal}
John Schulman, Filip Wolski, Prafulla Dhariwal, Alec Radford, and Oleg Klimov.
\newblock Proximal policy optimization algorithms.
\newblock \emph{arXiv preprint arXiv:1707.06347}, 2017.

\bibitem[Stiennon et~al.(2020)Stiennon, Ouyang, Wu, Ziegler, Lowe, Voss,
  Radford, Amodei, and Christiano]{stiennon2020learning}
Nisan Stiennon, Long Ouyang, Jeffrey Wu, Daniel Ziegler, Ryan Lowe, Chelsea
  Voss, Alec Radford, Dario Amodei, and Paul~F Christiano.
\newblock Learning to summarize with human feedback.
\newblock \emph{Advances in Neural Information Processing Systems},
  33:\penalty0 3008--3021, 2020.

\bibitem[Todorov et~al.(2012)Todorov, Erez, and Tassa]{todorov2012mujoco}
Emanuel Todorov, Tom Erez, and Yuval Tassa.
\newblock Mujoco: A physics engine for model-based control.
\newblock In \emph{2012 IEEE/RSJ international conference on intelligent robots
  and systems}, pp.\  5026--5033. IEEE, 2012.

\bibitem[Uehara et~al.(2023)Uehara, Kallus, Lee, and Sun]{uehara2023refined}
Masatoshi Uehara, Nathan Kallus, Jason~D Lee, and Wen Sun.
\newblock Refined value-based offline rl under realizability and partial
  coverage.
\newblock \emph{arXiv preprint arXiv:2302.02392}, 2023.

\bibitem[Wang et~al.(2021)Wang, Hua, Liu, Zhang, Yan, Qi, Yang, Zhou, and
  Yang]{wang2021bi}
Runzhong Wang, Zhigang Hua, Gan Liu, Jiayi Zhang, Junchi Yan, Feng Qi, Shuang
  Yang, Jun Zhou, and Xiaokang Yang.
\newblock A bi-level framework for learning to solve combinatorial optimization
  on graphs.
\newblock \emph{Advances in Neural Information Processing Systems},
  34:\penalty0 21453--21466, 2021.

\bibitem[Wulfmeier et~al.(2015)Wulfmeier, Ondruska, and
  Posner]{wulfmeier2015maximum}
Markus Wulfmeier, Peter Ondruska, and Ingmar Posner.
\newblock Maximum entropy deep inverse reinforcement learning.
\newblock \emph{arXiv preprint arXiv:1507.04888}, 2015.

\bibitem[Xiong et~al.(2024)Xiong, Dong, Ye, Wang, Zhong, Ji, Jiang, and
  Zhang]{xiong2024iterative}
Wei Xiong, Hanze Dong, Chenlu Ye, Ziqi Wang, Han Zhong, Heng Ji, Nan Jiang, and
  Tong Zhang.
\newblock Iterative preference learning from human feedback: Bridging theory
  and practice for rlhf under kl-constraint.
\newblock In \emph{Forty-first International Conference on Machine Learning},
  2024.

\bibitem[Xu et~al.(2024)Xu, Fu, Gao, Ye, Liu, Mei, Wang, Yu, and Wu]{xu2024dpo}
Shusheng Xu, Wei Fu, Jiaxuan Gao, Wenjie Ye, Weilin Liu, Zhiyu Mei, Guangju
  Wang, Chao Yu, and Yi~Wu.
\newblock Is dpo superior to ppo for llm alignment? a comprehensive study.
\newblock \emph{arXiv preprint arXiv:2404.10719}, 2024.

\bibitem[Xu et~al.(2020)Xu, Wang, and Liang]{xu2020improving}
Tengyu Xu, Zhe Wang, and Yingbin Liang.
\newblock Improving sample complexity bounds for (natural) actor-critic
  algorithms.
\newblock \emph{Advances in Neural Information Processing Systems},
  33:\penalty0 4358--4369, 2020.

\bibitem[Yuan et~al.(2023)Yuan, Yuan, Tan, Wang, Huang, and
  Huang]{yuan2023rrhf}
Zheng Yuan, Hongyi Yuan, Chuanqi Tan, Wei Wang, Songfang Huang, and Fei Huang.
\newblock Rrhf: Rank responses to align language models with human feedback
  without tears.
\newblock \emph{arXiv preprint arXiv:2304.05302}, 2023.

\bibitem[Zellers et~al.(2019)Zellers, Holtzman, Bisk, Farhadi, and
  Choi]{zellers2019hellaswag}
Rowan Zellers, Ari Holtzman, Yonatan Bisk, Ali Farhadi, and Yejin Choi.
\newblock Hellaswag: Can a machine really finish your sentence?
\newblock \emph{arXiv preprint arXiv:1905.07830}, 2019.

\bibitem[Zeng et~al.(2022{\natexlab{a}})Zeng, Hong, and
  Garcia]{zeng2022structural}
Siliang Zeng, Mingyi Hong, and Alfredo Garcia.
\newblock Structural estimation of markov decision processes in
  high-dimensional state space with finite-time guarantees.
\newblock \emph{arXiv preprint arXiv:2210.01282}, 2022{\natexlab{a}}.

\bibitem[Zeng et~al.(2022{\natexlab{b}})Zeng, Li, Garcia, and
  Hong]{zeng2022maximum}
Siliang Zeng, Chenliang Li, Alfredo Garcia, and Mingyi Hong.
\newblock Maximum-likelihood inverse reinforcement learning with finite-time
  guarantees.
\newblock \emph{Advances in Neural Information Processing Systems},
  2022{\natexlab{b}}.

\bibitem[Zhou et~al.(2017)Zhou, Bloem, and Bambos]{zhou2017infinite}
Zhengyuan Zhou, Michael Bloem, and Nicholas Bambos.
\newblock Infinite time horizon maximum causal entropy inverse reinforcement
  learning.
\newblock \emph{IEEE Transactions on Automatic Control}, 63\penalty0
  (9):\penalty0 2787--2802, 2017.

\bibitem[Zhu et~al.(2023)Zhu, Sharma, Frujeri, Dong, Zhu, Jordan, and
  Jiao]{zhu2023fine}
Banghua Zhu, Hiteshi Sharma, Felipe~Vieira Frujeri, Shi Dong, Chenguang Zhu,
  Michael~I Jordan, and Jiantao Jiao.
\newblock Fine-tuning language models with advantage-induced policy alignment.
\newblock \emph{arXiv preprint arXiv:2306.02231}, 2023.

\bibitem[Ziebart(2010)]{ziebart2010modeling}
Brian~D Ziebart.
\newblock \emph{Modeling purposeful adaptive behavior with the principle of
  maximum causal entropy}.
\newblock Carnegie Mellon University, 2010.

\bibitem[Ziebart et~al.(2013)Ziebart, Bagnell, and Dey]{ziebart2013principle}
Brian~D Ziebart, J~Andrew Bagnell, and Anind~K Dey.
\newblock The principle of maximum causal entropy for estimating interacting
  processes.
\newblock \emph{IEEE Transactions on Information Theory}, 59\penalty0
  (4):\penalty0 1966--1980, 2013.

\end{thebibliography}
\bibliographystyle{iclr2025_conference}

\appendix
\section{Appendix}

\subsection{Experiment Setup and Additional Result}\label{app:sec:experiments}
\subsubsection{MuJoCo Tasks} \label{mujuco}
In MuJoCo, we consider several robotic control tasks with continuous action space. We evaluate the performance of our proposed algorithm on aligning robot behaviors with provided demonstrations and preference data. After the robot training stage, we leverage the ground-truth reward function from the environment to evaluate the performance.

{\bf Data.} {Following the similar data generation pipeline in \cite{brown2019extrapolating}}, we generate the expert demonstrations and preference dataset as follows. We first train an expert agent by leveraging the ground-truth reward function and the popular Soft Actor-Critic (SAC) algorithm \cite{haarnoja2018soft},  {which is developed to solve policy optimization problems with continuous action space.} During the training process, we save the policy checkpoints and collect $30$k samples from each checkpoint. {To achieve precise control of dataset quality, we categorize the data collected into three different classes: low-, medium-, and high-quality datasets according to the performance of the checkpoints. 
Then we combine the low- and medium-quality data as the preference dataset and use high-quality as demonstration data.

\begin{figure}[ht]
    \centering
    \includegraphics[width=4.5cm]{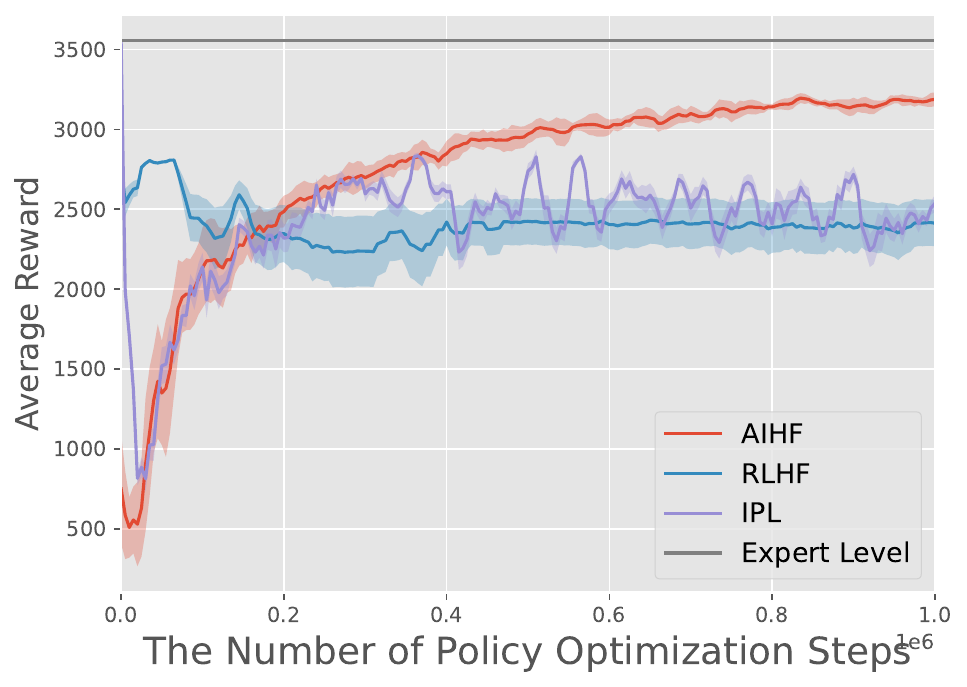}
    \includegraphics[width=4.5cm]{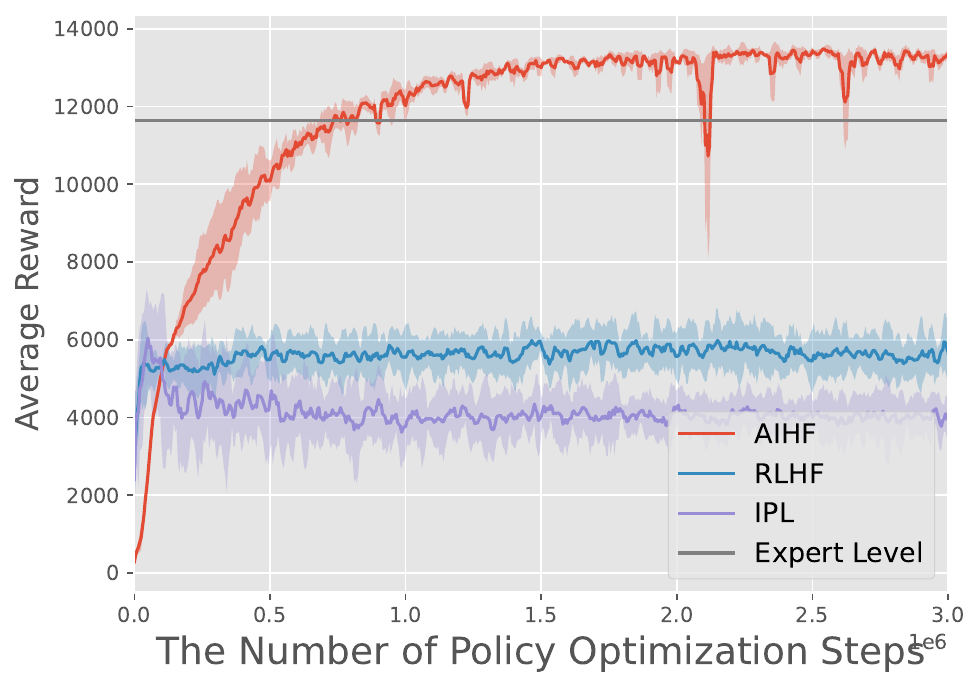}
    \includegraphics[width=4.5cm]{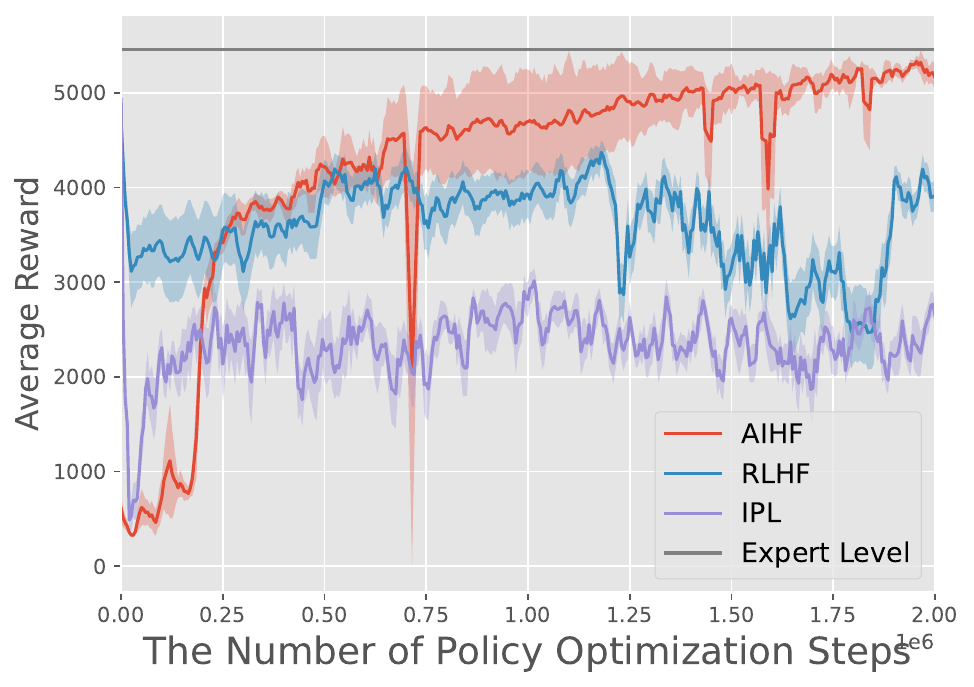}
    \caption{\textbf{Top-Left: Hopper Environment;} \textbf{Top-Right: HalfCheetah Environment;} \textbf{Bottom: Walker2d Environment}; \aname~ ({\color{orange}orange}) vs RLHF~({\color{blue}blue}) vs IPL~({\color{violet}purple}) \citep{hejna2024inverse}; results are averaged over 3 independent runs. {We use 10k demonstrations and 20k preferences.} The RLHF and IPL curve is initialized from a policy pre-trained by BC; the \aname~from a random policy. The performance is compared against the \# of SAC steps performed (for \aname~each policy alignment  performs 5k steps of SAC.)}
    \label{fig: MuJoCo Result}
\end{figure}

\noindent{\bf Results.} We show that \aname~is able to integrate (insufficient amount of) demonstration data and (not-so-high-quality) preference data to generate high-quality policy, and it significantly outperforms the RLHF and IPL. Note that IPL \citep{hejna2024inverse} is actually the DPO-type algorithm applied to multi-horizon MDP, therefore it is an ideal baseline for the MuJoCo setting (since the underlying problem is a multi-horizon MDP). In Figure \ref{fig: MuJoCo Result},  We can see that AIHF outperforms both RLHF and IPL algorithms. We also observe that due to the limited number of demonstration data, even BC could provide a high-performing initialization, RLHF still fails to improve policy quality in the following RL stage \cite{ross2010efficient,zeng2022maximum}. Moreover, since the preference data quality is only of low-to-medium quality, the RL step based on the learned reward model fails to significantly boost the fine-tuning performance. In contrast, clearly the proposed AIHF can effectively integrate preferences and demonstrations, leading to a more robust reward function and consequently, a high-quality policy. % {\color{red}[mention IPL as baseline.]}

In SAC, both the policy network and Q network are (64, 64) MLPs with ReLU activation function, and the step size is set to $3*10^{-3}$, we parameterize the reward function by a (64, 64) MLPs with ReLU activation function. For the reward network, we use Adam as the optimizer, and the step size is set to be $1*10^{-4}$.

The quality of the preference dataset and demonstration dataset are listed in Tab. \ref{tab:dataset_quality}.

\begin{table}[ht]
    \centering
    \resizebox{0.80\textwidth}{!}{%
    \begin{tabular}{c |c c c}
        \toprule
        \diagbox [width=3cm] {\textbf{Task}}{\textbf{Dataset}} & Non-prefer Data & Prefer Data & Demonstration Data\\
        \midrule
        Hopper-v2  & $2345.20 \pm 329.93$ & $3024.63 \pm 40.52$ & $3559.61 \pm 73.12$\\
       HalfCheetah-v2  &$7226.37 \pm 126.88$ &$9434.42 \pm 1315.13$ &$11635.42 \pm 236.51$\\
       Walker2d-v2 &$3952.60 \pm 444.45$ &$5091.71 \pm 291.73$ & $5453.41 \pm 71.07$\\
       \bottomrule
    \end{tabular}}
    \caption{The quality of preference and demonstration.}\label{tab:dataset_quality}
\end{table}

\subsubsection{Sentiment-Controlled Generation}

\textbf{Dataset Generation:} In the IMDb sentiment completion task, we generate the demonstrations and preference datasets using the following procedure. Initially, we train a Language Model by employing the ground-truth reward function \textsc{distilbert-imdb} and the Proximal Policy Optimization (PPO) algorithm on 30\% of the training dataset for IMDb. Throughout the training process, we save the policy checkpoint every 500 PPO steps. Subsequently, we select an additional 40\% of the training dataset and generate a response for each prompt for each checkpoint. According to the evaluation score of each generation, we categorize the data collected into different classes: low-, medium-, and high-quality datasets, then we combine low-quality and medium-quality as preference datasets, and use high-quality as demonstration datasets.

\textbf{Training:} After acquiring the preference and demonstration datasets, we train the proposed algorithm \aname~and baselines on the remaining 30\% of prompts from the training dataset. We evaluate the performance of each algorithm using the test datasets for IMDb and HH, along with their corresponding ground truth reward functions. For the GPU resources, we use 8$\times$ A100 40G for all the experiments.

\noindent \textbf{Results: Policy Quality.} We find that the proposed approach works well when either preference or demonstration data, or both, are limited. From the \ref{app:imdb:kl:AIHF}, we see that by using the same amount of data (10k preference, 10k demonstration), AIHF-based algorithms achieve faster convergence than their RLHF and DPO counterparts.

\noindent \textbf{Results: Distribution Mismatch.} We show that \aname~ is able to alleviate the distribution mismatch between the generated trajectories by the policy, and the data that the learned reward model is able to rank. To evaluate the extend of such mismatch, we use the following three steps: (1) use 1k preference, 1k demonstration to train policy and reward model for RLHF and \aname~; (2) for a given set of prompts from test dataset, use RLHF and \aname~ to perform generation; (3) use the trained reward models %,  take \textsc{lvwerra/distilbert-imdb} as the ground-truth reward, 
to rate the generation; (4) compare with the score generated by the ground truth reward \textsc{lvwerra/distilbert-imdb}.  Fig. \ref{fig:Mismatch-label} illustrates that the reward score distribution produced by \aname~ aligns closely with that of the ground truth reward, whereas that generated by RLHF exhibits a poor match. These results show that the reward model learned by \aname~is able to correctly evaluate the generation produced by the final policy.

%\dk{Where is result for helpfulness?}

\begin{figure}[H]
    \centering
    \includegraphics[width=4cm]{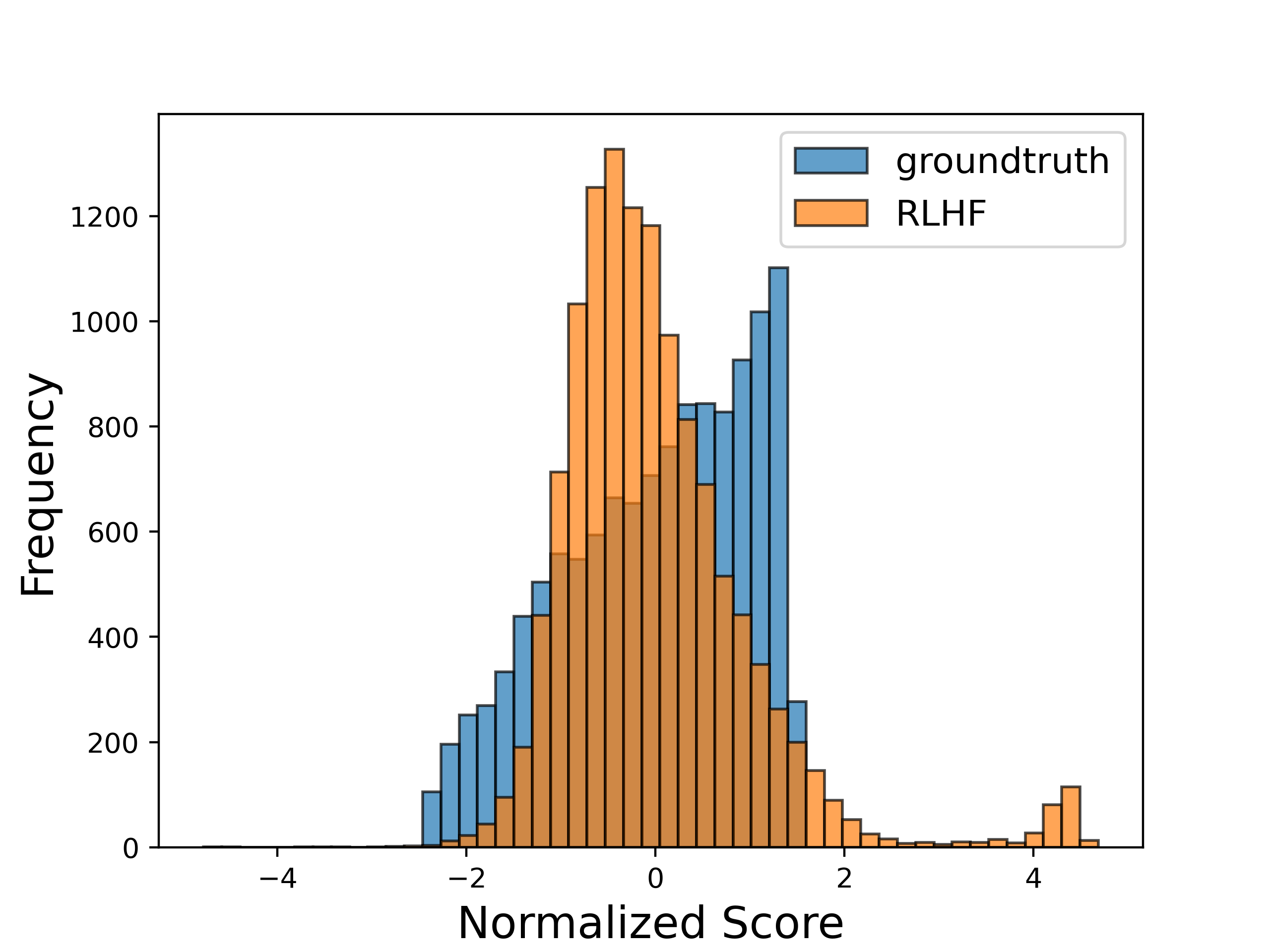}
    \includegraphics[width=4cm]{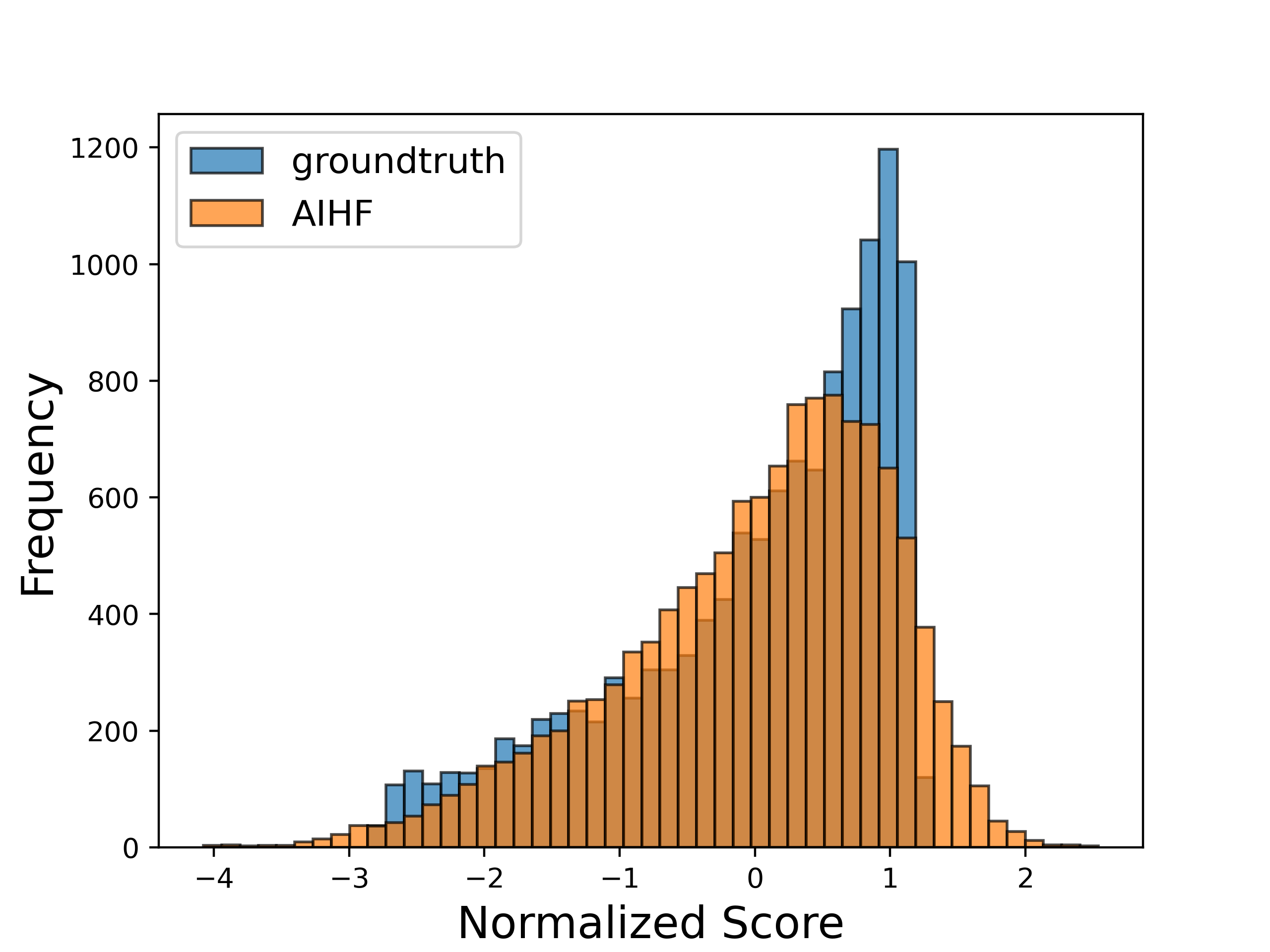}
    \caption{
    Comparison of the distribution of reward score generated by the trained reward models, and the ground truth reward model.  RLHF vs ground truth (left); \aname~vs ground truth (right).
% \dk{fonts are too tiny to read}
    }
    \label{fig:Mismatch-label}
\end{figure}

From Fig.\ref{app:imdb:kl:AIHF}, our proposed algorithm \aname~ could obtain higher rewards than baseline methods in the IMDb setting for almost all KL values. Although \aname~ might get a low score from the ground truth reward model in the earlier step, \aname~ would get a higher reward with more iteration and optimization steps. This indicates that with the mix of demonstration data and preference data, we could prevent the policy from known issues of reward hacking, especially when the policy learned more human-aligned features beyond base models (high KL value). Moreover, \aname~ is persistent in the number of preference data, presenting that \aname~ could still gain benefit from the limited preference data in more optimization steps as long as the demonstration data is high quality enough.

\begin{figure}[H]
    \centering
    \includegraphics[width=0.30\textwidth]{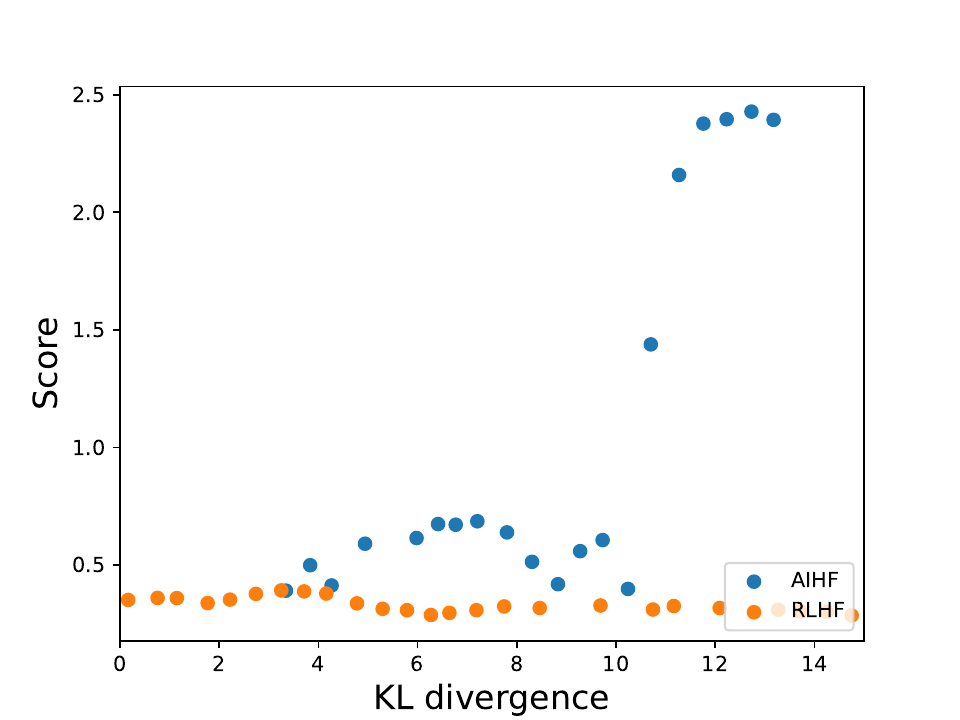}
    \includegraphics[width=0.30\textwidth]{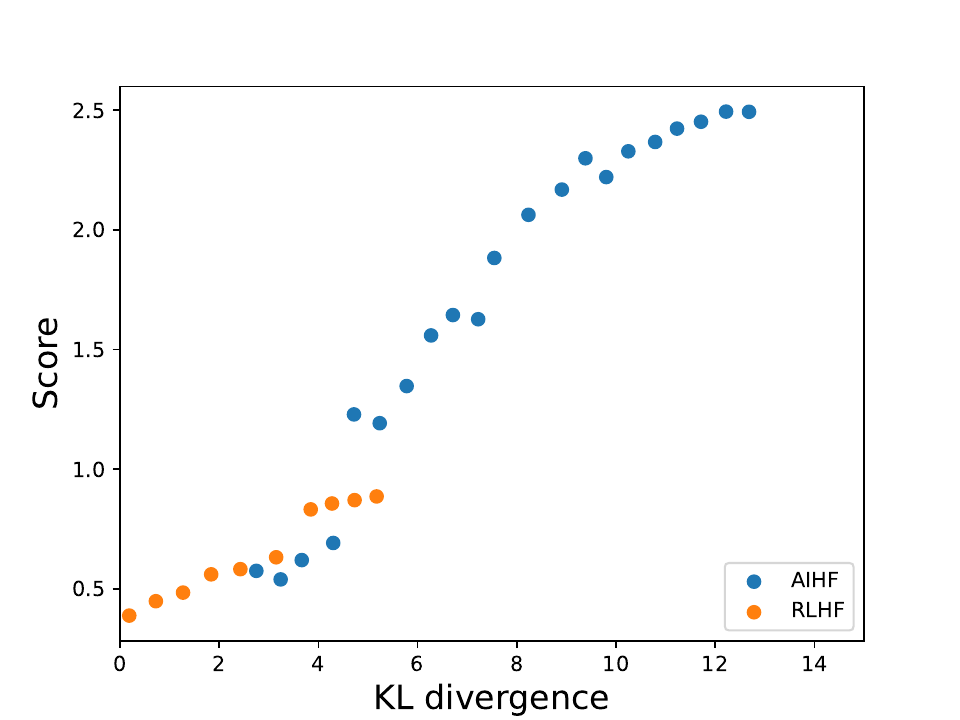}
    \includegraphics[width=0.30\textwidth]{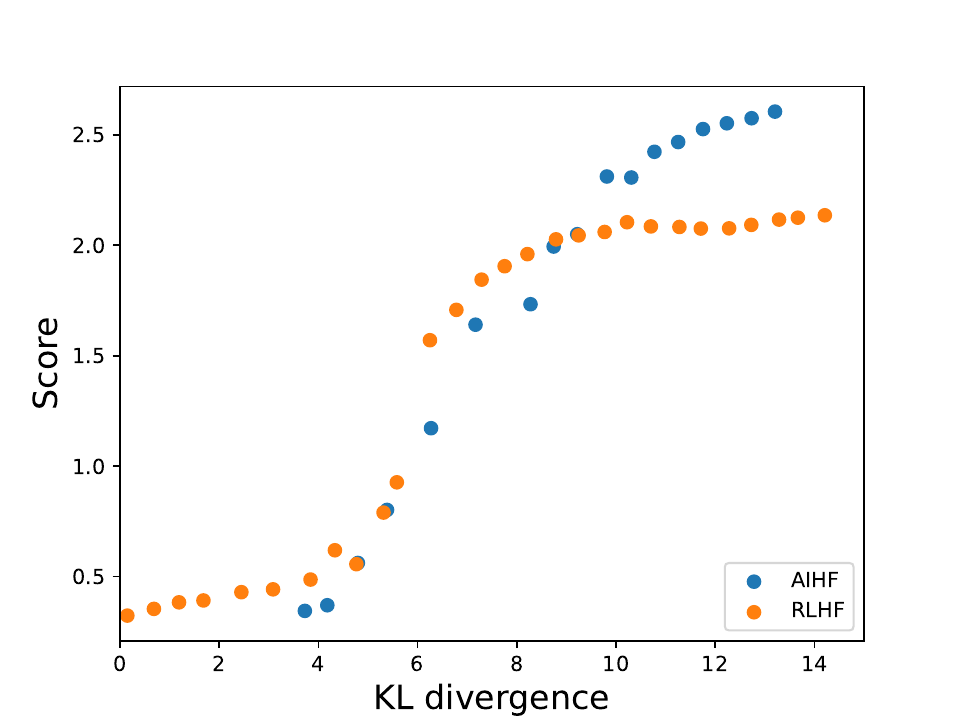}
    \caption{\textbf{The frontier of expected reward vs KL to the reference policy in IMDB dataset. fix the demonstration number to 3k} Left: Using 1k preference; Middle: Using 2k preferences; Right: Using 3k preference}
    \label{app:imdb:kl:AIHF}
\end{figure}

\subsubsection{The result of 1B and 2.8B experiments}

\noindent \textbf{Additional Results.} We show here the distribution of the reward values of the continuation generated by the reward models trained by AIHF and RLHF. In Fig. \ref{fig:reward distribution on HH}, we can observe the distribution of AIHF and RLHF have overlaps in low-quality continuation, however, AIHF can generate more high-quality continuations compared to RLHF, which shows that joint optimization can more effectively align the policy model with the demonstration distribution.

\begin{figure}[H]
    \centering
    \includegraphics[clip, trim={0cm 0cm 0cm 0cm}, width=0.49\textwidth]{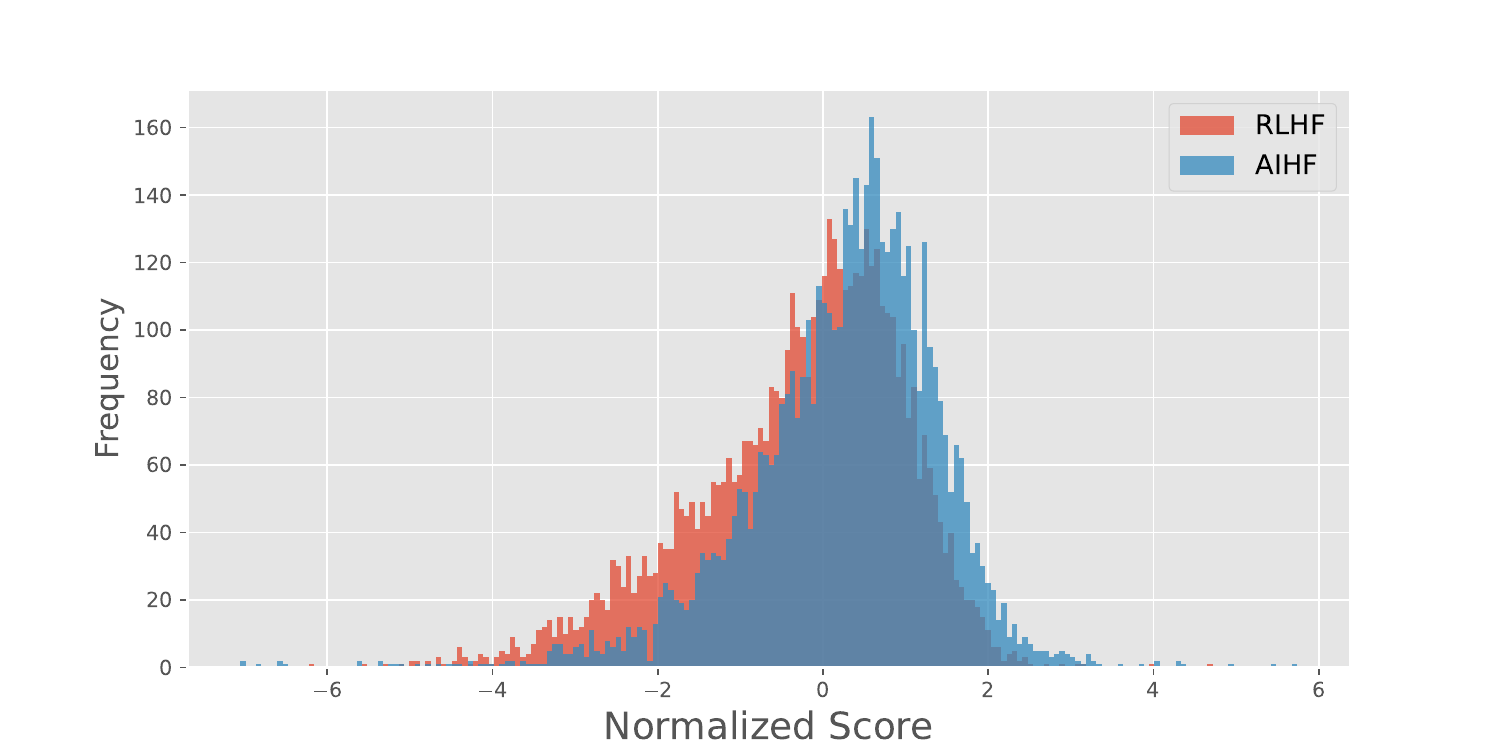}
    \includegraphics[clip, trim={0cm 0cm 1cm 1cm}, width=0.49\textwidth]{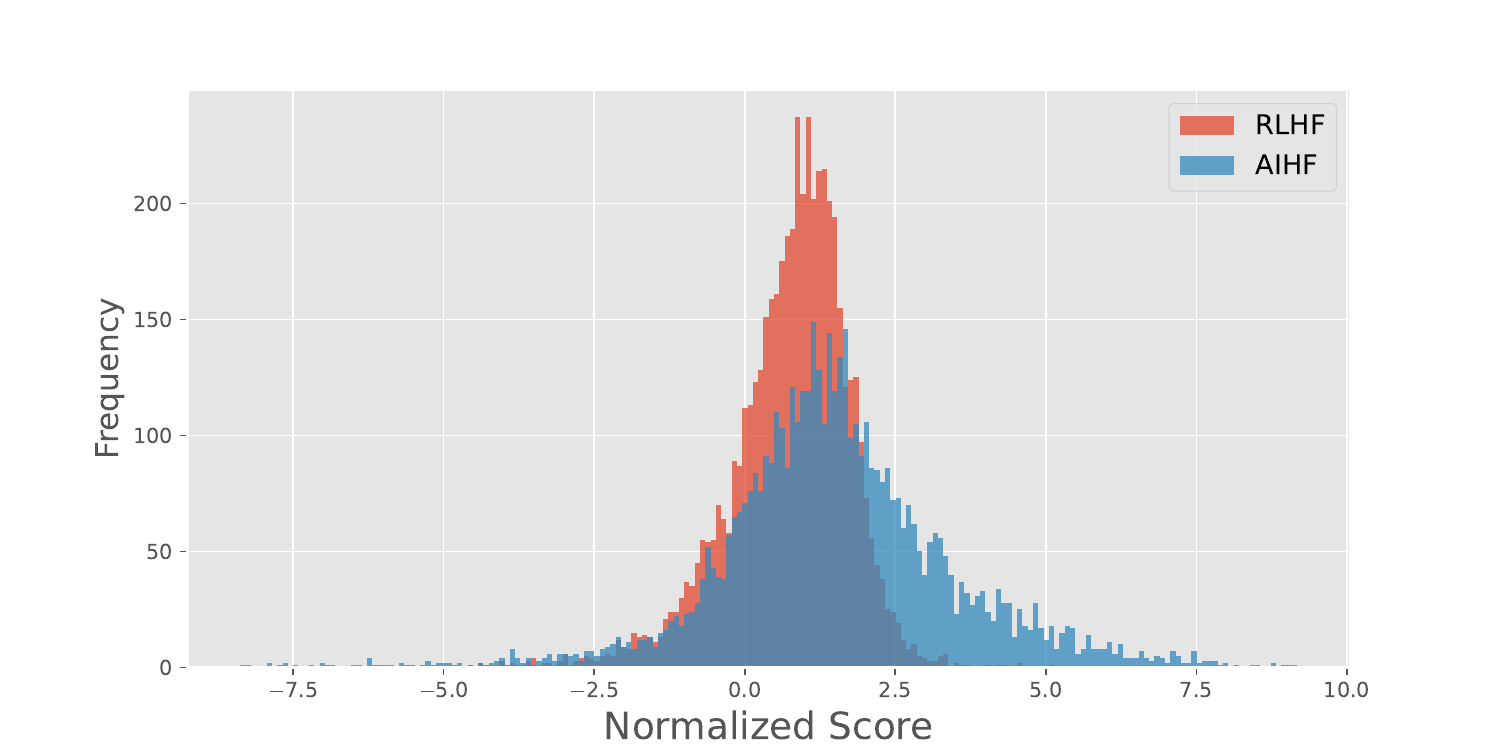}
    \caption{\textbf{The Reward Distribution of Helpfulness-controlled Generation. Left: Result on 160m model, Right: Results on 1B model}, This figure reports the reward distribution of generation evaluated by PKU-Alignment/beaver-7b-v3.0-reward for AIHF and RLHF.}
    \label{fig:reward distribution on HH}
\end{figure}

We record the performance of the two variants, namely Direct AIHF \eqref{eq:AFHF_redution_dpo} and Self-Play-AIHF \eqref{eq:AIHF_SPIN}, also RLHF, DPO, IPO and SPIN in Tab. \ref{tab:policy_quality}. Our proposed AIHF still outperforms all these methods in this experiment setting.

\begin{table}[ht]
    \centering
    \resizebox{0.99\textwidth}{!}{%
    \begin{tabular}{c |c c c c c c c}
        \toprule
        \textbf{Dataset} & \textbf{DPO} & \textbf{SPIN} & \textbf{IPO} & \textbf{RLHF} & \textbf{Direct AIHF} & \textbf{Self-Play-AIHF} &\textbf{AIHF} \\ \midrule
        10k Demonstrations, 5k preferences & $0.463 \pm 0.093$ & $0.625 \pm 0.048$ &$0.616 \pm 0.076$  & $0.710 \pm 0.085$  & $0.752 \pm 0.036$ & $0.640 \pm 0.102$& $1.167 \pm 0.157$ \\
        10k Demonstrations, 10k preferences & $0.474 \pm 0.052$ & $0.625 \pm 0.048 $& $0.650 \pm 0.017$  & $0.896 \pm 0.095$ & $0.798 \pm 0.026$ & $0.674 \pm 0.096$ & $1.190 \pm 0.069$ \\\bottomrule
    \end{tabular}}
    \caption{Evaluated reward scores of fine-tuned 1B Pythia models in HH dataset, evaluated by PKU-Alignment/beaver-7b-v3.0-reward. }\label{tab:policy_quality}
\end{table}
\subsubsection{The result of 7B experiments}

We utilize DeepSpeed ZeRO-3 \citep{Rajbhandari2020} to reduce the memory cost, and we use VLLM \citep{kwon2023efficient} for accelerate data generation/inference. We use eight NVIDIA A100-40G to do the training with a per-device batch size of 1 for 7B model. We train all models with bfloat16 precision. We set the learning rate to be 5e-7 for the 7B model with the cosine learning rate scheduler, and the max sequence length is set to be 512.

Our device cannot conduct the implementation of the standard AIHF \eqref{eq:AFHF} using PPO. Therefore, we employed an online DPO approach \citep{xiong2024iterative} for the lower-level policy optimization: 1) at each policy optimization iteration, we first generate multiple continuations from the current policy model over the given prompt dataset, 2) then we utilize our estimated explicit reward models to score each continuation, 3) over each prompt, choose the generated continuation with the highest reward score and the one with lowest reward score as a generated preference pair to run DPO algorithm to fine-tune the policy model.  

For Self-Play AIHF \eqref{eq:AIHF_SPIN} implementation, we adopt the same strategy as \cite{chen2024self}, at each epoch, we generate samples with picked 50k data and generate continuation $\tilde{a}\sim \pi(\cdot|s)$ using the current model $\pi$, then optimize \eqref{eq:AIHF_SPIN} with the sampled $\tilde{a}$. 

%\section{Details of the open LLM Leaderboard}
To evaluate the effectiveness of our approach, we assessed our fine-tuned models on the widely used HuggingFace Open
LLM Leaderboard \citep{open-llm-leaderboard} in Table \ref{tab:metrics}.
We also list the metric and number of shots used for LLM evaluation on each dataset in Table \ref{tab:leaderboard}. Moreover, in Tab. \ref{tab:reward_bench}, we also utilize the Reward-Bench \citep{lambert2024rewardbench} to evaluate the explicit reward models (estimated by standard preference learning in \eqref{eq:preference_gap_maximization} and AIHF) and the implicit reward model (extracted from DPO). The implicit reward is calculated by the policy model similar to what the DPO paper \cite{rafailov2023direct} proposed, i.e. $r(a|s) := \log(\frac{\pi(a|s)}{\pi^0(a|s)})$, the first explicit reward is estimated by BTL reward model \eqref{eq:preference_gap_maximization}, and second explicit reward is estimated by AIHF. The evaluation results show that AIHF can outperform the benchmark methods in terms of performance for both policy models and reward models. We also conduct an ablation study with different choices of $w_1$ in \eqref{eq:AFHF_redution_dpo}, as shown in Tab. \ref{tab:AIHF_weight_ablation},  the improvement of joint learning methods over baseline is robust.

\begin{table}[ht]
    \centering
    \resizebox{0.99\textwidth}{!}{%
    \begin{tabular}{c |c c c c c c}
        \toprule
        Dataset & Arc Challenge & TruthfulQA MC2 & Winogrande & GSM8K & HellaSwag & MMLU \\ \midrule
        Metric & acc & acc & acc & strict-match & acc\_norm & acc \\ 
        Num. of Shots & 25 & 0 & 5 & 5 & 10 & 5 \\ \bottomrule
    \end{tabular}}
    \caption{A summarization of the benchmarks we use in this work. We list the metric and number of shots used for LLM evaluation on each dataset.}\label{tab:metrics}
\end{table}

\begin{table}[ht]
    \centering
        \resizebox{0.99\textwidth}{!}{%
        \begin{tabular}{c | c c c c c c c}
        \toprule
        Tasks & Arc Challenge & TruthfulQA MC2 & Winogrande & GSM8k & HellaSwag & MMLU & Avg \\
        \midrule
        \texttt{mistral-7b-sft-beta} & 54.69\% & 42.96\% & 77.27\% & 39.88\% & 82.23\% & 59.72\% & 59.46\% \\
        \texttt{zephyr-7b-beta} & 59.64\% & 55.18\% & 77.82\% & 33.51\% & 84.19\% & 59.76\% & 61.68\% \\
        \midrule
        \texttt{SPIN} & 58.45\% & 43.66\% & 78.30\% & 39.50\% & 83.59\% & 58.60\% & 60.35\% \\
        \texttt{DPO} & 62.80\% & 53.17\% & \textbf{79.40\%} & 39.20\% & 85.13\% & 59.41\% & 63.19\% \\
        \texttt{IPO} & 58.02\% & 48.29\% & 79.24\% & 42.91\% & 83.93\% & 60.07\% & 62.08\% \\
        % \texttt{Direct AIHF($w_1=0.01$)} & 66.04\% & 57.55\% & 79.08\% & 36.61\% & 85.58\% & \textbf{60.09\%} & 64.15\% \\
        % \texttt{Direct AIHF($w_1=0.001$)} & \textbf{66.72}\% & 58.73\% & \textbf{79.16\%} & 36.84\% & 85.59\% & 59.26\% & 64.38\% \\
        \texttt{Direct AIHF} & 61.17\% & \textbf{60.03\%} & 79.00\% & 39.80\% & 85.71\% & 60.02\% & 64.29\% \\
        % \texttt{Self-play AIHF($w_1=1$) Iter1} & 63.22\% & 53.58\% & 77.74\% & 41.16\% & 84.58\% & 59.27\% & 63.25\% \\
        % \texttt{Self-play AIHF($w_1=1$) Iter2} & 64.41\% & 54.88\% & 78.61\% & 39.87\% & 85.14\% & 59.19\% & 63.68\% \\
        % \texttt{Self-play AIHF($w_1=1$) Iter3} & 65.95\% & 57.78\% & 78.53\% & 42.22\% & 85.50\% & 58.68\% & 64.77\% \\
        \texttt{Self-play AIHF} & 61.77\% & 58.29\% & 78.53\% & \textbf{44.20\%} & 85.53\% & 58.66\% & 64.50\% \\
        \texttt{AIHF} & \textbf{63.90\%} & 58.38\% & 79.24\% & 40.56\% & \textbf{86.23\%} & \textbf{60.18\%} & \textbf{64.75\%} \\
        \bottomrule
        \end{tabular}%
        }
     \caption{Test performance of Direct AIHF and Self-Play AIHF based on mistral-7b-sft-beta across HuggingFace Open LLM Leaderboard datasets.}
     \label{tab:leaderboard}
\end{table}

% \begin{table}[!h]
% \begin{center}
% \resizebox{0.99\textwidth}{!}{
% \begin{tabular}{|c|c|c|c|c|c|}
%     \hline
%     \textbf{Reward Model} & \textbf{Chat} & \textbf{Chat Hard} & \textbf{Safety} & \textbf{Reasoning} & \textbf{Average} \\
%     \hline
%     Base (mistral-7b-sft-full) & 79.19\% & {39.25\%} & 42.94\% & 47.29\% & 52.16\% \\
%     \hline
%     Implicit RM (UltraFeedback) & 37.43\% & 55.92\% & 64.14\% & 47.33\% & 51.205\% \\ \hline
%     Explicit RM (UltraFeedback)  & 95.11\% & {56.58\%} & 63.69\% & 69.22\% & 71.15\% \\
%     \hline
%      Explicit RM (UltraFeedback + UltraChat) &  94.41\%& 55.37\% & 63.98\% & 76.75\% & 72.63\% \\
%      \hline
% \end{tabular}
% }
% \end{center}
% \caption{Performance of Reward Models in Reward-Bench.}
% \label{tab:reward_bench}
% \end{table}

\begin{table}[ht]
    \centering
    \resizebox{0.85\textwidth}{!}{%
    \begin{tabular}{c |c c c c c }
        \toprule
        \textbf{Reward Model} & \textbf{Chat} & \textbf{Chat Hard} & \textbf{Safety} & \textbf{Reasoning} & \textbf{Average} \\ \midrule
        % Base (mistral-7b-sft-full) & 79.19\% & {39.25\%} & 42.94\% & 47.29\% & 52.16\% \\
        DPO Reward Model & 37.43\% & 55.92\% & 64.14\% & 47.33\% & 51.21\% \\ 
        BTL Reward Model  & \textbf{95.11\%} & \textbf{56.58\%} & 63.69\% & 69.22\% & 71.15\% \\
        AIHF Reward Model &  94.41\%& 55.37\% & \textbf{63.98\%} & \textbf{76.75\%} & \textbf{72.63\%} \\\bottomrule
    \end{tabular}}
    \caption{Evaluation of Reward Models in Reward-Bench. }\label{tab:reward_bench}
    % {\red[DPO RM, BTL RM, AIHF RM. Why were are empahsizing implicit and explicit here?]}
\end{table}

\begin{table}[ht]
    \centering
        \resizebox{0.99\textwidth}{!}{%
        \begin{tabular}{c | c c c c c c c}
        \toprule
        Tasks & Arc Challenge & TruthfulQA MC2 & Winogrande & GSM8k & HellaSwag & MMLU & Avg \\
        \midrule

        \texttt{Direct AIHF($w_1=0.01$)} & 61.86\% & 57.55\% & 79.08\% & 36.61\% & 85.58\% & \textbf{60.09\%} & 63.46\% \\
        \texttt{Direct AIHF($w_1=0.001$)} & \textbf{63.25}\% & 58.73\% & \textbf{79.16\%} & 36.84\% & 85.59\% & 59.26\% &  63.80\% \\
        \texttt{Direct AIHF($w_1=0.0001$)} & 61.17\% & \textbf{60.03\%} & 79.00\% & \textbf{39.80\%} & \textbf{85.71\%} & 60.02\% &  \textbf{64.28\%} \\

        \bottomrule
        \end{tabular}%
        }
     \caption{Test performance of Direct AIHF over different choices of the balancing coefficient across HuggingFace Open LLM Leaderboard datasets.}
     \label{tab:AIHF_weight_ablation}
\end{table}

\subsection{Why AIHF Can Outperform Two-Stage Alignment Approaches} \label{app:why}
In this section, we provide detailed analysis for Sec. \ref{sec:why}.

\subsubsection{SFT and RLHF Policy}\label{app:rlhf}
We revisit the RLHF pipeline in the context of a simple softmax choice model
where $\tau_i \in A$ is selected with probability 
\begin{align}\label{eq:discrete:policy}
    \pi^*_i(R)=\frac{\exp (R_i/\beta)}{\sum_{j=1}^N\exp (R_j/\beta)}.
\end{align}
 For simplicity, we have defined $R_i:=R(\tau_i)$. Using this policy model, below let us analyze different policies according to different ways that the rewards are learned. 
   
{\bf Supervised Fine-Tuning (SFT)}: Given a demonstration dataset $\mathcal{D}$ the goal is the find the policy $\pi_{\rm SFT}$ that maximizes likelihood, i.e.:
\begin{align} \label{eq:example:sft:policy}
\pi_{\rm SFT}&:=\arg \max_{\pi}\sum_{\ell=1}^{N}
\mathbb{E}_{\tau_\ell \sim \mathcal{D}}\left[
\log \pi^*_\ell(R)\right], \quad \mbox{\rm s.t.}\; \sum_{\ell=1}^{N} \pi^*_\ell(R)=1.
\end{align}
Next let us identify the optimal reward and its corresponding policy in this case. First, let us find the reward function. Write \eqref{eq:example:sft:policy} in terms of reward optimization, we have:
\begin{align}
  R^* = \arg\max_{R}\sum_{\ell=1}^{N}\mathbb{E}_{\tau_\ell\sim D} \log\frac{\exp(R_\ell/\beta)}{\sum_{j=1}^{N}\exp(R_j/\beta)} =: L_1(R)  
\end{align}
The partial derivative of the objective function w.r.t. a component $R_i, i\in[1,\cdots, N]$ is given by:
\begin{align}\label{eq:partial:sft}
   \frac{\partial L_1(R)}{\partial R_i} & = \sum_{\ell:\ell\ne i}\mathbb{E}_{\tau_\ell \sim D}\left [- \frac{1}{\beta}\frac{\exp(R_i/\beta)}{\sum_{j=1}^{N}\exp(R_j/\beta)}\right] + \mathbb{E}_{\tau_i \sim D}\left [\frac{1}{\beta} - \frac{1}{\beta}\frac{\exp(R_i/\beta)}{\sum_{j=1}^{N}\exp(R_j/\beta)}\right]\nonumber\\
   & =\frac{1}{\beta}\left(\frac{\#\{\tau_i\in \mathcal{D}\}}{|P|} - \sum_{\ell}\mathbb{E}_{\tau_{\ell}\in D} \left[\frac{\exp(R_i/\beta)}{\sum_{j=1}^{N}\exp(R_j/\beta)}\right]\right), \quad i\in[1:N].
\end{align}
Setting the above partial gradient to zero, we obtain 
\begin{align}\label{eq:sft:system}
\frac{\#\{\tau_i \in \mathcal{D}\}}{|\mathcal{D}|} =\frac{\exp (\widehat{R}_{\mathcal{D}}(\tau_i)/\beta)}{\sum_{j=1}^N\exp (\widehat{R}_{\mathcal{D}}(\tau_j)/\beta)} ~~~~i\in [1:N],
\end{align}
where we used $\widehat{R}_D(\tau_i)$'s to denote the optimal reward estimated from the set of SFT data $\mathcal{D}$. Then according to the definition of the policy given earlier, we obtain
$$\pi^*_i(\widehat{R}_{\mathcal{D}}) =\frac{\# \{\tau_i \in \mathcal{D}\}}{|\mathcal{D}|},\; \forall~i\in [1,\cdots N].$$
Note that the optimal reward learned from the SFT data $\widehat{R}_D$ should satisfy the system \eqref{eq:sft:system}. It turns out that there is a unique solution to the following system of equations:
\begin{align}
\frac{\#\{\tau_i \in \mathcal{D}\}}{|\mathcal{D}|} =\frac{\exp (\widehat{R}_{\mathcal{D}}(\tau_i)/\beta)}{\sum_{j=1}^N\exp (\widehat{R}_{\mathcal{D}}(\tau_j)/\beta)} ~~~~i\in \{2,\dots,N\}
\label{eq:SFT}
\end{align}
with $\widehat{R}_{\mathcal{D}}(\tau_1)=\bar{R}_1$ a fixed reference value. 
%This uniqueness result can be obtained from 
%\citet[Proposition 1]{hotz1993conditional}.
%{\color{red} Note: We don't need to cite Hotz and Miller which prove uniqueness for the dynamic model.}
A simple argument is provided below. The system \eqref{eq:SFT} can be re-written as:
\begin{align*}
\log \frac{\#\{\tau_i \in \mathcal{D}\}}{|\mathcal{D}|}
-
\log \frac{\#\{\tau_1 \in \mathcal{D}\}}{|\mathcal{D}|}
&=
\frac{1}{\beta}\big(\widehat{R}_{\mathcal{D}}(\tau_i)
-
\bar{R}_1 \big)~~~~~i \in \{2,\dots,N\}
\end{align*}
Hence, rewards are uniquely determined by demonstration data with a fixed reference value $\widehat{R}_{\mathcal{D}}(\tau_1)=\bar{R}_1$. 

{It is important to note that, despite the fact that in the SFT we may not {\it directly} learn a reward function, the above analysis says that we are {\it implicitly} learning a reward $\widehat{R}_D$, assuming that the policy is parameterized as \eqref{eq:discrete:policy}. This {\it implicit} reward will be used later to analyze the RLHF policy. Indeed, if we solve the following SFT {\it policy optimization} problem directly
\begin{align}
    \max \sum_{\ell}\mathbb{E}_{\tau_\ell\in \mathcal{D}} \log(\pi_\ell), \quad {\rm s.t.}\quad \sum_{i\in\mathcal{D}}\pi_i = 1,
\end{align}
one can easily obtain that the optimal policy is given by $\pi^*_i=\frac{\# \{\tau_i \in \mathcal{D}\}}{|\mathcal{D}|},\; \forall~i\in [1,\cdots N]$.
}

\noindent{\bf Reward Modeling with Preference Data}: Now let us analyze the case where the reward is only learned from a preference dataset. 
With preference data $\mathcal{P}:=\{(\tau_i\succ \tau_j)\}$, the BTL model is:
\begin{align*}
P(\tau_i \succ \tau_j)&= \sigma \big( \frac{1}{\beta}\big(R(\tau_i) - R(\tau_j) \big)=\frac{\pi^*_i(R)}{ \pi^*_i(R)+\pi^*_j(R)}. 
\end{align*}

The reward estimation problem is defined as:
\begin{align}
\widehat{R}_{\mathcal{P}}= \arg \max_R L_2(R) & := \sum_{\ell,j: \ell\ne j}\mathbb{E}_{(\tau_\ell \succ \tau_j) \sim \mathcal{P}}\Big[\log \frac{\pi^*_\ell(R)}{\pi^*_\ell(R) + \pi^*_j(R)}\Big]\\ &= \sum_{\ell,j: \ell\ne j}\mathbb{E}_{(\tau_\ell \succ \tau_j) \sim \mathcal{P}}\left[\log \left(\frac{\exp(R_{\ell}/\beta)}{\exp(R_{\ell}/\beta)+ \exp(R_{j}/\beta)}\right) \right]
\end{align}
Take the gradient w.r.t. $R_i$, we obtain 
\begin{align}\label{eq:partial:rm}
\frac{\partial \ell_{RM}({R})}{\partial R_i}
& = \sum_{j: j\ne i}\mathbb{E}_{(\tau_i \succ \tau_j)}\left(\frac{1}{\beta}- \frac{1}{\beta}\left(\frac{\exp(R_{\ell}/\beta)}{\exp(R_{\ell}/\beta)+ \exp(R_{j}/\beta)}\right)\right)\nonumber\\
&\quad + \sum_{j: j\ne i}\mathbb{E}_{(\tau_j \succ \tau_i)}\left[- \frac{1}{\beta}\left(\frac{\exp(R_{\ell}/\beta)}{\exp(R_{\ell}/\beta)+ \exp(R_{j}/\beta)}\right)\right] \nonumber\\
& = \frac{1}{\beta}\sum_{j: j\ne i}\frac{|\mathcal{P}_{i\succ j}|}{|\mathcal{P}|}\left(1- \left(\frac{\exp(R_{\ell}/\beta)}{\exp(R_{\ell}/\beta)+ \exp(R_{j}/\beta)}\right)\right)\nonumber\\
&\quad + \frac{1}{\beta}\sum_{j: j\ne i}\frac{|\mathcal{P}_{j\succ i}|}{|\mathcal{P}|}\left(- \left(\frac{\exp(R_{\ell}/\beta)}{\exp(R_{\ell}/\beta)+ \exp(R_{j}/\beta)}\right)\right)\nonumber\\
& = \frac{1}{\beta}\left(\sum_{j: j\ne i}\frac{|\mathcal{P}_{i\succ j}|}{|\mathcal{P}|}- \frac{|\mathcal{P}_{i,j}|}{|\mathcal{P}|}\frac{\exp(R_{\ell}/\beta)}{\exp(R_{\ell}/\beta)+ \exp(R_{j}/\beta)}\right)\nonumber\\
& =\frac{1}{\beta}\sum_{j: j \neq i} \Big(\frac{|\mathcal{P}_{i\succ j}|}{|\mathcal{P}_{i,j}|} - \frac{\pi^*_i(R)}{ \pi^*_i(R)+\pi^*_j(R)} \Big)\frac{|\mathcal{P}_{i,j}|}{|\mathcal{P}|}
\end{align}
where in the above derivation we have defined $|\mathcal{P}_{i\succ j}|:=\# \{\tau_i \succ \tau_j ~\mbox{in}~ \mathcal{P}\}$ and $|\mathcal{P}_{i,j}|:=|\mathcal{P}_{i\succ j}|+|\mathcal{P}_{j\succ i}|$. Setting the gradient to zero, we obtain
\begin{align}\label{eq:policy:preference}
    \sum_{j: j \neq i} \Big(\frac{|\mathcal{P}_{i\succ j}|}{|\mathcal{P}_{i,j}|} - \frac{\pi^*_i(\widehat{R}_{P})}{ \pi^*_i(\widehat{R}_{P})+\pi^*_j(\widehat{R}_{P})} \Big)\frac{|\mathcal{P}_{i,j}|}{|\mathcal{P}|} =0. 
\end{align}
We shall denote by $\widehat{R}_{\mathcal{P}}$ as  a solution to the above system of equations.
The first-order condition \eqref{eq:policy:preference} can be written in implicit form as:
\begin{align}\label{eq:preference :policy}
    \pi^*_i(\widehat{R}_P) = \frac{\sum_{j:j\ne i}|\mathcal{P}_{i\succ j}|}{\sum_{j:j\ne i}\frac{|\mathcal{P}_{ij}|}{\pi_i^*(\widehat{R}_P)+\pi_j^*(\widehat{R}_P)}} = \frac{\sum_{j:j\ne i}|\mathcal{P}_{i\succ j}|}{\sum_{j:j\ne i}\frac{|\mathcal{P}_{ij}|}{1-\sum_{\ell \in A\backslash\{i,j\}} \pi^*_\ell(\widehat{R}_P) }} : = \frac{\sum_{j:j\ne i}|\mathcal{P}_{i\succ j}|}{\sum_{j:j\ne i}{|\mathcal{P}_{ij}|}\rho_{-(i,j)}(\pi^*(\widehat{R}_P))}
\end{align}
% \begin{align}
% \pi^*_i(\widehat{R}_{\mathcal{P}}) &=\frac{
% \sum_{j \neq i}|\mathcal{P}_{i\succ j}| }
% {
% \sum_{j\neq i}|\mathcal{P}_{i,j}|
% \rho_{-i}(\pi^*(\widehat{R}_{\mathcal{P}}))}
% \label{estimator_preferences}
% \end{align}
%where $\rho_{-i}(\pi):=\sum_{j \neq i}\rho_{-(i,j)}(\pi)$ and 
where we have defined $\rho_{-(i,j)}(\pi):=\Big(1-\sum_{\ell\in A \backslash \{i,j\}}\pi_\ell\Big)^{-1}$ as the expected number of times an action {\em other} than $\tau_i$ or $\tau_j$ is selected when sampling actions from $\pi$ infinitely many times. 

\noindent{\bf RLHF Policy.}
Based on the above two analysis, we are ready to analyze the RLHF policy. The RLHF policy is defined as follows:
\begin{align*}
    &\pi^{\rm RLHF}=\arg \max_{\pi \in \Delta^N} ~~\mathbb{E}_{\tau_i \sim \pi}\big[{\widehat{R}_{\mathcal{P}}}(\tau_i)\big] - \beta \mathrm{KL} (\pi||\pi^{\rm SFT}) 
\end{align*}
where {$\widehat{R}_{\mathcal{P}}$ is the estimator obtained from preference data, $\pi^{\rm SFT}$ is the SFT model trained with demonstration dataset $\mathcal{D}$, and $\Delta^N$ is the probability simplex. 

It can be easily shown that the solution $\pi^{\rm RLHF}$ is of the form:
\begin{align}
\pi^{\rm RLHF}(\tau_i)&=\frac{\pi^{\rm SFT}(\tau_i) \exp \left(\frac{1}{\beta}\widehat{R}_{\mathcal{P}}(\tau_i)\right)}
{\sum_{j=1}^N \pi^{\rm SFT}(\tau_j) \exp \left(\frac{1}{\beta}\widehat{R}_{\mathcal{P}}(\tau_j)\right)} \nonumber \\
&= {\frac{\exp \left(\frac{1}{\beta}\widehat{R}_{\mathcal{D}}(\tau_i)\right) \exp \left(\frac{1}{\beta}\widehat{R}_{\mathcal{P}}(\tau_i)\right)}{\sum_{j=1}^N \exp \left(\frac{1}{\beta}\widehat{R}_{\mathcal{D}}(\tau_j)\right) \exp \left({\frac{1}{\beta}\widehat{R}_{\mathcal{P}}}(\tau_j)\right)}} ~~~~ (\mbox{using} ~\eqref{eq:SFT})\nonumber \\
& ={\frac{\exp\big(\frac{1}{\beta}(\widehat{R}_{\mathcal{D}}(\tau_i)+\widehat{R}_{\mathcal{P}}(\tau_i))\big)}
{\sum_{j=1}^N \exp\big(\frac{1}{\beta}(\widehat{R}_{\mathcal{D}}(\tau_j) + \widehat{R}_{\mathcal{P}}(\tau_j))\big)}}\nonumber \\
&=\pi^*_i\Big(\widehat{R}_{\mathcal{D}}
+\widehat{R}_{\mathcal{P}}\Big).
\label{eq:ref_optimal_RLHF}
\end{align}

\subsubsection{The AIHF Policy}\label{appendix:aihf}

\noindent{\bf The AIHF Policy.} The proposed AIHF estimation problem is
\begin{align}\label{eq:AIHF:reward}
    \widehat{R}^{\rm AIHF} = \arg\max_{R}  \ell_{\mathcal{D}+\mathcal{P}}(R) :=  |\mathcal{D}| L_1(R)+ |\mathcal{P}|L_2(R)
\end{align}
 where  
$L_1(R):=\mathbb{E}_{\tau_i\sim \mathcal{D}}[\log \pi^*_i(R)]$ and $L_2(R):=\mathbb{E}_{(\tau_j \prec \tau_i) \sim \mathcal{P}}[\log \frac{\pi^*_i(R)}{\pi^*_i(R) + \pi^*_j(R)}]$, and $\pi^*_i(R)$ is given by \eqref{eq:discrete:policy}.

Similarly as before, taking gradient w.r.t. $R_i$, and leverage \eqref{eq:partial:sft} and \eqref{eq:partial:rm}, we obtain
\begin{align}
    \frac{\partial \ell_{\mathcal{D}+\mathcal{P}}(R)}{\partial R_i}& =\frac{\# \{\tau_i ~\mbox{in}~ \mathcal{D}\}}{|\mathcal{D}|} |\mathcal{D}| - \pi^*_i(R) |\mathcal{D}| + \sum_{j:j\neq i} \left(|\mathcal{P}_{i\succ j}|  -  \sum_{j \neq i}  \frac{\pi^{*}_i(R)}{ \pi^{*}_i(R)+\pi^{*}_j(R)} |\mathcal{P}_{i,j}|\right)
\end{align}
Setting the above condition to zero, we obtain that the AIHF reward should satisfy the following
% The first order condition is:
% \begin{align*}
% \frac{\# \{\tau_i ~\mbox{in}~ \mathcal{D}\}}{|\mathcal{D}|} |\mathcal{D}| + \sum_{j\neq i} |\mathcal{P}_{i\succ j}| - \pi^{*}_i |\mathcal{D}| -  \sum_{j \neq i}  \frac{\pi^{*}_i(\widehat{R}^{\rm AIHF})}{ \pi^{*}_i(\widehat{R}^{\rm AIHF})+\pi^{*}_j(\widehat{R}^{\rm AIHF})} |\mathcal{P}_{i,j}|=0
% \end{align*}
% Hence, the first-order condition can be re-written as:
%As an example, for the case with  $A=\{\tau_1,\tau_2,\tau_3\}$ the first order condition leads to:
\begin{align*}
\# \{\tau_i ~\mbox{in}~ \mathcal{D}\} + \sum_{j: j \neq i} |\mathcal{P}_{i \succ j}|
& =\pi^{*}_i(\widehat{R}^{\rm AIHF}) \Big( |\mathcal{D}|+\sum_{j:j \neq i}   \frac{|\mathcal{P}_{i,j}|}{ \pi^{*}_i(\widehat{R}^{\rm AIHF})+\pi^{*}_j(\widehat{R}^{\rm AIHF})}\Big) \\
&=\pi^{*}_i(\widehat{R}^{\rm AIHF}) \Big( |\mathcal{D}|+\sum_{j:j \neq i}|\mathcal{P}_{i,j}|\rho_{-(i,j)}\big(\pi^*(\widehat{R}^{\rm AIHF})\big)\Big)
\end{align*}
Or equivalently,
\begin{align}
\pi^{*}_i(\widehat{R}^{\rm AIHF})&=
\frac{\# \{\tau_i ~\mbox{in}~ \mathcal{D}\} + \sum_{j \neq i}|\mathcal{P}_{i \succ j}|} 
{|\mathcal{D}|+\sum_{j:j \neq i} |\mathcal{P}_{i,j}|
\rho_{-(i,j)}\big(\pi^*(\widehat{R}^{\rm AIHF})\big)} \label{eq:AIHF}
\end{align}
The system (\ref{eq:AIHF}) has a unique solution $\widehat{R}^{\rm AIHF}$ with a fixed reference value
 $\widehat{R}^{\rm AIHF}(\tau_1)=\bar{R}_1$.

\noindent{\bf Discussion.} Now let us compare the RLHF policy \eqref{eq:ref_optimal_RLHF} and AIHF policy \eqref{eq:AIHF}. First, observe that the RLHF policy is a function of the {\it sum} of two reward functions, one learned from the demonstration data and one learned from the preference data, while the AIHF policy takes a more complex form. To have a better understanding of the two policies, let us consider the extreme cases where the SFT and the preference datasets are {\it unbalanced}, where either $|\mathcal{D}|\gg |\mathcal{P}|$ or $|\mathcal{P}|\gg |\mathcal{D}|$. 
\begin{itemize}
    \item Case $|\mathcal{D}|\gg |\mathcal{P}|$. In this case, $\pi^{*}_i(\widehat{R}^{\rm AIHF})\approx \frac{\# \{\tau_i ~\mbox{in}~ \mathcal{D}\}}{|\mathcal{D}|} = \pi^*_i(\widehat{R}_{\mathcal{D}})$, so the AIHF policy will ignore the preference dataset, while focusing on the SFT policy; however, for the RLHF policy, one needs to first separately performs the reward estimation, but in this case $\hat{R}_P$ can be very noisy due to lack of data; Then the RLHF policy will also be noisy since it will be influenced by the noisy reward function $\hat{R}_P$. 
    \item Case $|\mathcal{P}|\gg |\mathcal{D}|$. In this case, let's further assume that $\sum_{j:j\ne i}|\mathcal{P}_{i\succ j}|\gg \# \{\tau_i ~\mbox{in}~ \mathcal{D}\}, \;\forall~i$, that is, for each trajectory $\tau_i$, the number of times it appears in the preference data is much larger than its appearance in the demonstration data. In this case, it is easy to see that $\pi^{*}_i(\widehat{R}^{\rm AIHF})\approx \pi^*_i(\widehat{R}_{\mathcal{P}})$, while the RLHF policy will still be noisy as in the first case. 
\end{itemize}

\subsubsection{Numerical Examples}\label{sec:numerical_example}
{\bf Example 1:} With $\beta=1$ and only two actions $\tau_1$ and $\tau_2$.
Since $\rho_{-(1,2)}(\pi)=\rho_{-(2,1)}(\pi)=1$, it follows from equations (\ref{eq:SFT}),  \eqref{eq:preference :policy}, (\ref{eq:AIHF}) that:
\begin{align*}
\pi^{\rm AIHF}_1:=\pi^*_1(\widehat{R}^{\rm AIHF})&=\frac{\# \{\tau_1 ~\mbox{in}~ \mathcal{D}\} +  \# \{\tau_2 \prec \tau_1 ~\mbox{in}~ \mathcal{P}\}}{|\mathcal{D}|+|\mathcal{P}|}\\
&=\frac{|\mathcal{D}|}{|\mathcal{D}|+|\mathcal{P}|}\pi^*_1(\widehat{R}_{\mathcal{D}})
+\frac{|\mathcal{P}|}{|\mathcal{D}|+|\mathcal{P}|}\pi^*_1(\widehat{R}_{\mathcal{P}}). 
\end{align*}

Slightly abusing notations,  let $\pi^*_1:=\pi^*_1(R^*)$ where $R^*$ is the ground-truth reward. It follows that $\mbox{Var}(\pi^*_1(\widehat{R}_{\mathcal{D}}))=\frac{\pi^*_1(1-\pi^*_1)}{|\mathcal{D}|}$, 
$\mbox{Var}(\pi^*_1(\widehat{R}_{\mathcal{P}}))=\frac{\pi^*_1(1-\pi^*_1)}{|\mathcal{P}|}$
and
$$
\mbox{Var}(\pi^{\rm AIHF}_1)=\frac{\pi^*_1(1-\pi^*_1)}{|\mathcal{D}|+|\mathcal{P}|} < \min\{\mbox{Var}(\pi^*_1(\widehat{R}_{\mathcal{D}})),\mbox{Var}(\pi^*_1(\widehat{R}_{\mathcal{P}}))\}
$$
Hence, the AIHF policy estimate has less variance than either the policy obtained from {\em only} demonstrations or preferences.

\medskip
To further illustrate,
suppose the ground truth is $R^*(\tau_1)=R^*(\tau_2)$ and in the dataset there are more demonstrations than preferences, i.e.  $|\mathcal{D}|\gg |\mathcal{P}|$:
\begin{align*}
    \#\{ \tau_1~\mbox{in}~ \mathcal{D}\}=\#\{ \tau_2~\mbox{in}~ \mathcal{D}\}=50, \quad 
  \#\{ \tau_1 \succ \tau_2 ~\mbox{in}~ \mathcal{P}\}&=6, \quad 
   \#\{ \tau_2 \succ \tau_1 ~\mbox{in}~ \mathcal{P}\}=4.
\end{align*}
With the given data,
$\pi^{\rm SFT}_1 = \pi^*_1(\widehat{R}_{\mathcal{D}})=
\frac{\#\{ \tau_1~\mbox{in}~ \mathcal{D}\}}{|\mathcal{D}|}=\frac{50}{100}$ and the solution to \eqref{eq:preference :policy} yields
$$
\pi^*_1(\widehat{R}_{\mathcal{P}})
= 
\frac{\exp \widehat{R}_{\mathcal{P}}(\tau_1)}{\exp \widehat{R}_{\mathcal{P}}(\tau_1)+\exp \widehat{R}_{\mathcal{P}}(\tau_2)}
=\frac{6}{10}.
$$
Hence,
$\pi_1^{\rm AIHF}=
\frac{100}{10+100}\pi^*_1(\widehat{R}_{\mathcal{D}})+
\frac{10}{10+100}\pi^*_1(\widehat{R}_{\mathcal{P}})=\frac{56}{110}$.
It follows from \eqref{eq:ref_optimal_RLHF} that: 
\begin{align*}
    \pi^{\rm RLHF}_1 &= \frac{\pi^{\rm SFT}_1 \exp \widehat{R}_{\mathcal{P}}(\tau_1)}{\pi^{\rm SFT}_1\exp \widehat{R}_{\mathcal{P}}(\tau_1)+\pi^{\rm SFT}_2\exp \widehat{r}_{\mathcal{P}}(\tau_2)}\\
    &
    =\frac{\exp \widehat{R}_{\mathcal{P}}(\tau_1)}{\exp \widehat{R}_{\mathcal{P}}(\tau_1)+\exp \widehat{R}_{\mathcal{P}}(\tau_2)}=\frac{6}{10}.
\end{align*}
}
In this example, 
the RLHF policy estimator is the furthest away from ground-truth, because it does not correctly use the information provided by the demonstration data which in this case happens by chance to be correct $\pi^{\rm SFT}_1 = \pi^*_1(\widehat{R}_{\mathcal{P}})=\frac{1}{2}$. 

\medskip
As a second example, again suppose the ground truth is $R^*(\tau_1)=R^*(\tau_2)$ and in the dataset there are more preferences than demonstrations, $|\mathcal{P}|\gg |\mathcal{D}|$:
\begin{align}
    \{\# \tau_1~\mbox{in}~ \mathcal{D}\} = 6, \quad \{\#\tau_2~\mbox{in}~ \mathcal{D}\}=4, \quad 
  \{\# \tau_1 \succ \tau_2 ~\mbox{in}~ \mathcal{P}\} = \{\# \tau_2 \succ \tau_1 ~\mbox{in}~ \mathcal{P}\} =50.
\end{align}
In this case, $\pi^{\rm SFT}_1 =\pi^*_1(\widehat{R}_{\mathcal{D}})= \frac{6}{10}$ and 
the solution to \eqref{eq:preference :policy} yields
$$
\pi^*_1(\widehat{R}_{\mathcal{P}})
= 
\frac{\exp \widehat{R}_{\mathcal{P}}(\tau_1)}{\exp \widehat{R}_{\mathcal{P}}(\tau_1)+\exp \widehat{R}_{\mathcal{P}}(\tau_2)}
=\frac{50}{100}.
$$
It follows from \eqref{eq:ref_optimal_RLHF} that: 
\begin{align*}
    \pi^{\rm RLHF}_1 &= \frac{\pi^{\rm SFT}_1 \exp \widehat{R}_{\mathcal{P}}(\tau_1)}{\pi^{\rm SFT}_1\exp \widehat{R}_{\mathcal{P}}(\tau_1)+\pi^{\rm SFT}_2\exp \widehat{R}_{\mathcal{P}}(\tau_2)}\\
    &
    =\frac{\pi_1^{SFT}}{\pi_1^{SFT}+\pi_2^{SFT}}=\frac{6}{10}.
\end{align*}
Hence,
$\pi_1^{\rm AIHF}=
\frac{10}{10+100}\pi^*_1(\widehat{R}_{\mathcal{D}})+
\frac{100}{10+100}\pi^*_1(\widehat{R}_{\mathcal{P}})=\frac{56}{110}$.
In this example, 
the RLHF policy estimator is again farthest from ground-truth, because it does not correctly dismiss the information provided by the demonstration data which is less informative than preference.

{\bf Example 2:} Let us use an illustrative example to show that RLHF method will result in significant data under-utilization when the demonstration coverage is limited. With $\beta=1$, assume that there are 50 actions, i.e. $A=\{1,2,\dots,50\}$ and each with a ground-truth reward defined by $R^*(\tau_i) = \frac{1}{\sigma\sqrt{2\pi}} e^{-\frac{(\frac{i}{50}-\mu)^2}{2\sigma^2}}$, where $\mu=0.5$ and $\sigma=2$. Assume we can sample demonstration and preference from the ground truth reward distribution: demonstrations are sampled from the multinomial distribution, while preferences are sampled from the BTL model.

In an extreme scenario, let demonstrations only include actions 1 through 45, i.e. $\mathcal{D} \cap \{45,46,\dots,50\}=\varnothing$, while preferences have full coverage across all actions. In the subsequent experiment, we initially sample 2000 demonstrations using the multinomial distribution $\pi^*_i=\frac{\exp R^*_i}{\sum^{j=45}_{j=1}\exp R^*_j}$, and obtain 200 preferences for each preference pair with $P(i \succ j)=\frac{\exp R^*_i}{\exp R^*_j+ \exp R^*_i}$. We then calculate the RLHF and AIHF policies as in in (\ref{eq:ref_optimal_RLHF}) and (\ref{eq:AIHF}) to obtain the result depicted in Figure \ref{fig:policy_distribution}:
%In this case, we will use numerical examples to show that RLHF suffers from this partial data coverage while AIHF does not.
\begin{figure}[H]
    \centering
    \includegraphics[width=0.47\textwidth]{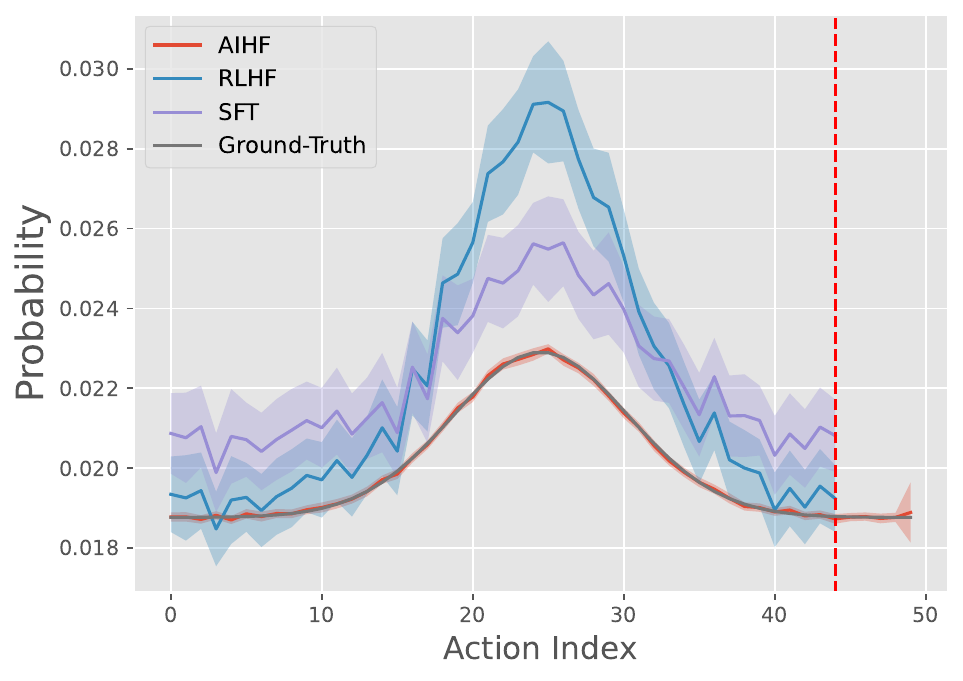}
    \caption{The optimal policy of RLHF, SFT, AIHF, and Ground-truth distribution. The left region of the red dotted line is included in the demonstration, while the right region is uncovered. We report the results with 100 random repeats.}
    \label{fig:policy_distribution}
\end{figure}

From the result shown in Figure \ref{fig:policy_distribution}, we demonstrate that both SFT and RLHF transfer the weight from uncovered action to covered actions when demonstration coverage is limited, as indicated by $\pi_{SFT}(\tau_i)=0, \tau_i \in \{45,46,\dots, 50\}$. Consequently, the weight of covered actions is significantly higher than the ground truth. However, this issue does not occur when jointly optimizing the demonstration and preference in the AIHF method.

\subsection{Proofs}\label{sec:convergence_result}

\subsubsection{Useful Lemmas}
\begin{assumption}[Ergodicity] \label{Assumption:Ergodicity_Markov_chain}
	For any policy $\pi$, assume the Markov chain with transition kernel $\mathcal{P}$ is irreducible and aperiodic under policy $\pi$. Then there exist constants $ \kappa > 0 $ and $ \rho \in (0, 1) $ such that 
	\begin{align}
		\sup_{s \in \mathcal{S}} ~ \| \mathbb{P}(s_t \in \cdot| s_0 = s, \pi) -  \mu_{\pi}(\cdot) \|_{TV} \leq \kappa \rho^t, \quad \forall ~ t \geq 0  \nonumber
	\end{align}
	where $ \| \cdot \|_{TV} $ is the total variation (TV) norm; $ \mu_{\pi} $ is the stationary state distribution under $ \pi $.
\end{assumption}
Assumption \ref{Assumption:Ergodicity_Markov_chain} assumes the Markov chain mixes at a geometric rate. It is a common assumption in the literature of RL, which holds for any time-homogeneous Markov chain with finite-state space or any uniformly ergodic Markov chain with general-state space.

\begin{assumption}\label{Assumption:reward_grad_bound}
	For any $s \in \mathcal{S}$, $a \in \mathcal{A}$ and any reward parameter $\theta$, the following holds: 
	\begin{subequations}
		\begin{align}
		\big \| \nabla_{\theta} r(s, a; \theta) \big  \| 	&  \leq L_r, \label{ineq:reward_grad_bound} \\
		\big \|	\nabla_{\theta} r(s, a; \theta_1)  - \nabla_{\theta} r(s, a; \theta_2)	\big \|	&  \leq  L_g \|  \theta_{1} - \theta_{2}	\|  \label{ineq:Lipschitz_smooth_reward}
		\end{align}
	\end{subequations}
	where $L_r$ and $L_g$ are positive constants.
\end{assumption}

\ref{Assumption:reward_grad_bound}, we next provide the following Lipschitz properties:
\begin{lemma} \label{lemma:Lipschitz_properties}
	Suppose Assumptions \ref{Assumption:Ergodicity_Markov_chain} - \ref{Assumption:reward_grad_bound} hold. For any reward parameter $\theta_1$ and $\theta_2$, the following results hold:
	\begin{subequations}
    \begin{align}
	    | Q^{soft}_{r_{\theta_1}, \pi_{\theta_1}}(s,a)  - Q^{soft}_{r_{\theta_2}, \pi_{\theta_2}}(s,a) |& \leq L_q \| \theta_{1} - \theta_2 \|, \quad \forall s \in \mathcal{S}, a \in \mathcal{A}  \label{ineq:soft_Q_Lipschitz} \\
		\| \nabla L(\theta_{1}) - \nabla L(\theta_{2}) \|& \leq L_c \| \theta_1 - \theta_2 \|  \label{ineq:objective_lipschitz_smooth}
	\end{align}
	\end{subequations}
	where $Q^{soft}_{r_{\theta}, \pi_{\theta}}(\cdot, \cdot)$ denotes the soft Q-function under the reward function $r(\cdot, \cdot; \theta)$ and the policy $\pi_{\theta}$. The positive constants $ L_q$ and $L_c$ are defined in Appendix \ref{proof:Lemma_Lipschitz_Properties}.
\end{lemma}

\subsubsection{Proof of Lemma \ref{Lemma:gradient}}\label{appendix:Lemma:gradient}

Before we proceed to the proof of Lemma \ref{Lemma:gradient}, we have the following remark.

{\bf Remark}: The KL-regularized MDP problem described by \eqref{eq:lower} and \eqref{eq:L3} has a closed-form solution:
\begin{align}\label{eq:opt:policy}
    \pi_{\theta}(a|s) = \frac{\pi^0(a|s)\exp(Q_{\theta}(s,a)/\beta)}{\sum_{\tilde{a}}\pi^0(\tilde{a}|s)\exp(Q_{\theta}(s,a)/\beta)},
\end{align}
where the corresponding value function $V_{\theta}$ and the Q-function $Q_{\theta}$ are defined as below:
%Define the soft Q and soft value functions for a given policy $\pi$ and a reward parameter $\theta$ by:
{\small
\begin{subequations}
    \begin{align}
      &V_{\theta}(s) := \mathbb{E}_{\tau\sim \pi_{\theta}} \bigg[ R(\tau;\theta)- \beta \sum_{t = 0}^{\infty} \gamma^t {D}_{\rm KL}\Big( \pi(\cdot | s_t) \| \pi^{0}(\cdot | s_t) \Big)\bigg| s_0 = s \bigg] \label{def:soft_Value_function} \\ 
      &Q_{\theta}(s,a) := r(s,a; \theta) + \gamma 
\mathbb{E}_{s^\prime \sim P(\cdot | s,a)} \big[ V_{\theta}(s') \big] \label{def:soft_Q_function}.
    \end{align}
\end{subequations}
}%

Further, assuming that $T=1$, i.e., $\tau = (s_0,a_0)$, and considering the LLM alignment problem as a sequence-level training problem (this is a popular simplification in language models, see, e.g., \cite{rafailov2023direct}), the closed-form expression of $\pi_{\theta}$ in \eqref{eq:opt:policy} can be reduced to:
\begin{align}\label{eq:dpo:pi}
    \pi_\theta(a|s)=\frac{\pi^0(a|s)\exp(\frac{1}{\beta}r(s,a;\theta))}{\sum_{a \in A}\big(\pi^0(a|s)\exp(\frac{1}{\beta}r(s,a;\theta))\big)}.
\end{align}

% \noindent \textbf{Proof of Lemma \ref{Lemma:gradient}.} 
Here, under a reward parameter $\theta$ and the corresponding optimal policy $\pi_{\theta}$ of \eqref{eq:opt:policy}.

Moreover, under a fixed reward parameter $\theta$, we have defined the optimal policy $\pi_{\theta}$ as below:
\begin{align}
    \pi_{\theta} := \arg \max_{\pi} ~\mathbb{E}_{\tau^{\rm A}\sim \pi} \bigg[ \sum_{t=0}^{\infty} \gamma^t \bigg( r(s_t, a_t; \theta) - \beta \mathcal{D}_{KL}\Big( \pi(\cdot | s_t) || \pi^0(\cdot | s_t) \Big) \bigg) \bigg]. \nonumber
\end{align}

According to \cite{uehara2023refined}, the optimal policy $\pi_{\theta}$ of \eqref{eq:L3} has the closed form expression as below:
\begin{align}
    \pi_{\theta}(a|s) = \frac{ \pi^0( a | s) \exp \big( \frac{Q_{\theta}(s,a;\theta)}{\beta} \big) }{\sum_{\tilde{a} \in \mathcal{A}} \pi^0( \tilde{a} | s) \exp \big( \frac{Q_{\theta}(s,\tilde{a};\theta)}{\beta} \big) }, \quad \forall s \in \mathcal{S}, a \in \mathcal{A}. \label{eq:optimal_policy_PPO}
\end{align}

Based on the closed form of $\pi_{\theta}$, we can also obtain the closed form of $V_{\theta}$ as following:
\begin{align}
    V_{\theta}(s) := \beta \log \Big( \sum_{a \in \mathcal{A}} \pi^0(a|s) \exp \big( \frac{Q_{\theta}(s,a)}{\beta} \big) \Big). \label{closed_form:optimal_soft_V}
\end{align}

Then we can re-write the demonstration loss $L_1(\theta)$ as below:
\begin{align}
    L_1(\theta) &= \mathbb{E}_{\tau^{\rm E} \sim \pi^{\rm E}} \bigg[ \sum_{t=0}^{\infty} \gamma^t \log \pi_{\theta}(a_t|s_t) \bigg] \nonumber \\
    &= \mathbb{E}_{\tau^{\rm E} \sim \pi^{\rm E}} \bigg[ \sum_{t=0}^{\infty} \gamma^t \log \bigg(  \frac{ \pi^0( a_t | s_t) \exp \big( \frac{Q_{\theta}(s_t,a_t)}{\beta} \big) }{\sum_{\tilde{a} \in \mathcal{A}} \pi^0( \tilde{a} | s_t) \exp \big( \frac{Q_{\theta}(s_t,\tilde{a})}{\beta} \big) }  \bigg) \bigg]  \nonumber \\
    &= \mathbb{E}_{\tau^{\rm E} \sim \pi^{\rm E}} \bigg[ \sum_{t=0}^{\infty} \gamma^t \bigg(   \log \Big( \pi^0( a_t | s_t) \exp \big( \frac{Q_{\theta}(s_t,a_t)}{\beta} \big) \Big)  - \log \Big( \sum_{\tilde{a} \in \mathcal{A}} \pi^0( \tilde{a} | s_t) \exp \big( \frac{Q_{\theta}(s_t,\tilde{a})}{\beta} \big) \Big) \bigg) \bigg]  \nonumber \\
    &= \mathbb{E}_{\tau^{\rm E} \sim \pi^{\rm E}} \bigg[ \sum_{t=0}^{\infty} \gamma^t \bigg(   \log \pi^0( a_t | s_t) +  \frac{Q_{\theta}(s_t,a_t)}{\beta}  - \log \Big( \sum_{\tilde{a} \in \mathcal{A}} \pi^0( \tilde{a} | s_t) \exp \big( \frac{Q_{\theta}(s_t,\tilde{a})}{\beta} \big) \Big) \bigg) \bigg]  \nonumber \\
    &= \frac{1}{\beta}\mathbb{E}_{\tau^{\rm E} \sim \pi^{\rm E}} \bigg[ \sum_{t=0}^{\infty} \gamma^t \bigg(  \beta \log \pi^0( a_t | s_t) +  Q_{\theta}(s_t,a_t) - \beta \log \Big( \sum_{\tilde{a} \in \mathcal{A}} \pi^0( \tilde{a} | s_t) \exp \big( \frac{Q_{\theta}(s_t,\tilde{a})}{\beta} \big) \Big) \bigg) \bigg]  \nonumber \\
     &= \frac{1}{\beta}\mathbb{E}_{\tau^{\rm E} \sim \pi^{\rm E}} \bigg[ \sum_{t=0}^{\infty} \gamma^t \bigg(  \beta \log \pi^0( a_t | s_t) +  Q_{\theta}(s_t,a_t) - V_{\theta}(s_t) \bigg) \bigg]  \label{eq:demonstration_obj_expression}
\end{align}

Then we can take gradient of $L_1(\theta)$ w.r.t. the reward parameter $\theta$, we have the following expression:
\begin{align}
    \nabla L_1(\theta) &:= \frac{1}{\beta}\mathbb{E}_{\tau^{\rm E} \sim \pi^{\rm E}} \bigg[ \sum_{t=0}^{\infty} \gamma^t \bigg( \nabla_{\theta} \beta \log \pi^0( a_t | s_t) + \nabla_{\theta} Q_{\theta}(s_t,a_t) - \nabla_{\theta} V_{\theta}(s_t) \bigg) \bigg]  \nonumber \\
    &= \frac{1}{\beta}\mathbb{E}_{\tau^{\rm E} \sim \pi^{\rm E}} \bigg[ \sum_{t=0}^{\infty} \gamma^t \bigg( \nabla_{\theta} Q_{\theta}(s_t,a_t) - \nabla_{\theta} V_{\theta}(s_t) \bigg) \bigg]  \nonumber \\
    &= \frac{1}{\beta}\mathbb{E}_{\tau^{\rm E} \sim \pi^{\rm E}} \bigg[ \sum_{t=0}^{\infty} \gamma^t \bigg( \nabla_{\theta} r(s_t, a_t;\theta) + \gamma \nabla_{\theta} V_{\theta}(s_{t+1}) - \nabla_{\theta} V_{\theta}(s_t) \bigg) \bigg]  \nonumber \\
    &= \frac{1}{\beta}\mathbb{E}_{\tau^{\rm E} \sim \pi^{\rm E}} \bigg[ \sum_{t=0}^{\infty} \gamma^t \nabla_{\theta} r(s_t, a_t;\theta) \bigg] - \frac{1}{\beta} \mathbb{E}_{s_0 \sim \rho} \bigg[ \nabla_{\theta} V_{\theta}(s_0) \bigg]  \label{eq:demo_obj_grad}
\end{align}

In order to calculate the expression of $\nabla L_1(\theta)$, we further derive the expression of $\nabla_{\theta} V_{\theta}(s_0)$:
\begin{align}
    \nabla_{\theta} V_{\theta}(s_0) &= \nabla_{\theta} \bigg( \beta \log \Big( \sum_{a \in \mathcal{A}} \pi^0(a|s_0) \exp \big( \frac{Q_{\theta}(s_0,a)}{\beta} \big) \Big) \bigg) \nonumber \\
    &=  \beta \sum_{a \in \mathcal{A}} \frac{\pi^0(a|s_0) \exp \big( \frac{Q_{\theta}(s_0,a)}{\beta} \big)}{\sum_{a \in \mathcal{A}} \pi^0(a|s_0) \exp \big( \frac{Q_{\theta}(s_0,a)}{\beta} \big)} \frac{\nabla_{\theta} Q_{\theta}(s,a)}{\beta}  \nonumber \\
    &= \mathbb{E}_{a \sim \pi_{\theta}(\cdot | s_0)} \Big[ \nabla_{\theta} Q_{\theta}(s_0,a) \Big] \nonumber \\
    &= \mathbb{E}_{a_0 \sim \pi_{\theta}(\cdot | s_0), s_1 \sim P(\cdot | s_0, a_0)} \Big[ \nabla_{\theta} r(s_0, a_0; \theta) + \gamma \nabla_{\theta} V_{\theta}(s_1) \Big] \nonumber \\
    &= \mathbb{E}_{\tau^{\rm A} \sim \pi_{\theta}} \bigg[ \sum_{t=0}^{\infty} \gamma^t \nabla_{\theta} r(s_t, a_t; \theta) \mid s_0 \bigg] \label{expression:V_grad}
\end{align}

By plugging \eqref{expression:V_grad} into \eqref{eq:demo_obj_grad}, we obtain the following expression:
\begin{align}
    \nabla L_1(\theta) = \frac{1}{\beta}\mathbb{E}_{\tau^{\rm E} \sim \pi^{\rm E}} \bigg[ \sum_{t=0}^{\infty} \gamma^t \nabla_{\theta} r(s_t, a_t;\theta) \bigg] - \frac{1}{\beta} \mathbb{E}_{\tau^{\rm A} \sim \pi_{\theta}} \bigg[ \sum_{t=0}^{\infty} \gamma^t \nabla_{\theta} r(s_t, a_t; \theta) \bigg] \label{eq:demo_gradient}
\end{align}

\subsubsection{Proof of Lemma \ref{lemma:Lipschitz_properties}} 
\label{proof:Lemma_Lipschitz_Properties}

To prove Lemma \ref{lemma:Lipschitz_properties}, we will prove the equality \eqref{ineq:soft_Q_Lipschitz} and the equality \eqref{ineq:objective_lipschitz_smooth} respectively. The constants $L_q$ and $L_c$ in Lemma \ref{lemma:Lipschitz_properties} has the expression: $$L_q := \frac{L_r}{1 - \gamma}, \quad L_c := \frac{2L_q L_r C_d \sqrt{ | \mathcal{S}| \cdot |\mathcal{A}| } }{1 - \gamma}  + \frac{2 L_g}{1 - \gamma}. $$

\subsubsection{Proof of Inequality \eqref{ineq:soft_Q_Lipschitz}}
In this subsection, we prove the inequality \eqref{ineq:soft_Q_Lipschitz} in Lemma \ref{lemma:Lipschitz_properties}.

 We show that $ Q^\soft_{r_\theta, \pi_\theta} $ has a bounded gradient with respect to any reward parameter $\theta$, then the inequality \eqref{ineq:soft_Q_Lipschitz} holds due to the mean value theorem. According to the soft Bellman equation, we have shown the explicit expression of $ \nabla_\theta Q^\soft_{r_\theta, \pi_\theta}(s,a) $ for any $s \in \mathcal{S}$ and $a \in \mathcal{A}$. Using this expression, we have the following series of relations:
	\begin{align}
		 \|  \nabla_\theta Q^\soft_{r_\theta, \pi_\theta}(s,a) \| 
        &=\bigg \|  \mathbb{E}_{a_0 \sim \pi_{\theta}(\cdot | s_0), s_1 \sim P(\cdot | s_0, a_0)} \Big[ \nabla_{\theta} r(s_0, a_0; \theta) + \gamma \nabla_{\theta} V_{\theta}(s_1) \Big]\bigg \|  \nonumber \\
        & \overset{(i)}{ = }  \bigg \| \ee_{\tau \sim \pi_{\theta}} \bigg[ \sum_{t \geq 0} \gamma^t \nabla_{\theta} r(s_t, a_t; \theta) ~ \bigg| (s_0, a_0) = (s, a) \bigg]  \bigg \| \nonumber \\
		&\overset{(ii)}{\leq} \ee_{\tau \sim \pi_{\theta}} \bigg[ \sum_{t \geq 0} \gamma^t  \bigg \| \nabla_{\theta} r(s_t, a_t; \theta) \bigg \| ~ \bigg| (s_0, a_0) = (s, a) \bigg] \nonumber \\
		& \overset{(iii)}{\leq } \ee_{\tau \sim \pi_{\theta}} \bigg[ \sum_{t \geq 0} \gamma^t  L_r ~ \bigg| (s_0, a_0) = (s, a) \bigg] \nonumber \\
		& = \frac{L_r}{1 - \gamma} \label{ineq:soft_Q_grad_bound}
	\end{align}
	where (i) is from the equality \eqref{expression:V_grad} in the proof of Lemma \ref{lemma:Lipschitz_properties}, (ii) follows Jensen's inequality and (iii) follows the inequality \eqref{ineq:reward_grad_bound} in Assumption \ref{Assumption:reward_grad_bound}. To complete this proof, we use the mean value theorem to show that 
	\begin{align}
	    |  Q^\soft_{r_{\theta_1}, \pi_{\theta_1}}(s,a)  - Q^\soft_{r_{\theta_2}, \pi_{\theta_2}}(s,a) | \leq \| \max_{\theta} \nabla_{\theta} Q^\soft_{r_{\theta}, \pi_{\theta}}(s,a) \| \cdot \| \theta_1 - \theta_2 \| \leq L_q \| \theta_{1} - \theta_2 \| 
	\end{align}
	where the last inequality follows \eqref{ineq:soft_Q_grad_bound} and we denote $L_q := \frac{L_r}{1 - \gamma} $. Therefore, we have proved the Lipschitz continuous inequality in  \eqref{ineq:soft_Q_Lipschitz}.

\subsubsection{Proof of Inequality \eqref{ineq:objective_lipschitz_smooth}}
In this section, we prove the inequality \eqref{ineq:objective_lipschitz_smooth} in Lemma \ref{lemma:Lipschitz_properties}.

According to Lemma \ref{lemma:Lipschitz_properties}, the gradient $\nabla L_1(\theta)$ is expressed as:
\begin{align}
    \nabla  L_1(\theta) = \mathbb{E}_{\tau \sim \pi^{\rm E}}\bigg[ \sum_{t \geq 0} \gamma^t \nabla_{\theta} r(s_t, a_t; \theta) \bigg] - \mathbb{E}_{\tau \sim \pi_{\theta}}\bigg[ \sum_{t \geq 0} \gamma^t \nabla_{\theta} r(s_t, a_t; \theta) \bigg]. \label{restate:gradient_expression}
\end{align}

	Using the above relation, we have  
	\begin{align}
		& \| \nabla L_1(\theta_{1}) - \nabla L_1(\theta_{2}) \|  \nonumber \\
		&\overset{(i)}{=} \bigg \| \bigg( \mathbb{E}_{\tau \sim \pi^{\rm E}}\bigg[ \sum_{t \geq 0} \gamma^t \nabla_{\theta} r(s_t, a_t; \theta_1) \bigg] - \mathbb{E}_{\tau \sim \pi_{\theta_1}}\bigg[ \sum_{t \geq 0} \gamma^t \nabla_{\theta} r(s_t, a_t; \theta_1) \bigg] \bigg)  - \nonumber \\
		& \quad \quad \bigg( \mathbb{E}_{\tau \sim \pi^{\rm E}}\bigg[ \sum_{t \geq 0} \gamma^t \nabla_{\theta} r(s_t, a_t; \theta_2) \bigg] - \mathbb{E}_{\tau \sim \pi_{\theta_2}}\bigg[ \sum_{t \geq 0} \gamma^t \nabla_{\theta} r(s_t, a_t; \theta_2) \bigg] \bigg)  \bigg  \|  \nonumber \\
		& \leq \underbrace{\bigg \|   \mathbb{E}_{\tau \sim \pi^{\rm E}}\bigg[ \sum_{t \geq 0} \gamma^t \nabla_{\theta} r(s_t, a_t; \theta_1) \bigg] -  \mathbb{E}_{\tau \sim \pi^{\rm E}}\bigg[ \sum_{t \geq 0} \gamma^t \nabla_{\theta} r(s_t, a_t; \theta_2) \bigg] \bigg \|}_{\rm :=term ~ A}  +  \nonumber \\
		& \quad \quad  \underbrace{\bigg \|   \mathbb{E}_{\tau \sim \pi_{\theta_1}}\bigg[ \sum_{t \geq 0} \gamma^t \nabla_{\theta} r(s_t, a_t; \theta_1) \bigg] -  \mathbb{E}_{\tau \sim \pi_{\theta_2}}\bigg[ \sum_{t \geq 0} \gamma^t \nabla_{\theta} r(s_t, a_t; \theta_2) \bigg] \bigg \|}_{\rm :=term ~ B}    \label{ineq:grad_difference_separate}
	\end{align}
	where (i) follows the exact gradient expression in equation \eqref{restate:gradient_expression}. Then we separately analyze term A and term B in \eqref{ineq:grad_difference_separate}. 
	
	For term A, it follows that
	\begin{align}
		&\bigg \|   \mathbb{E}_{\tau \sim \pi^{\rm E}}\bigg[ \sum_{t \geq 0} \gamma^t \nabla_{\theta} r(s_t, a_t; \theta_1) \bigg] -  \mathbb{E}_{\tau \sim \pi^{\rm E}}\bigg[ \sum_{t \geq 0} \gamma^t \nabla_{\theta} r(s_t, a_t; \theta_2) \bigg] \bigg \| \nonumber \\
		& \overset{(i)}{\leq} \ee_{\tau \sim \pi^{\rm E}} \bigg[ \sum_{t \geq 0} \gamma^t \big \|  \nabla_{\theta} r(s_t, a_t; \theta_1)  - \nabla_{\theta} r(s_t, a_t; \theta_2)  \big \| \bigg] \nonumber \\
		& \overset{(ii)}{\leq} \ee_{\tau \sim \pi^{\rm E}} \bigg[ \sum_{t \geq 0} \gamma^t L_g \| \theta_{1}  - \theta_{2} \| \bigg] \nonumber \\
		& = \frac{L_g}{1 - \gamma} \| \theta_{1}  -  \theta_{2} \| \label{ineq:gradient_diff_term_A}
	\end{align}
	where (i) follows Jensen's inequality and (ii) is from \eqref{ineq:Lipschitz_smooth_reward} in Assumption \ref{Assumption:reward_grad_bound}. 
	
	For the term B, it holds that 
	\begin{align}
		& \bigg \|   \mathbb{E}_{\tau \sim \pi_{\theta_1}}\bigg[ \sum_{t \geq 0} \gamma^t \nabla_{\theta} r(s_t, a_t; \theta_1) \bigg] -  \mathbb{E}_{\tau \sim \pi_{\theta_2}}\bigg[ \sum_{t \geq 0} \gamma^t \nabla_{\theta} r(s_t, a_t; \theta_2) \bigg] \bigg \|  \nonumber \\
		&\overset{(i)}{\leq} \bigg \|	\mathbb{E}_{\tau \sim \pi_{\theta_1}}\bigg[ \sum_{t \geq 0} \gamma^t \nabla_{\theta} r(s_t, a_t; \theta_1) \bigg]  -  \mathbb{E}_{\tau \sim \pi_{\theta_2}}\bigg[ \sum_{t \geq 0} \gamma^t \nabla_{\theta} r(s_t, a_t; \theta_1) \bigg]	\bigg \|    \nonumber \\
		&\quad +  \bigg \| \mathbb{E}_{\tau \sim \pi_{\theta_2}}\bigg[ \sum_{t \geq 0} \gamma^t \nabla_{\theta} r(s_t, a_t; \theta_1) \bigg] -  \mathbb{E}_{\tau \sim \pi_{\theta_2}} \bigg[ \sum_{t \geq 0} \gamma^t \nabla_{\theta} r(s_t, a_t; \theta_2) \bigg]  \bigg \| 	\nonumber \\
		&\overset{(ii)}{ \leq }  \frac{1}{1 - \gamma}   \bigg  \|  \mathbb{E}_{(s,a) \sim d(\cdot, \cdot; \pi_{\theta_1})}\bigg[ \nabla_{\theta} r(s_t, a_t; \theta_1) \bigg] - \mathbb{E}_{(s,a) \sim d(\cdot, \cdot;\pi_{\theta_2})}\bigg[ \nabla_{\theta} r(s_t, a_t; \theta_1) \bigg]   \bigg \|  \nonumber\\
		& \quad + \mathbb{E}_{\tau \sim \pi_{\theta_2}} \bigg[ \sum_{t \geq 0} \gamma^t  \bigg \|	\nabla_{\theta} r(s_t, a_t; \theta_1)  -  \nabla_{\theta} r(s_t, a_t; \theta_2) \bigg \|  \bigg]   \nonumber \\
		&\overset{(iii)}{\leq} \frac{1}{1 - \gamma}   \bigg  \| \sum_{s \in \mathcal{S}, a \in \mathcal{A}}  \nabla_{\theta} r(s_t, a_t; \theta_1) \bigg(d(s,a; \pi_{\theta_1}) - d(s,a; \pi_{\theta_2})\bigg) \bigg \|  + \mathbb{E}_{\tau \sim \pi_{\theta_2}} \bigg[ \sum_{k \geq 0} \gamma^k L_g \| \theta_{1}  -  \theta_{2} \|  \bigg] \nonumber \\
		&\overset{(iv)}{\leq} \frac{2L_r}{1 - \gamma} \| d(\cdot, \cdot; \pi_{\theta_1}) - d(\cdot, \cdot; \pi_{\theta_2}) \|_{TV}  + \frac{L_g}{1 - \gamma} \| \theta_{1}  -  \theta_{2} \|  \label{ineq:gradient_diff_term_B}
	\end{align}
	where (i) follows the triangle inequality, (ii) is from Jensen's inequality and the definition of the discounted state-action visitation measure $d(s,a;\pi) := (1-\gamma)\pi(a|s)\sum_{t\geq0} \gamma^t \mathcal{P}^{\pi}(s_t = s | s_0 \sim \eta)$; (iii) is from \eqref{ineq:Lipschitz_smooth_reward} in Assumption \ref{Assumption:reward_grad_bound};(iv) is from \eqref{ineq:reward_grad_bound} and the definition of the total variation norm.

    Consider the $L_2$ term: 
    \begin{align}
        L_2(\theta) := \mathbb{E}_{(\tau_i, \tau_w) \sim \pi^P} \left[ \log \left( \sigma \left( R(\tau_w; \theta) - R(\tau_i; \theta) \right) \right) \right] \nonumber
    \end{align}
    where \(\sigma(x)\) is sigmoid function defined by: $\sigma(x) = \frac{1}{1 + e^{-x}}$.
    We have
    \begin{align}
        &\nabla_\theta L_2(\theta) = \mathbb{E}_{(\tau_i, \tau_w) \sim \pi^P} \left[ (1 - \sigma(R(\tau_w; \theta) - R(\tau_i; \theta))) \cdot \left( \nabla_{\theta} R(\tau_w; \theta) - \nabla_{\theta} R(\tau_i; \theta) \right) \right]. \nonumber \\
        & =  \mathbb{E}_{(\tau_i, \tau_w) \sim \pi^P} \big[\left( \nabla_{\theta} R(\tau_w; \theta) - \nabla_{\theta} R(\tau_i; \theta) \right) -  \sigma(R(\tau_w; \theta) - R(\tau_i; \theta))(\nabla_{\theta} R(\tau_w; \theta) - \nabla_{\theta} R(\tau_i; \theta)) \big] \nonumber
    \end{align}
    Using the triangle inequality, we obtain the following equation:
    \begin{align}\label{eq:difference_of_L}
        &\|   \nabla L_2(\tau_w,\tau_l;\theta_{1}) - \nabla L_2(\tau_w,\tau_l;\theta_{2})  \|   \nonumber \\
        \leq& \underbrace{\bigg\| \mathbb{E}_{(\tau_i, \tau_w) \sim \pi^P} \big[ \left( \nabla_{\theta} R(\tau_w; \theta_1) - \nabla_{\theta} R(\tau_i; \theta_1) \right) - \left( \nabla_{\theta} R(\tau_w; \theta_2) - \nabla_{\theta} R(\tau_i; \theta_2) \right)\big] \bigg\|}_{\rm :=term ~ A} \nonumber \\
        &+ \underbrace{\bigg\| \mathbb{E}_{(\tau_i, \tau_w) \sim \pi^P} \big [ \sigma(R(\tau_w; \theta_1) - R(\tau_i; \theta_1))(\nabla_{\theta} R(\tau_w; \theta_1) - \nabla_{\theta} R(\tau_i; \theta_1))} \nonumber \\
        & \underbrace{ - \sigma(R(\tau_w; \theta_2) - R(\tau_i; \theta_2))(\nabla_{\theta} R(\tau_w; \theta_2) - \nabla_{\theta} R(\tau_i; \theta_2)) \big ]\bigg\|}_{\rm :=term ~ B} 
    \end{align}
    First we bound the term A of \eqref{eq:difference_of_L}
    \begin{align}
		&\rm term ~ A = \bigg \| \bigg(\bigg[ \sum_{t \geq 0} \gamma^t \nabla_{\theta} r(s^w_t, a^w_t; \theta_1) - \gamma^t \nabla_{\theta} r(s^l_t, a^l_t; \theta_1) \bigg] \bigg) - \bigg(\bigg[ \sum_{t \geq 0} \gamma^t \nabla_{\theta} r(s^w_t, a^w_t; \theta_2) -  \gamma^t \nabla_{\theta} r(s^l_t, a^l_t; \theta_2) \bigg] \bigg)   \bigg \| \nonumber \\
        &\leq \bigg \| \sum_{t \geq 0} \gamma^t \nabla_{\theta} r(s^w_t, a^w_t; \theta_1) - \gamma^t \nabla_{\theta} r(s^w_t, a^w_t; \theta_2) \bigg\| + \bigg \| \sum_{t \geq 0} \gamma^t \nabla_{\theta} r(s^l_t, a^l_t; \theta_1) -\gamma^t \nabla_{\theta} r(s^l_t, a^l_t; \theta_2) \bigg\| \nonumber \\
        &\leq  \frac{2 L_g}{1 - \gamma} \| \theta_{1}  -  \theta_{2} \|  
	\end{align}
    Then we bounded term B of \eqref{eq:difference_of_L}:
    \begin{align}
        &\rm term ~ B= \nonumber \\
        &{=} \bigg \| \sigma \bigg(\bigg[ \sum_{t \geq 0} \gamma^t  r(s^w_t, a^w_t; \theta_1) - \gamma^t  r(s^l_t, a^l_t; \theta_1) \bigg] \bigg) \bigg(\bigg[ \sum_{t \geq 0} \gamma^t \nabla_{\theta} r(s^w_t, a^w_t; \theta_1) -\gamma^t \nabla_{\theta} r(s^l_t, a^l_t; \theta_1) \bigg] \bigg) \nonumber  \\
        & -  \sigma \bigg(\bigg[ \sum_{t \geq 0} \gamma^t  r(s^w_t, a^w_t; \theta_2) - \gamma^t  r(s^l_t, a^l_t; \theta_2) \bigg] \bigg) \bigg(\bigg[ \sum_{t \geq 0} \gamma^t \nabla_{\theta} r(s^w_t, a^w_t; \theta_2) - \gamma^t \nabla_{\theta} r(s^l_t, a^l_t; \theta_2) \bigg] \bigg) \bigg \| \nonumber \\
        & = \bigg \| \sigma \bigg(\sum_{t \geq 0} \gamma^t r(s^w_t, a^w_t; \theta_1) - \gamma^t r(s^l_t, a^l_t; \theta_1) \bigg) \bigg(\sum_{t \geq 0} \gamma^t \nabla_\theta r(s^w_t, a^w_t; \theta_1)-\gamma^t \nabla_\theta r(s^l_t, a^l_t; \theta_1) \bigg) \nonumber \\
        &-\sigma \bigg(\sum_{t \geq 0} \gamma^t r(s^w_t, a^w_t; \theta_1) - \gamma^t r(s^l_t, a^l_t; \theta_1) \bigg) \bigg(\sum_{t \geq 0} \gamma^t \nabla_\theta r(s^w_t, a^w_t; \theta_2) - \gamma^t \nabla_\theta r(s^l_t, a^l_t; \theta_2) \bigg) \nonumber \\
        &+\sigma \bigg(\sum_{t \geq 0} \gamma^t r(s^w_t, a^w_t; \theta_1) - \gamma^t r(s^l_t, a^l_t; \theta_1) \bigg) \bigg(\sum_{t \geq 0} \gamma^t \nabla_\theta r(s^w_t, a^w_t; \theta_2)- \gamma^t \nabla_\theta r(s^l_t, a^l_t; \theta_2) \bigg) \nonumber \\
        &\sigma \bigg(\sum_{t \geq 0} \gamma^t r(s^w_t, a^w_t; \theta_2) - \gamma^t r(s^l_t, a^l_t; \theta_2) \bigg) \bigg(\sum_{t \geq 0} \gamma^t \nabla_\theta r(s^w_t, a^w_t; \theta_2) - \gamma^t \nabla_\theta r(s^l_t, a^l_t; \theta_2) \bigg) \bigg \| \nonumber \\
        & \leq \bigg \| \sigma \bigg(\sum_{t \geq 0} \gamma^t r(s^w_t, a^w_t; \theta_1) - \gamma^t r(s^l_t, a^l_t; \theta_1) \bigg) \bigg(\sum_{t \geq 0} \gamma^t \nabla_\theta r(s^w_t, a^w_t; \theta_1) - \gamma^t \nabla_\theta r(s^l_t, a^l_t; \theta_1) \nonumber \\
        &\quad + \sum_{t \geq 0} \gamma^t \nabla_\theta r(s^w_t, a^w_t; \theta_2) - \gamma^t \nabla_\theta r(s^l_t, a^l_t; \theta_2) \bigg) \bigg \| + \bigg \| \bigg( \sum_{t \geq 0} \gamma^t \nabla_\theta r(s^w_t, a^w_t; \theta_1) - \gamma^t \nabla_\theta r(s^l_t, a^l_t; \theta_1) \bigg) \nonumber \\ 
        &\quad \bigg [ \sigma \bigg(\sum_{t \geq 0} \gamma^t \nabla_\theta r(s^w_t, a^w_t; \theta_1) - \gamma^t \nabla_\theta r(s^l_t, a^l_t; \theta_1)\bigg) -\sigma \bigg(\sum_{t \geq 0} \gamma^t \nabla_\theta r(s^w_t, a^w_t; \theta_1) - \gamma^t \nabla_\theta r(s^l_t, a^l_t; \theta_1)\bigg) \bigg ] \bigg\| \nonumber \\
        &\leq \frac{2 L_g}{1 - \gamma} \| \theta_{1}  -  \theta_{2} \| + \frac{L_g}{1 - \gamma} \| \theta_{1}  -  \theta_{2} \|  = \frac{3 L_g}{1 - \gamma} \| \theta_{1}  -  \theta_{2} \|
    \end{align}

	Plugging the inequalities \eqref{ineq:gradient_diff_term_A}, \eqref{ineq:gradient_diff_term_B} to \eqref{ineq:grad_difference_separate}, it holds that 
	\begin{align}
		&\|   \nabla L(\theta_{1}) - \nabla L(\theta_{2})  \|   \nonumber \\
		& \leq \frac{2L_r}{1 - \gamma} \| d(\cdot, \cdot;\pi_{\theta_1}) - d(\cdot, \cdot;\pi_{\theta_2}) \|_{TV}  + \frac{6 L_g}{1 - \gamma} \| \theta_{1}  -  \theta_{2} \|   \nonumber \\
		& \overset{(i)}{\leq}  \frac{2L_r C_d}{1 - \gamma} \| Q^\soft_{r_{\theta_1}, \pi_{\theta_1}} - Q^\soft_{r_{\theta_2}, \pi_{\theta_2}}  \|  + \frac{6 L_g}{1 - \gamma} \| \theta_{1}  -  \theta_{2} \|   \nonumber \\
		& \overset{(ii)}{\leq}  \frac{2L_r C_d \sqrt{ | \mathcal{S}| \cdot |\mathcal{A}| } }{1 - \gamma} \| Q^\soft_{r_{\theta_1}, \pi_{\theta_1}} - Q^\soft_{r_{\theta_2}, \pi_{\theta_2}}   \|_{\infty}  + \frac{6 L_g}{1 - \gamma} \| \theta_{1}  -  \theta_{2} \|  \nonumber \\
		& \overset{(iii)}{\leq}  \bigg( \frac{2L_q L_r C_d \sqrt{ | \mathcal{S}| \cdot |\mathcal{A}| } }{1 - \gamma}  + \frac{6 L_g}{1 - \gamma}  \bigg) \| \theta_{1}  - \theta_{2}  \|.  \label{ineq:Lipschitz_Obj_derivation}
	\end{align}
	% Given the fact that $\pi_{\theta}$ is a Boltzmann policy parameterized by $Q^\soft_{r_{\theta}, \pi_{\theta}}$ where $\pi_{\theta}(a|s) \propto \exp(Q^\soft_{r_{\theta}, \pi_{\theta}}(s,a))$, we show the inequality (i) from the inequality \eqref{ineq:lipschitz_measure} in Lemma \ref{lemma:Lipschitz_visitation_measure}. Moreover, the inequality (ii) follows the equivalence relation between Frobenius norm and infinity norm and (iii) is from the inequality \eqref{ineq:soft_Q_Lipschitz} in Lemma \ref{lemma:Lipschitz_properties}. 
	
	Define the constant $L_c := \frac{2L_q L_r C_d \sqrt{ | \mathcal{S}| \cdot |\mathcal{A}| } }{1 - \gamma}  + \frac{5 L_g}{1 - \gamma}$, we have the following inequality:
	\begin{align}
	    \| \nabla L(\theta_{1}) - \nabla L(\theta_{2}) \| \leq L_c \| \theta_1 - \theta_2 \|. \nonumber
	\end{align}
	Therefore, we complete the proof of the inequality \eqref{ineq:objective_lipschitz_smooth} in Lemma \ref{lemma:Lipschitz_properties}. 

\subsection{Proof of Theorem \ref{theorem:main_convergence_results} }
\label{proof:main_convergence_theorem}

In this section, we prove \eqref{rate:lower_error} and \eqref{rate:upper_grad_norm} respectively, to show the convergence of the lower-level problem and the upper-level problem.
\label{proof:main_convergence_results}
\subsubsection{Proof of \eqref{rate:lower_error}}
In this proof, we first show the convergence of the lower-level variable $\{\pi_k\}_{k\geq0}$. Recall that we approximate the optimal policy $\pi_{\theta_k}$ by $\pi_{k+1}$ at each iteration $k$. We first analyze the approximation error between $\pi_{\theta_k}$ and $\pi_{k+1}$ as follows. For any $s \in \mathcal{S}$ and $a \in \mathcal{A}$, we have the following relation:
\begin{align}
    &\big| \log \big(\pi_{k+1}(a|s) \big) - \log \big(\pi_{\theta_k}(a|s) \big) \big| \nonumber \\
    &\overset{(i)}{=} \bigg| \log \bigg(\frac{\pi^0(a|s)\exp\big(Q^{\soft}_{r_{\theta_k}, \pi_k}(s,a)\big)}{\sum_{\tilde{a}} \exp \pi^0(\tilde{a}|s) \big(Q^{\soft}_{r_{\theta_k},\pi_k}(s,\tilde{a})\big)} \bigg) - \log \bigg(\frac{\pi^0(a|s) \exp\big(Q^{\soft}_{r_{\theta_k},\pi_{\theta_k}}(s,a)\big)}{\sum_{\tilde{a}} \pi^0(\tilde{a}|s) \exp\big(Q^{\soft}_{r_{\theta_k},\pi_{\theta_k}}(s,\tilde{a})\big)} \bigg) \bigg|  \nonumber \\
    &\overset{(ii)}{\leq} \big| Q^{\soft}_{r_{\theta_k},\pi_k}(s,a) - Q^{\soft}_{r_{\theta_k},\pi_{\theta_k}}(s,a) \big| + \bigg| \log\bigg( \sum_{\tilde{a}}  \pi^0(\tilde{a}|s) \exp\big(Q^{\soft}_{r_{\theta_k},\pi_k}(s,\tilde{a})\big) \bigg) \nonumber \\
    & \quad -\log \bigg( \sum_{\tilde{a}}  \pi^0(\tilde{a}|s) \exp\big(Q^{\soft}_{r_{\theta_k},\pi_{\theta_k}}(s,\tilde{a})\big) \bigg) \bigg| \label{ineq:log_policy_gap}
\end{align}

where (i) follows \eqref{eq:opt:policy}; (ii) is by the triangle inequality. We further analyze the second term in \eqref{ineq:log_policy_gap}. 
	
	We first denote the operator $\log(\| w \exp(v) \|_1) := \log(\| \sum_{\tilde{a} \in \mathcal{A}} w \exp(v_{\tilde{a}}) \|_1)$, where the vector $w,v \in \mathbb{R}^{|\mathcal{A}|}$ and $v = [v_1, v_2, \cdots, v_{|\mathcal{A}|}],w = [w_1, w_2, \cdots, w_{|\mathcal{A}|}]$. Then for any $v^\prime, v^{\prime\prime} \in \mathbb{R}^{|\mathcal{A}|}$, we have the following relation:
	\begin{align}
	    \big| \log\big( \|w^{\prime} \exp(v^\prime) \|_1 \big) - \log\big( \|w^{\prime\prime} \exp(v^{\prime\prime}) \|_1 \big) &\overset{(i)}{=} \big \langle v^\prime - v^{\prime\prime}, \nabla_{v} \log\big( \| w \exp(v) \|_1 \big) |_{v = v^c}  \big \rangle \nonumber \\
	    &\leq \|  v^\prime - v^{\prime\prime}  \|_{\infty} \cdot \|  \nabla_{v} \log\big( \| w \exp(v) \|_1 \big) |_{v = v^c}  \|_1 \nonumber \\
	    &\overset{(ii)}{=} \|  v^\prime - v^{\prime\prime}  \|_{\infty} \label{ineq:Lipscitz_log_exp_operator}
	\end{align}
	where (i) follows the mean value theorem and $v_c$ is a convex combination of $v^\prime$ and $v^{\prime\prime}$; (ii) follows the following equalities:
    \begin{align}
	    [\nabla_{v} \log\big( \| w \exp(v) \|_1 \big)]_i = \frac{w_i \exp(v_i)}{\sum_{1 \leq a \leq |\mathcal{A}|}w_a \exp(v_a)}, \quad  \|\nabla_{v} \log\big( \| w \exp(v) \|_1 \big)\|_1 = 1, \quad \forall v \in \mathbb{R}^{|\mathcal{A}|}. \nonumber
	\end{align}
	
	Through plugging \eqref{ineq:Lipscitz_log_exp_operator} into \eqref{ineq:log_policy_gap}, it holds that 
	\begin{align}
	    &\big| \log \big(\pi_{k+1}(a|s) \big) - \log \big(\pi_{\theta_k}(a|s) \big) \big| \nonumber \\
	    &\leq \big| Q^{\soft}_{r_{\theta_k},\pi_k}(s,a) - Q^{\soft}_{r_{\theta_k},\pi_{\theta_k}}(s,a) \big| + \max_{\tilde{a} \in \mathcal{A}} \big| Q^{\soft}_{r_{\theta_k},\pi_k}(s,\tilde{a}) - Q^{\soft}_{r_{\theta_k},\pi_{\theta_k}}(s,\tilde{a}) \big| \label{ineq:log_pi_soft_Q}
	\end{align}
	Taking the infinity norm over $\mathbb{R}^{|\mathcal{S}| \cdot |\mathcal{A}|}$, the following result holds:
	\begin{align}
	   \| \log \pi_{k+1} - \log \pi_{\theta_k} \|_{\infty} \leq 2 \| Q^{\soft}_{r_{\theta_k},\pi_k} - Q^{\soft}_{r_{\theta_k},\pi_{\theta_k}} \|_{\infty} \label{ineq:infty_policy_gap}
	\end{align}
	where $\| \log \pi_{k+1} - \log \pi_{\theta_k} \|_{\infty} = \max_{s \in \mathcal{S}, a \in \mathcal{A}} | \log \pi_{k+1}(a|s) - \log \pi_{\theta_k}(a|s) |$ and $ \| Q^{\soft}_{r_{\theta_k},\pi_k} - Q^{\soft}_{r_{\theta_k},\pi_{\theta_k}} \|_{\infty} = \max_{s \in \mathcal{S}, a \in \mathcal{A}} | Q^{\soft}_{r_{\theta_k},\pi_k}(s,a) - Q^{\soft}_{r_{\theta_k},\pi_{\theta_k}}(s,a) |$.

 Based on the inequality \eqref{ineq:infty_policy_gap}, we analyze $\| Q^{\soft}_{r_{\theta_k},\pi_k} - Q^{\soft}_{r_{\theta_k},\pi_{\theta_k}} \|_{\infty}$ to show the convergence of the policy estimates. It leads to the following analysis:
	\begin{align}
		& \|  Q^{\soft}_{r_{\theta_k},\pi_k} - Q^{\soft}_{r_{\theta_k},\pi_{\theta_k}} \|_{\infty} \nonumber \\
		&=  \|  Q_{r_{\theta_k}, \pi_{k}}^\soft -  Q^\soft_{r_{\theta_k}, \pi_{\theta_k}} +  Q^\soft_{r_{\theta_{k-1}}, \pi_{\theta_{k-1}}} -  Q^\soft_{r_{\theta_{k-1}}, \pi_{\theta_{k-1}}}  +  Q_{r_{\theta_{k-1}}, \pi_{k}}^\soft  -  Q_{r_{\theta_{k-1}}, \pi_{k}}^\soft \|_{\infty}  \nonumber  \\
		&\leq  \| Q^\soft_{r_{\theta_k}, \pi_{\theta_k}}  -  Q^\soft_{r_{\theta_{k-1}}, \pi_{\theta_{k-1}}}  \|_{\infty}  + \|  Q_{r_{\theta_{k-1}}, \pi_{k}}^\soft -  Q^\soft_{r_{\theta_{k-1}}, \pi_{\theta_{k-1}}}  \|_{\infty} + \| Q_{r_{\theta_k}, \pi_{k}}^\soft -  Q_{r_{\theta_{k-1}}, \pi_{k}}^\soft \|_{\infty}		\nonumber \\
		&\overset{(i)}{\leq}  L_q \| \theta_{k} - \theta_{k-1} \|  +  \|  Q_{r_{\theta_{k-1}}, \pi_{k}}^\soft -  Q^\soft_{r_{\theta_{k-1}}, \pi_{\theta_{k-1}}}  \|_{\infty} + \| Q_{r_{\theta_k}, \pi_{k}}^\soft -  Q_{r_{\theta_{k-1}}, \pi_{k}}^\soft \|_{\infty} \nonumber \\
		&\overset{(ii)}{\leq}  \|  Q_{r_{\theta_{k-1}}, \pi_{k}}^\soft -  Q^\soft_{r_{\theta_{k-1}}, \pi_{\theta_{k-1}}}  \|_{\infty} + 2 L_q \| \theta_k - \theta_{k-1} \|  \label{bound:soft_Q_difference}
	\end{align}
	where (i) is from \eqref{ineq:soft_Q_Lipschitz} in Lemma \ref{lemma:Lipschitz_properties}; (ii) follows \eqref{ineq:soft_Q_Lipschitz}. Based on \eqref{bound:soft_Q_difference}, we further analyze the two terms in \eqref{bound:soft_Q_difference} as below.

 Recall we have the “soft” Bellman operator expressed as below: \begin{align}
		\mathcal{T}_{\theta}(Q)(s,a) = r(s,a; \theta) + \gamma \mathbb{E}_{s^\prime \sim P(\cdot | s^\prime, a^\prime)}\bigg[ \log \left( \sum_{a^\prime} \pi^0(a^\prime|s^\prime)\exp\big(Q(s^\prime, a^\prime) \big) \right) \bigg] \label{operator:soft_bellman}
	\end{align}
	
	According to the soft Bellman operator, it holds that
	\begin{align}
		Q^\soft_{r_{\theta_k}, \pi_{k+1}}(s,a) &= r(s,a; \theta_k) + \gamma \ee_{s^\prime \sim \mathcal{P}(\cdot | s,a)}[ V^\soft_{r_{\theta_k}, \pi_{k+1}}(s^\prime) ] \nonumber \\
		&= r(s,a; \theta_k) + \gamma \ee_{s^\prime \sim \mathcal{P}(\cdot | s,a), a^\prime \sim \pi_{k+1}(\cdot | s^\prime)}[ -\frac{\log \pi_{k+1}(a^\prime | s^\prime)}{\log \pi_{0}(a^\prime | s^\prime)} +  Q^\soft_{r_{\theta_k}, \pi_{k+1}}(s^\prime,a^\prime) ] \nonumber \\
		&\overset{(i)}{\geq} r(s,a; \theta_k) + \gamma \ee_{s^\prime \sim \mathcal{P}(\cdot | s,a), a^\prime \sim \pi_{k+1}(\cdot | s^\prime)}[ -\frac{\log \pi_{k+1}(a^\prime | s^\prime)}{\log \pi_{0}(a^\prime | s^\prime)} +  Q^\soft_{r_{\theta_k}, \pi_{k}}(s^\prime,a^\prime) ] \nonumber \\
		&\overset{(ii)}{=} r(s,a; \theta_k) + \gamma \ee_{s^\prime \sim \mathcal{P}(\cdot | s,a)} \left[ \log \bigg( \sum_{a^\prime} \pi^0(a^\prime|s^\prime)\exp\big(Q^\soft_{r_{\theta_k}, \pi_{k}}(s^\prime,a^\prime) \big) \bigg) \right] \nonumber \\
		&\overset{(iii)}{=} \mathcal{T}_{\theta_k}(Q^\soft_{r_{\theta_k}, \pi_{k}})(s,a)  \label{ineq:policy_improvement_inequality}
	\end{align}
 
where (i) follows the policy improvement result(ii) follows the definition $ \pi_{k+1}(a | s):= \frac{\pi^0(a|s)\exp\big(Q^\soft_{r_{\theta_k}, \pi_{k}}(s, a) \big)}{\sum_{\tilde{a}}\pi^0(\tilde{a|s}) \exp\big(Q^\soft_{r_{\theta_k}, \pi_{k}}(s, \tilde{a}) \big)}$ (iii) follows the definition of the soft Bellman operator in \eqref{operator:soft_bellman}.

For any $s \in \mathcal{S}$ and $a \in \mathcal{A}$, it holds that 
	\begin{align}
		0 \overset{(i)}{\leq} Q^\soft_{r_{\theta_k}, \pi_{\theta_k}}(s, a) - Q^\soft_{r_{\theta_k}, \pi_{k+1}}(s,a) \overset{(ii)}{\leq} Q^\soft_{r_{\theta_k}, \pi_{\theta_k}}(s, a) - \mathcal{T}_{\theta_k}(Q^\soft_{r_{\theta_k}, \pi_{k}})(s,a) \label{ineq:policy_gap}
	\end{align} 
	where (i) is due to the fact that $\pi_{\theta_k}$ is the optimal policy under reward parameter $\theta_k$; (ii) is from \eqref{ineq:policy_improvement_inequality}.
	
	Hence, it further leads to \begin{align}
		\| Q^\soft_{r_{\theta_k}, \pi_{\theta_k}} - Q^\soft_{r_{\theta_k}, \pi_{k+1}} \|_{\infty} &\overset{(i)}{\leq} \| Q^\soft_{r_{\theta_k}, \pi_{\theta_k}} - \mathcal{T}_{\theta_k}(Q^\soft_{r_{\theta_k}, \pi_{k}}) \|_{\infty} \nonumber \\
		&\overset{(ii)}{=} \| \mathcal{T}_{\theta_k}( Q^\soft_{r_{\theta_k}, \pi_{\theta_k}}) - \mathcal{T}_{\theta_k}(Q^\soft_{r_{\theta_k}, \pi_{k}}) \|_{\infty} \nonumber \\
		&\overset{(iii)}{\leq} \gamma \|  Q^\soft_{r_{\theta_k}, \pi_{\theta_k}} - Q^\soft_{r_{\theta_k}, \pi_{k}}  \|_{\infty} \label{contraction:policy_update}
	\end{align}
	where (i) is from \eqref{ineq:policy_gap}; (ii) is from the fixed-point property in \eqref{fixed_point:soft_Bellman}; (iii) is from the contraction property in \eqref{ineq:soft_Bellman_contraction}. Therefore, we have the following result:
	\begin{align}
	    & \|  Q^{\soft}_{r_{\theta_k},\pi_k} - Q^{\soft}_{r_{\theta_k},\pi_{\theta_k}} \|_{\infty} \nonumber \\
		&\overset{(i)}{\leq}  \|  Q_{r_{\theta_{k-1}}, \pi_{k}}^\soft -  Q^\soft_{r_{\theta_{k-1}}, \pi_{\theta_{k-1}}}  \|_{\infty} + 2 L_q \| \theta_k - \theta_{k-1} \| \nonumber \\
		&\overset{(ii)}{\leq} \gamma \|  Q_{r_{\theta_{k-1}}, \pi_{k-1}}^\soft -  Q^\soft_{r_{\theta_{k-1}}, \pi_{\theta_{k-1}}}  \|_{\infty} + 2 L_q \| \theta_k - \theta_{k-1} \| \label{constraction:soft_q_update}
	\end{align}
	where (i) is from \eqref{bound:soft_Q_difference}; (ii) is from \eqref{contraction:policy_update}.
	
	To show the convergence of the soft Q-function based on \eqref{constraction:soft_q_update}, we further analyze the error between the reward parameters $\theta_k$ and $\theta_{k-1}$. 
    Recall in Alg.\ref{alg:offline_ML_IRL}, the updates in reward parameters \eqref{eq:sto_gradient_expression}:	
	\begin{align}
		& \theta_k = \theta_{k-1} + \alpha g_{k-1} \nonumber 
	\end{align}
 
 where we denote $\tau = \{ (s_t, a_t) \}_{t = 0}^{\infty}$, $h(\theta, \tau) := \sum_{t \geq 0} \gamma^t \nabla_{\theta} r(s_t,a_t; \theta) $ and $g_{k-1}$ is the stochastic gradient estimator at iteration $k-1$. Here, $\tau^E_{k-1}$ denotes the trajectory sampled from the expert's dataset $D$ at iteration $k-1$ and $\tau^A_{k-1}$ denotes the trajectory sampled from the agent's policy $\pi_k$ at time $k-1$,$\tau_w,\tau_i$ denote the trajectory sampled from the preference dataset. Then according to the inequality \eqref{ineq:reward_grad_bound} in Assumption \ref{Assumption:reward_grad_bound}, we could show that 
 \begin{align}
		\|  g_{k-1}  &\| \leq \| h(\theta_{k - 1}, \tau^E_{k-1}) -  h(\theta_{k - 1}, \tau^A_{k-1})\| +\| h(\theta_{k - 1}, \tau^W_{k-1}) -  h(\theta_{k - 1}, \tau^L_{k-1}) \|  \nonumber \\
        &\leq \frac{2 L_r}{1 - \gamma}+\frac{2 L_r}{1 - \gamma} = 4L_q \label{bound:upper_grad}
	\end{align}
	where the last equality follows the fact that we have defined the constant $L_q := \frac{L_r}{1 - \gamma}$. Then we could further show that
	\begin{align}
	    & \|  Q^{\soft}_{r_{\theta_k},\pi_k} - Q^{\soft}_{r_{\theta_k},\pi_{\theta_k}} \|_{\infty} \nonumber \\
		&\overset{(i)}{\leq} \gamma \|  Q_{r_{\theta_{k-1}}, \pi_{k-1}}^\soft -  Q^\soft_{r_{\theta_{k-1}}, \pi_{\theta_{k-1}}}  \|_{\infty} + 4 L_q \| \theta_k - \theta_{k-1} \| \nonumber \\
		&\overset{(ii)}{=} \gamma \|  Q_{r_{\theta_{k-1}}, \pi_{k-1}}^\soft -  Q^\soft_{r_{\theta_{k-1}}, \pi_{\theta_{k-1}}}  \|_{\infty} + 4 \alpha L_q \| g_{k-1} \| \nonumber \\
		&\overset{(iii)}{\leq} \gamma \|  Q_{r_{\theta_{k-1}}, \pi_{k-1}}^\soft -  Q^\soft_{r_{\theta_{k-1}}, \pi_{\theta_{k-1}}}  \|_{\infty} + 8\alpha L_q^2 \label{contraction_bound:optimal_gap_soft_Q}
	\end{align}
	where (i) is from \eqref{constraction:soft_q_update}; (ii) follows the reward update scheme; (iii) is from \eqref{bound:upper_grad}. 
	
	Summing the inequality \eqref{contraction_bound:optimal_gap_soft_Q} from $k = 1$ to $k = K$, it holds that \begin{align}
		\sum_{k = 1}^{K} \| Q_{r_{\theta_k}, \pi_{k}}^\soft - Q^\soft_{r_{\theta_k}, \pi_{\theta_k}} \|_{\infty} \leq \gamma \sum_{k = 0}^{K - 1} \| Q_{r_{\theta_k}, \pi_{k}}^\soft - Q^\soft_{r_{\theta_k}, \pi_{\theta_k}} \|_{\infty} +  8\alpha K L_q^2  \label{bound:sum_lower_error}
	\end{align}
	Rearranging the inequality \eqref{bound:sum_lower_error} and divided \eqref{bound:sum_lower_error} by $K$ on both sides, it holds that
	\begin{align}
		& \frac{1 - \gamma}{K} \sum_{k = 1}^{K} \| Q_{r_{\theta_k}, \pi_{k}}^\soft - Q^\soft_{r_{\theta_k}, \pi_{\theta_k}} \|_{\infty} \leq \frac{\gamma}{K} \bigg( \| Q_{r_{\theta_0}, \pi_{0}}^\soft - Q^\soft_{r_{\theta_0}, \pi_{\theta_0}} \|_{\infty} -  \| Q_{r_{\theta_K}, \pi_{K}}^\soft - Q^\soft_{r_{\theta_K}, \pi_{\theta_K}} \|_{\infty} \bigg) +  8\alpha L_q^2  \label{bound:lower_rate}
	\end{align}
	Dividing the constant $ 1 - \gamma $ on both sides of \eqref{bound:lower_rate}, it holds that 
	\begin{align}
		\frac{1}{K} \sum_{k = 1}^{K} \| Q_{r_{\theta_k}, \pi_{k}}^\soft - Q^\soft_{r_{\theta_k}, \pi_{\theta_k}} \|_{\infty} \leq \frac{\gamma C_0}{K(1 - \gamma)}  + \frac{8 L_q^2}{1 - \gamma} \alpha \nonumber
	\end{align}
	where we denote $C_0 := \| Q_{r_{\theta_0}, \pi_{0}}^\soft - Q^\soft_{r_{\theta_0}, \pi_{\theta_0}} \|_{\infty} $. We could also write the inequality above as 
	\begin{align}
		& \frac{1}{K} \sum_{k = 0}^{K-1} \| Q_{r_{\theta_k}, \pi_{k}}^\soft - Q^\soft_{r_{\theta_k}, \pi_{\theta_k}} \|_{\infty} \nonumber \\
		& \leq \frac{\gamma C_0}{K(1 - \gamma)} + \frac{C_0}{K} - \frac{\| Q_{r_{\theta_K}, \pi_{K}}^\soft - Q^\soft_{r_{\theta_K}, \pi_{\theta_K}} \|_{\infty}}{K}  + \frac{8 L_q^2}{1 - \gamma} \alpha \nonumber \\
		& \leq \frac{C_0}{K(1 - \gamma)} +  \frac{8 L_q^2}{1 - \gamma} \alpha. \nonumber
	\end{align}
	
	Recall the stepsize is defined as $\alpha = \frac{\alpha_0}{K^\sigma}$ where $\sigma>0$. Then we have the following result:
	\begin{align}
		\frac{1}{K} \sum_{k = 0}^{K-1} \| Q_{r_{\theta_k}, \pi_{k}}^\soft - Q^\soft_{r_{\theta_k}, \pi_{\theta_k}} \|_{\infty}   = \mathcal{O}(K^{-1}) + \mathcal{O}(K^{-\sigma}). \label{convergence_rate:soft_Q_gap}
	\end{align}
	With the inequality \eqref{ineq:infty_policy_gap}, it follows that
	\begin{align}
	   \frac{1}{K} \sum_{k = 0}^{K-1} \| \log \pi_{k+1} - \log \pi_{\theta_k} \|_{\infty} \leq \frac{2}{K} \sum_{k = 0}^{K-1} \| Q^{\soft}_{r_{\theta_k},\pi_k} - Q^{\soft}_{r_{\theta_k},\pi_{\theta_k}} \|_{\infty} = \mathcal{O}(K^{-1}) + \mathcal{O}(K^{-\sigma}). \nonumber
	\end{align}
	Therefore, we complete the proof of \eqref{rate:lower_error} in Theorem \ref{theorem:main_convergence_results}.
\subsubsection{Proof of \eqref{rate:upper_grad_norm}}

In this part, we prove the convergence of reward parameters $\{\theta_k\}_{k \geq 0}$.
    
	We have the following result of the objective function $L(\theta)$:
	\begin{align}
		L(\theta_{k+1}) &\overset{(i)}{ \geq } L(\theta_{k}) + \langle \nabla L(\theta_{k}), \theta_{k+1} - \theta_{k}  \rangle - \frac{L_c}{2} \| \theta_{k+1}  - \theta_{k} \|^2 \nonumber \\
		&\overset{(ii)}{=} L(\theta_{k}) + \alpha \langle \nabla L(\theta_{k}), g_k \rangle - \frac{L_c \alpha^2}{2} \| g_k \|^2 \nonumber \\
		&=  L(\theta_{k}) + \alpha \langle \nabla L(\theta_{k}), g_k - \nabla L(\theta_{k}) \rangle + \alpha \|  \nabla L(\theta_{k}) \|^2  - \frac{L_c \alpha^2}{2} \| g_k \|^2 \nonumber \\
		&\overset{(iii)}{ \geq } L(\theta_{k}) + \alpha \langle \nabla L(\theta_{k}), g_k - \nabla L(\theta_{k}) \rangle + \alpha \|  \nabla L(\theta_{k}) \|^2  - 8L_c L_q^2 \alpha^2 \label{ineq:upper_grad_ascent}
	\end{align}
	where (i) is from the Lipschitz smooth property in \eqref{ineq:objective_lipschitz_smooth} of Lemma \ref{lemma:Lipschitz_properties}; (ii) follows the update scheme \eqref{eq:sto_gradient_expression}; (iii) is from constant bound in \eqref{bound:upper_grad}.  
	Taking an expectation over the both sides of \eqref{ineq:upper_grad_ascent}, it holds that
	\begin{align}
		& \mathbb{E} \left[ L(\theta_{k+1}) \right] \nonumber \\
		&\geq \mathbb{E} \left[ L(\theta_{k}) \right] + \alpha \mathbb{E} \bigg[ \langle \nabla L(\theta_{k}) , g_k - \nabla L(\theta_{k}) \rangle \bigg] + \alpha \mathbb{E} \bigg[ \|  \nabla L(\theta_{k}) \|^2 \bigg] - 8L_c L_q^2 \alpha^2  \nonumber \\ 
		&= \mathbb{E} \left[ L(\theta_{k}) \right] + \alpha \mathbb{E} \bigg[ \langle \nabla L(\theta_{k}), \mathbb{E} \big[g_k - \nabla L(\theta_{k}) \big | \theta_k] \rangle \bigg] + \alpha \mathbb{E} \bigg[ \|  \nabla L(\theta_{k}) \|^2 \bigg] - 8L_c L_q^2 \alpha^2  \nonumber \\
		&{=}  \mathbb{E} \left[ L(\theta_{k}) \right] + \alpha \ee \bigg[ \bigg \langle  \nabla L(\theta_{k}), \ee_{\tau \sim \pi_{\theta_{k}}}\bigg[ \sum_{t \geq 0} \gamma^t \nabla_{\theta} r(s_t, a_t; \theta_{k}) \bigg]  -  \ee_{\tau \sim \pi_{k+1}}\bigg[ \sum_{t \geq 0} \gamma^t \nabla_{\theta} r(s_t, a_t; \theta_{k}) \bigg] \nonumber \\
        &+ \ee_{(\tau_l \prec \tau_w)\sim \pi^P}\bigg[ \sum_{t \geq 0} (1-\sigma(\gamma^t r(s^w_t, a^w_t; \theta_{k}) - \gamma^t  r(s^l_t, a^l_t; \theta_{k}) )(\gamma^t \nabla_{\theta} r(s^w_t, a^w_t; \theta_{k}) - \gamma^t \nabla_{\theta} r(s^l_t, a^l_t; \theta_{k})) \bigg] \bigg  \rangle \bigg] \nonumber \\
        & + \alpha \mathbb{E} \bigg[ \|  \nabla L(\theta_{k}) \|^2 \bigg] - 8L_c L_q^2 \alpha^2 \nonumber  \\
		&\overset{(i)}{ \geq }  \mathbb{E} \left[ L(\theta_{k}) \right]  - 4\alpha L_q \underbrace{ \ee \bigg [ \bigg \| \ee_{\tau \sim \pi_{\theta_{k}}}\bigg[ \sum_{t \geq 0} \gamma^t \nabla_{\theta} r(s_t, a_t; \theta_{k}) \bigg]  -  \ee_{\tau \sim \pi_{k+1}}\bigg[ \sum_{t \geq 0} \gamma^t \nabla_{\theta} r(s_t, a_t; \theta_{k}) }_{\rm term~A} \bigg] \bigg \| \nonumber \\
        & + \alpha \mathbb{E} \bigg[ \|  \nabla L(\theta_{k}) \|^2 \bigg] - 8L_c L_q^2 \alpha^2 \label{bound:objective_gradient_ascent}
	\end{align}
     (i) is due to the fact that $\| \nabla L(\theta) \| \leq 4L_q$ and $\ee[g_{k,2}-\nabla_\theta L_2(\theta_k)|\theta_k]=0$.
	
	Then we further analyze the term A as below:
	\begin{align}
		& \ee \left[ \bigg \| \ee_{\tau \sim \pi_{\theta_{k}}}\bigg[ \sum_{t \geq 0} \gamma^t \nabla_{\theta} r(s_t, a_t; \theta_{k}) \bigg]  -  \ee_{\tau \sim \pi_{k+1}}\bigg[ \sum_{t \geq 0} \gamma^t \nabla_{\theta} r(s_t, a_t; \theta_{k}) \bigg] \bigg \| \right]  \nonumber \\
		&\overset{(i)}{=}  \ee \left[ \bigg \|  \frac{1}{1 - \gamma} \ee_{ (s,a) \sim d(\cdot, \cdot; \pi_{\theta_k}) }\big[ \nabla_{\theta} r(s, a; \theta_{k}) \big]  -  \frac{1}{1 - \gamma} \ee_{(s,a) \sim d(\cdot, \cdot; \pi_{k+1})}  \big[ \nabla_{\theta} r(s, a; \theta_{k}) \big] \bigg \| \right]  \nonumber \\
		&\overset{(ii)}{\leq} \frac{2}{1 - \gamma} \cdot \max_{s \in \mathcal{S},a \in \mathcal{A}} \| \nabla_{\theta} r(s, a; \theta_{k}) \| \cdot \ee \big[ \| d(\cdot, \cdot;\pi_{\theta_k}) - d(\cdot, \cdot;\pi_{k+1}) \|_{TV} \big] \nonumber \\
		&\overset{(iii)}{\leq} \frac{2L_r}{1 - \gamma}   \ee \big[ \| d(\cdot, \cdot;\pi_{\theta_k}) - d(\cdot, \cdot;\pi_{k+1}) \|_{TV} \big] \nonumber \\
		&\overset{(iv)}{\leq}  2L_q C_d  \ee \left[  \|  \log \frac{\pi^0(a|s)\exp Q^\soft_{r_{\theta_k}, \pi_{\theta_k} }(s,a)}{\sum_{\tilde{a}}\pi^0(\tilde{a}|s)\exp Q^\soft_{r_{\theta_k}, \pi_{\theta_k} }(s,\tilde{a})} - \log \frac{\pi^0(a|s)\exp Q^\soft_{r_{\theta_k}, \pi_{k+1} }(s,a)}{\sum_{\tilde{a}}\pi^0(\tilde{a}|s)\exp Q^\soft_{r_{\theta_k}, \pi_{k+1} }(s,\tilde{a})}  \| \right]  \nonumber \\
        &\overset{(v)}{\leq} 2L_q C_d \ee \left[  \|  Q^\soft_{r_{\theta_k}, \pi_{\theta_k} } - Q^\soft_{r_{\theta_k}, \pi_{k}}  \| +\| \log \sum_a \pi^0(\tilde{a}|s) \exp  Q^\soft_{r_{\theta_k}, \pi_{\theta_k}}(s,\tilde{a})-\log \sum_a \pi^0(\tilde{a}|s) \exp  Q^\soft_{r_{\theta_k}, \pi_{k+1}}(s,\tilde{a}) \|\right]  \nonumber \\
		&\overset{(vi)}{\leq} 2L_q C_d \sqrt{|\mathcal{S} | \cdot | \mathcal{A} |}  \ee \left[  \|  Q^\soft_{r_{\theta_k}, \pi_{\theta_k} } - Q^\soft_{r_{\theta_k}, \pi_{k}}  \|_{\infty} +\|  Q^\soft_{r_{\theta_k}, \pi_{\theta_k} } - Q^\soft_{r_{\theta_k}, \pi_{k}}  \|_{\infty} \right] \nonumber \\
  &= 4L_q C_d \sqrt{|\mathcal{S} | \cdot | \mathcal{A} |}  \ee \left[  \|  Q^\soft_{r_{\theta_k}, \pi_{\theta_k} } - Q^\soft_{r_{\theta_k}, \pi_{k}}  \|_{\infty}  \right]
  \label{bound:trajectory_mismatch}
	\end{align}
	where (i) follows the definition $d(s,a;\pi) = (1-\gamma)\pi(a|s)\sum_{t\geq0} \gamma^t \mathcal{P}^{\pi}(s_t = s | s_0 \sim \eta)$; (ii) is due to distribution mismatch between two visitation measures; (iii) follows the inequality \eqref{ineq:reward_grad_bound} in Assumption \ref{Assumption:reward_grad_bound}; the inequality (iv) follows Lemma \ref{lemma:Lipschitz_visitation_measure} and the fact that $\pi_{\theta_k}(\cdot | s) \propto \pi^0(\cdot|s) \exp \big(Q^\soft_{r_{\theta_k}, \pi_{\theta_k}}(s, \cdot) \big)$, $\pi_{k + 1}(\cdot | s) \propto \pi^0(\cdot|s) \exp \big(Q^\soft_{r_{\theta_k}, \pi_k}(s, \cdot) \big)$ and the constant $L_q := \frac{L_r}{1 - \gamma}$; (v) follows the \eqref{ineq:Lipscitz_log_exp_operator};(vi) follows the conversion between Frobenius norm and infinity norm. 
 
    Through plugging the inequality \eqref{bound:trajectory_mismatch} into \eqref{bound:objective_gradient_ascent}, it leads to 
	\begin{align}
		& \mathbb{E} \left[ L(\theta_{k+1}) \right] \nonumber \\
		&\geq  \mathbb{E} \left[ L(\theta_{k}) \right]  - 2\alpha L_q  \ee \left[ \big \| \ee_{\tau \sim \pi_{\theta_{k}}}\big[ \sum_{t \geq 0} \gamma^t \nabla_{\theta} r(s_t, a_t; \theta_{k}) \big]  -  \ee_{\tau \sim \pi_{k+1}}\big[ \sum_{t \geq 0} \gamma^t \nabla_{\theta} r(s_t, a_t; \theta_{k}) \big] \big \| \right] \nonumber \\ 
		& \quad + \alpha \mathbb{E} \bigg[ \|  \nabla L(\theta_{k}) \|^2 \bigg] - 8L_c L_q^2 \alpha^2 \nonumber \\
		&\overset{(i)}{\geq} \mathbb{E} \left[ L(\theta_{k}) \right] - 8\alpha C_d L_q^2 \sqrt{|\mathcal{S}| \cdot | \mathcal{A} |}  \ee \left[  \|  Q^\soft_{r_{\theta_k}, \pi_{\theta_k} } - Q^\soft_{r_{\theta_k}, \pi_{k}} \|_{\infty} \right] + \alpha \mathbb{E} \bigg[ \|  \nabla L(\theta_{k}) \|^2 \bigg] - 8L_c L_q^2 \alpha^2 \nonumber
	\end{align}
	where (i) follows the inequality \eqref{bound:trajectory_mismatch}. 
	
	Rearranging the inequality above and denote $C_1 := 8C_d L_q^2 \sqrt{|\mathcal{S}| \cdot |\mathcal{A}|}$, it holds that \begin{align}
		\alpha \mathbb{E} \big[ \|  \nabla L(\theta_{k}) \|^2 \big] \leq 8L_c L_q^2 \alpha^2 +  \alpha C_1 \ee \left[ \|  Q_{r_{\theta_{k}}, \pi_{\theta_{k}}}^\soft -  Q^\soft_{r_{\theta_{k}}, \pi_{k} }  \|_{\infty} \right] + \ee \big[ L(\theta_{k+1}) - L(\theta_{k}) \big] \nonumber \nonumber
	\end{align} 
	Summing the inequality above from $k = 0$ to $K-1$ and dividing both sides by $\alpha K$, it holds that 
	\begin{align}
		\frac{1}{K}\sum_{k = 0}^{K-1} \ee \left[ \|  \nabla L(\theta_{k}) \|^2 \right] \leq 8L_c L_q^2 \alpha + \frac{C_1}{K} \sum_{k = 0}^{K-1} \ee \left[ \|  Q_{r_{\theta_{k}}, \pi_{\theta_{k}}}^\soft -  Q^\soft_{r_{\theta_{k}}, \pi_{k} }  \|_{\infty} \right] + \ee \left[ \frac{L(\theta_{K}) - L(\theta_{0})}{K \alpha} \right]  \label{bound:sum_upper_grad}
	\end{align}
	Note that the log-likelihood function $L(\theta_K)$ is negative and $L(\theta_0)$ is a bounded constant. Then we could plug \eqref{convergence_rate:soft_Q_gap} into \eqref{bound:sum_upper_grad}, it holds that \begin{align}
		\frac{1}{K}\sum_{K = 0}^{K-1} \ee \left[ \|  \nabla L(\theta_{K}) \|^2 \right]  = \mathcal{O}(K^{-\sigma}) + \mathcal{O}(K^{-1}) + \mathcal{O}(K^{-1 + \sigma}) \label{rate_analysis:lower_problem}
	\end{align}
	which completes the proof for the inequality \eqref{rate:upper_grad_norm}.

\subsection{Auxiliary Lemmas}
\begin{lemma} \label{lemma:Lipschitz_visitation_measure}
	(\cite[Lemma 3]{xu2020improving}) Consider the initialization distribution $\eta(\cdot)$ and transition kernel $\mathcal{P}( \cdot | s,a)$. Under $\eta(\cdot)$ and $\mathcal{P}( \cdot | s,a)$, denote $d_w(\cdot, \cdot)$ as the state-action visitation distribution of MDP with the Boltzman policy parameterized by parameter $w$. Suppose Assumption \ref{Assumption:Ergodicity_Markov_chain} holds, for all policy parameter $w$ and $w^\prime$, we have 
	\begin{align}
		\| d_{w}(\cdot, \cdot) - d_{w^\prime}(\cdot, \cdot)  \|_{TV} \leq C_d \| w - w^\prime \| 	\label{ineq:lipschitz_measure}
	\end{align}
	where $C_d$ is a positive constant.
\end{lemma}

Next, to facilitate analysis for KL-regularized MDPs, we introduce a “soft” Bellman optimality operator $ \mathcal{T}: \mathbb{R}^{|\mathcal{S}| \times |\mathcal{A}|} \to \mathbb{R}^{|\mathcal{S}| \times |\mathcal{A}|}  $ as follows:
	\begin{align}
		\mathcal{T}(Q)(s,a) := r(s,a) + \gamma \ee_{s^\prime \sim \mathcal{P}(\cdot | s,a)} \left[ \max_{\pi(\cdot | s)} ~ \mathbb{E}_{a^\prime \sim \pi(\cdot | s^\prime)} \left[ Q(s^\prime, a^\prime) -\frac{\log \pi(a^\prime | s^\prime)}{\log \pi^0(a^\prime | s^\prime)} \right] \right]. \label{def:soft_Bellman}
	\end{align}
	In the following lemma, the properties of KL-regularized MDPs are characterized.

\begin{lemma} \label{lemma:soft_Bellman_operator}
	(The operator $\mathcal{T}$ as defined in \eqref{def:soft_Bellman} satisfies the properties below:
	\begin{itemize}
		\item  $\mathcal{T}$ has the following closed-form expression:
		\begin{align}
			\mathcal{T}(Q)(s,a) = r(s,a) + \gamma \ee_{s^\prime \sim \mathcal{P}(\cdot | s,a)} \left[ \log \left( \sum_{a^\prime}\pi^0(a^\prime|s^\prime) \exp \big( Q(s^\prime, a^\prime) \big) \right) \right].  \label{closed_form:soft_Bellman}
		\end{align}
		
		\item $ \mathcal{T} $ is a $\gamma$-contraction in the $\ell_\infty$ norm, namely, for any $Q_1, Q_2 \in \mathbb{R}^{|\mathcal{S}| \times |\mathcal{A}|}$, it holds that
		\begin{align}
			\|  \mathcal{T}(Q_1) - \mathcal{T}(Q_2) \|_{\infty} \leq \gamma \| Q_1 - Q_2 \|_{\infty} . \label{ineq:soft_Bellman_contraction}
		\end{align}
		
		\item Under a given reward function $r(\cdot, \cdot)$, the corresponding optimal soft $Q$-function $Q^\soft_{r, \pi^*}$ is a unique fixed point of the operator $\mathcal{T}$, namely, \begin{align}
			\mathcal{T}(Q^\soft_{r, \pi^*}) = Q^\soft_{r, \pi^*} \label{fixed_point:soft_Bellman}
		\end{align}
	\end{itemize}
\end{lemma}

We refine its analysis as below.

We first show that 
\begin{align}
    \mathbb{E}_{a \sim \pi(\cdot | s )} \bigg[ Q(s, a) - \frac{\log \pi(a | s) }{\log \pi^0(a|s)} \bigg]  = \sum_{a} \pi(a|s) \log \left( \frac{ \pi^0(a | s)  \exp( Q(s, a ) )}{\pi( a | s )}  \right) \overset{(i)}{\leq}  \log \bigg( \sum_a  \pi^0(a | s)  \exp \big( Q(s, a ) \big) \bigg) \label{ineq:soft_V_bound}
\end{align}
where (i) is from Jensen's inequality. Moreover, the equality between both sides of (i) holds when the policy $\pi$ has the expression $ \pi(\cdot | s) \propto \pi^0(a | s)  \exp( Q(s, \cdot) ) $. Therefore, through applying the inequality \eqref{ineq:soft_V_bound} to \eqref{def:soft_Bellman}, it obtains that 
\begin{align}
    \mathcal{T}(Q)(s,a) = r(s,a) + \gamma \ee_{s^\prime \sim \mathcal{P}(\cdot | s,a)} \left[ \log \left( \sum_{a^\prime} \pi^0(a | s) \exp \big( Q(s^\prime, a^\prime) \big) \right) \right], \label{eq:soft_bellman_closed_form}
\end{align}
which proves the equality \eqref{closed_form:soft_Bellman}.

We define  $\|Q_1 - Q_2\|_{\infty} := \max_{s \in \mathcal{S}, a \in \mathcal{A}} |Q_1(s, a) - Q_2(s, a)|$ and $\epsilon = \|Q_1 - Q_2\|_{\infty}$. Then for any $s \in \mathcal{S}$ and $a \in \mathcal{A}$, it follows that
\begin{align}
    \log \left( \sum_a  \pi^0(a | s) \exp\big(Q_1 (s,a) \big) \right) &\leq \log \left( \sum_a \pi^0(a | s)  \exp \big(Q_2 (s,a) + \epsilon\big) \right) \nonumber \\
    &= \log \left( \exp(\epsilon) \sum_a  \pi^0(a | s)  \exp \big(Q_2 (s,a)\big) \right) \nonumber \\
    &= \epsilon + \log \left( \sum_a  \pi^0(a | s)  \exp \big(Q_2 (s,a)\big) \right) \nonumber
\end{align}

Similarly, it is easy to obtain that $ \log \left( \sum_a \pi^0(a | s) \exp\big(Q_1 (s,a) \big) \right) \geq -\epsilon + \log \left( \sum_a \pi^0(a | s) \exp \big(Q_2 (s,a)\big) \right)$. Hence, it leads to the contraction property that \begin{align}
    \| \mathcal{T}( Q_1 ) - \mathcal{T} ( Q_2 ) \|_{\infty} \leq  \gamma \epsilon = \gamma \|  Q_1 - Q_2 \|_{\infty} \label{contraction:soft_operator}
\end{align}
which proves the contraction property \eqref{ineq:soft_Bellman_contraction}.
Moreover, we have 
	\begin{align}
		\mathcal{T}(Q^\soft_{r, \pi^*})(s,a) \overset{(i)}{=} r(s,a) + \gamma \ee_{s^\prime \sim \mathcal{P}(\cdot | s,a)} \left[ \log \left( \sum_{a^\prime} \pi^0(a^\prime|s^\prime) \exp \big( Q^\soft_{r, \pi^*}(s^\prime, a^\prime) \big) \right) \right] \overset{(ii)}{=} Q^\soft_{r, \pi^*}(s,a) \label{eq:fixed_optimal_soft_Q}
	\end{align}
	where (i) follows the equality \eqref{eq:soft_bellman_closed_form}. Based on the definition of the soft Q-function $ Q^\soft_{r, \pi^*} $, we have 
	\begin{align}
	    Q^\soft_{r, \pi^*}(s,a) = r(s,a) + \gamma \ee_{s^\prime \sim \mathcal{P}(\cdot | s,a)} \big[ \ee_{a^\prime \sim \pi^*(\cdot | s^\prime)} [ -\frac{\log \pi^*(a^\prime|s^\prime)}{\log \pi^0(a^\prime|s^\prime)} + Q^\soft_{r, \pi^*}(s^\prime,a^\prime) ] \big]. \label{rewrite:soft_Q_definition}
	\end{align}
	We prove the equality (ii) in \eqref{eq:fixed_optimal_soft_Q} through combining \eqref{rewrite:soft_Q_definition} and the fact that the optimal soft policy has the closed form $\pi^*(\cdot | s) \propto \pi^0(\cdot |s^\prime) \exp\big( Q^\soft_{r, \pi^*}(s, \cdot) \big)$. Suppose two different fixed points of the soft Bellman operator exist, then it contradicts with the contraction property in \eqref{contraction:soft_operator}. 
	
	Hence, we proved the uniqueness of the optimal soft $Q$-function $Q^\soft_{r, \pi^*}$. Moreover, the optimal soft $Q$-function $Q^\soft_{r, \pi^*}$ is a fixed point to the soft Bellman operator $\mathcal{T}$ in \eqref{fixed_point:soft_Bellman}.
% Moreover, we have 
% \begin{align}
%     \mathcal{T}(Q^\soft_{r, \pi^*})(s,a) \overset{(i)}{=} r(s,a) + \gamma \ee_{s^\prime \sim \mathcal{P}(\cdot | s,a)} \left[ \log \left( \sum_{a^\prime} \exp \big( Q^\soft_{r, \pi^*}(s^\prime, a^\prime) \big) \right) \right] \overset{(ii)}{=} Q^\soft_{r, \pi^*}(s,a) \label{eq:fixed_optimal_soft_Q}
% \end{align}
% where (i) follows the equality \eqref{eq:soft_bellman_closed_form}. Based on the definition of the soft Q-function $ Q^\soft_{r, \pi^*} $, we have 
% \begin{align}
%     Q^\soft_{r, \pi^*}(s,a) = r(s,a) + \gamma \ee_{s^\prime \sim \mathcal{P}(\cdot | s,a)} \big[ \ee_{a^\prime \sim \pi^*(\cdot | s^\prime)} [ -\log \pi^*(a^\prime|s^\prime) + Q^\soft_{r, \pi^*}(s^\prime,a^\prime) ] \big]. \label{rewrite:soft_Q_definition}
% \end{align}
% We prove the equality (ii) in \eqref{eq:fixed_optimal_soft_Q} through combining \eqref{rewrite:soft_Q_definition} and the fact that the optimal soft policy has the closed form $\pi^*(\cdot | s) \propto \exp\big( Q^\soft_{r, \pi^*}(s, \cdot) \big)$. Suppose two different fixed points of the soft Bellman operator exist, then it contradicts with the contraction property in \eqref{contraction:soft_operator}. 

% Hence, we proved the uniqueness of the optimal soft $Q$-function $Q^\soft_{r, \pi^*}$. Moreover, the optimal soft $Q$-function $Q^\soft_{r, \pi^*}$ is a fixed point to the soft Bellman operator $\mathcal{T}$ in \eqref{fixed_point:soft_Bellman}.

\end{document}